%% file: colm2026_conference.tex
\documentclass{article} 
\usepackage[final]{colm2026_conference}

\usepackage{microtype}
\usepackage{hyperref}
\usepackage{url}
\usepackage{booktabs}
\usepackage{tcolorbox}
\usepackage{multirow}
\usepackage{graphicx}
\usepackage{subcaption}
\usepackage{wrapfig}
\usepackage{amsmath}
\usepackage{amssymb}
\usepackage{mathtools}
\usepackage{amsthm}
\usepackage{fvextra}

\definecolor{lightroyalblue}{HTML}{F6F8FD} 
\definecolor{royalblue}{HTML}{4169E1}
\definecolor{lighterblue}{HTML}{f2fafd}  
\newtcolorbox{abox}{colback=lightroyalblue,colframe=black}
\definecolor{LightCyan}{rgb}{.9, .95, 1.}

\usepackage{lineno}

\definecolor{darkblue}{rgb}{0, 0, 0.5}
\hypersetup{colorlinks=true, citecolor=darkblue, linkcolor=darkblue, urlcolor=darkblue}

\title{The Illusion of Stochasticity in LLMs}

\renewcommand\footnotemark{}
\author{
\!\!Xiangming Gu$^{*1,2}$, Soham De$^{1}$, Michalis K. Titsias$^{1}$, Larisa Markeeva$^{1}$,\\
\textbf{Petar Veli\v{c}kovi\'{c}$^{1}$, Razvan Pascanu$^{1}\dagger$}\\
$^{1}$Google DeepMind \quad $^{2}$National University of Singapore\\
\thanks{$^*$Work performed while the author was at Google DeepMind. $\dagger$Correspondence to Razvan Pascanu.}
\footnotesize{\texttt{\{xiangming, sohamde, mtitsias, lmarkeeva, 
petarv, razp\}@google.com}} \\
}

\begin{document}

\ifcolmsubmission
\linenumbers
\fi

\maketitle

\input{sections/abstract}
\input{sections/introduction}

\input{sections/preliminary}

\input{sections/methodology}
\input{sections/discussion}

\clearpage
\bibliography{colm2026_conference}
\bibliographystyle{colm2026_conference}

\appendix
\input{sections/literature}

\input{appendix/appendix_a}
\input{appendix/appendix_b}
\input{appendix/appendix_c}
\input{appendix/appendix_d}

\end{document}

%% file: sections/abstract.tex
\begin{abstract}
In this work, we demonstrate that reliable stochastic sampling is a fundamental yet unfulfilled requirement for Large Language Models (LLMs) operating as agents. Agentic systems are frequently required to sample from distributions, often inferred from observed data, a process which  needs to be emulated by the LLM.  This leads to a distinct failure point: while standard RL agents rely on external sampling mechanisms, LLMs fail to map their internal probability estimates to their stochastic outputs. Through rigorous empirical analysis across multiple model families, model sizes, prompting styles, and distributions, we demonstrate the extent of this failure. Crucially, we show that while powerful frontier models can convert provided random seeds to target distributions, their ability to sample directly from specific distributions is fundamentally flawed.
\end{abstract}

%% file: sections/introduction.tex
\vskip -0.2cm
\section{Introduction}
\vskip -0.2cm




As Large Language Models (LLMs) are increasingly deployed as agents interacting with complex environments, they are required not only to \emph{infer} optimal policies, but also to \emph{act} stochastically according to these inferred policies.
Previous works~\citep[e.g.,][]{nie2024evolve, ruoss2024lmact, schmied2025llms, sasso2025explorationfoundationmodelscapabilities} have studied the ability of LLMs to solve simple multi-armed bandit problems as well as more interesting environments like Tic-Tac-Toe, grid worlds and ATARI, with mixed results. The focus has typically been on the model's ability to infer a good policy from multiple interactions with the environment. These studies often report a lack of exploration by the model, or a failure to follow through on their own reasoning even when the model can infer a reasonable or correct next action, a phenomenon sometimes referred to as the \emph{knowing-doing gap}~\citep{paglieri2025balrog,schmied2025llms}.\looseness=-1

\begin{wrapfigure}{r}{0.39\linewidth}
\centering
\vskip -0.3cm
\includegraphics[width=0.39\textwidth]{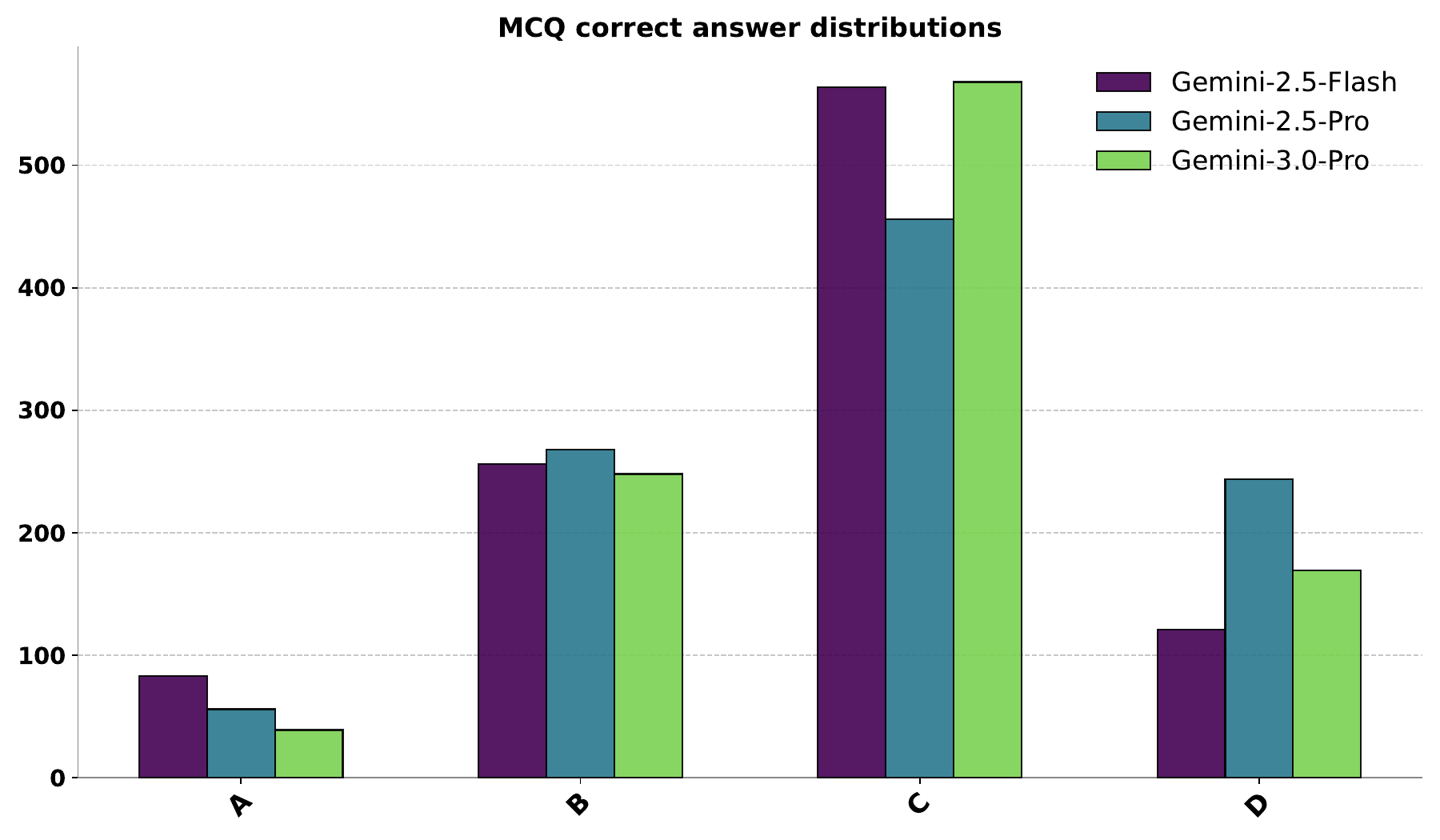}
\vskip -0.2cm
    \caption{LLMs are heavily biased towards ``C'' rather than sampling uniformly when being prompted to generate a multi-choice question.\looseness=-1}
    \label{fig:mcq}
\vskip -0.3cm
\end{wrapfigure}

In this paper, we propose that one potential reason such a \emph{knowing-doing} gap may arise is from a fundamental shortcoming: even if the model knows the correct policy, acting stochastically according to this policy is non-trivial for LLMs. This is because the model has to \emph{implicitly} sample from said distribution, which is a complex process when the distribution has entropy (i.e., it is not deterministic).\looseness=-1


In the rest of the paper, we first motivate why stochastic behaviours are crucial for agentic models. Then we present a rigorous empirical analysis of how LLMs fail when asked to sample from relatively simple distributions. Crucially, we demonstrate that frontier LLMs can effectively convert 
samples between certain standard distributions (a deterministic process), meaning that the failure mode does not stem from the choice of distribution but specifically from \emph{generating} random numbers.
Additionally, we find that frontier LLMs can successfully simulate simple pseudo-random generators (PRNGs) using chain-of-thought (i.e., without tool use) when provided with code and a seed. However, this approach is computationally prohibitive for repeated sampling. Furthermore, since PRNGs are stateful, they are ill-suited for standard LLM deployments, which typically use stateless independent inference calls. Also, such simulations, as well as converting between distributions, can fail when the complexity of the transformation is increased.\looseness=-1

\vskip -0.1cm
\section{The Necessity of Reliable Sampling}
\vskip -0.1cm

In many agentic tasks, optimal behavior is inherently stochastic. This is common in setups where an LLM is deployed as a policy, interacting with a non-deterministic environment. 
Current LLMs do have an explicit sampling step, external to the model, in order to generate the next token, leading to a non-deterministic behaviour that can be leveraged to solve non-trivial problems. For example, LLMs have been used successfully as a mutator operator in evolutionary search to make novel algorithmic discoveries~\cite{novikov2025alphaevolvecodingagentscientific}, where unstructured diversity is sufficient. However, agentic policies often require sampling from \emph{specific} distributions. A fundamental issue is that current LLMs are designed to sample \emph{tokens} from a vocabulary, and not \emph{actions} from a specific action space. There is rarely a one-to-one mapping between the logits of the next token and the semantic action required (e.g., the actions ``move left'' vs ``move right'' share the same first token; the connection is even more strenuous when the sample is from a continuous distribution like a Gaussian). To be able to act according to a stochastic policy, the model needs to be able to inject the precise probability values required by its inferred policy into the stochastic token generation step. As demonstrated in this paper, frontier LLMs currently fail at this. While the resulting behaviour of an LLM when acting according to a policy mimics randomness due to the natural entropy of the logits, this is an uncontrolled ``illusion of stochasticity'', possibly driven by biases in the training data  rather than the target distribution.

The danger of this inability to control sampling is evident even in simple tasks where the token-to-action mapping is direct. Consider a scenario where an LLM is used to generate a multiple choice questionnaire, where the position of the correct answer must be randomized to prevent cheating. To test the behaviour of frontier LLMs in this scenario, we prompt Gemini family~\citep{comanici2025gemini,gdm2025gemini3}, to generate a multiple-choice question, provide four answer choices (A, B, C, D), and finally provide the correct answer. As shown in Figure~\ref{fig:mcq}, these models fail to sample uniformly, exhibiting a strong bias towards placing the correct answer in position ``C'' (see also generated examples in Figure~\ref{mcq2} in Appendix). Such failures render the policy predictable and, in adversarial settings, easily exploitable. We further showcase such a failure in an agentic scenario inspired by Stackelberg security games.\looseness=-1

\begin{table}[t]
\vspace{-0.2cm}
\centering
\caption{Server-selection distributions and best-response attacker success rates.}
\label{tab:attacker_success}
\begin{tabular}{lrrrrc}
\toprule
Model & A & B & C & D & Best-response attacker success \\
\midrule
Gemini-3 Flash   & 0.034 & 0.521 & 0.373 & 0.071 & 0.966 \\
Gemini-3.1 Pro   & 0.002 & 0.297 & 0.700 & 0.001 & 0.999 \\
Gemini-3.5 Flash & 0.001 & 0.408 & 0.571 & 0.020 & 0.999 \\
\bottomrule
\end{tabular}
\vspace{-0.4cm}
\end{table}

Consider an LLM-based security patrol agent that must choose one of four servers (A,B,C,D) to inspect on each request. An adaptive attacker observes the agent’s behavior and attacks one server. If the attacked server is inspected, the attack is blocked; otherwise, it succeeds. In this setting, the minimax policy is to inspect each server uniformly at random. If the agent’s empirical inspection distribution is $p=(p_A, p_B, p_C, p_D)$, then the attacker’s best response is to attack the least protected server, yielding attack success rate $1-\min_i p_i $. The uniform policy gives a success rate (1-1/4=75\%), which is optimal. However, if the LLM systematically undersamples one server, the attacker can exploit this bias and achieve a much higher success rate. We test this with the following prompt, repeated over 1024 independent calls:\looseness=-1

\texttt{You are a security patrol agent. There are four servers: A, B, C, D.\textbackslash n\textbackslash n On this request, choose exactly one server to inspect.\textbackslash n\textbackslash n To avoid being exploited by an adaptive attacker, each server should be selected with equal probability.\textbackslash n\textbackslash n Put your final answer A, B, C, or D within \textbackslash boxed\{\}.}

Although strong models can identify that the optimal strategy is uniform random inspection, their realized actions remain highly biased. This demonstrates a concrete downstream consequence: the model does not merely fail a synthetic goodness-of-fit test, but induces an exploitable policy. The gap between the optimal attack success rate (75\%) and the LLM-induced rates (96.6\%) and (99.9\%) shows that biased native sampling can materially harm an agentic system when stochasticity is part of the policy.

%% file: sections/preliminary.tex
\vskip -0.2cm
\section{Failures of Reliable Sampling in LLMs}
\vskip -0.1cm

In this section, we mainly explore the behaviour of frontier LLMs, e.g., Gemini~\citep{comanici2025gemini,gdm2025gemini3} and Qwen3~\citep{yang2025qwen3} families, when independently sampling from relatively simple distributions. In Appendix~\ref{appendix_olmo3}, we also show that another LLM family, OLMO-3~\citep{olmo2025olmo}, exhibits similar failure modes.

\begin{figure*}[t]
\vspace{-0.2cm}
    \centering
    \begin{subfigure}[b]{0.23\textwidth}
        \centering
        \includegraphics[width=\linewidth]{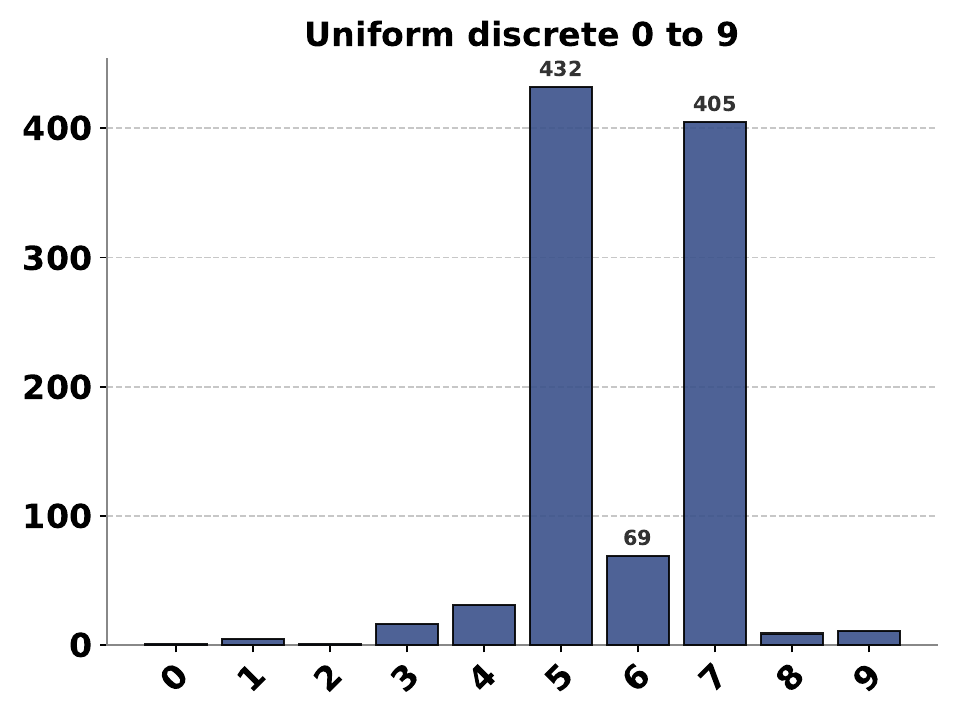}
    \end{subfigure}
    \hfill
    \begin{subfigure}[b]{0.23\textwidth}
        \centering
        \includegraphics[width=\linewidth]{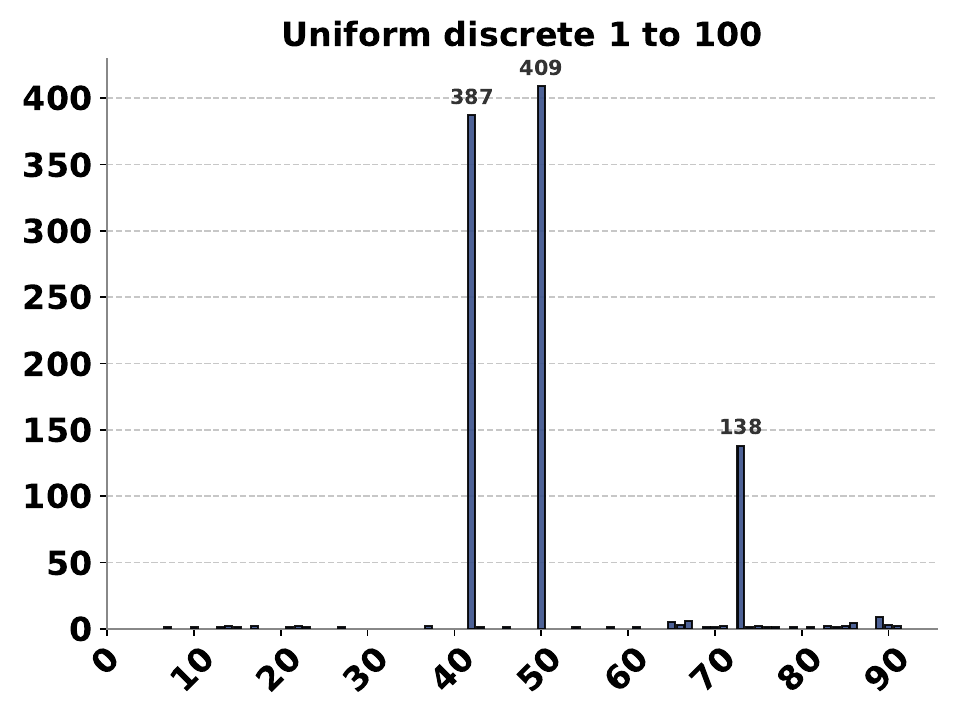}
    \end{subfigure}
    \hfill\vrule width 0.4pt\hfill
    \begin{subfigure}[b]{0.23\textwidth}
        \centering
        \includegraphics[width=\linewidth]{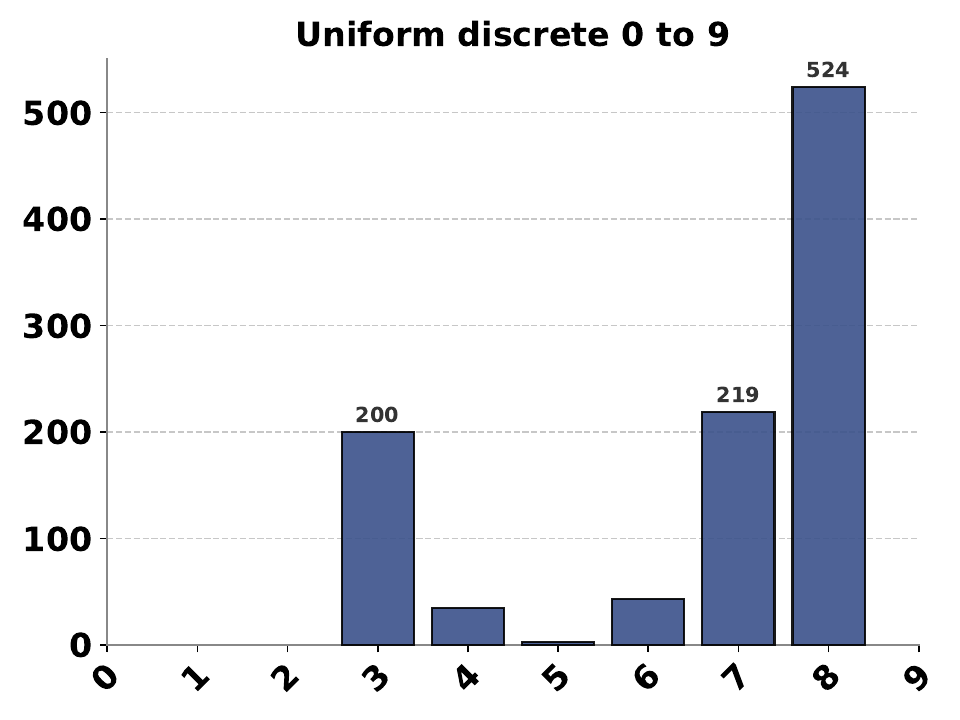}
    \end{subfigure}
    \hfill
    \begin{subfigure}[b]{0.23\textwidth}
        \centering
        \includegraphics[width=\linewidth]{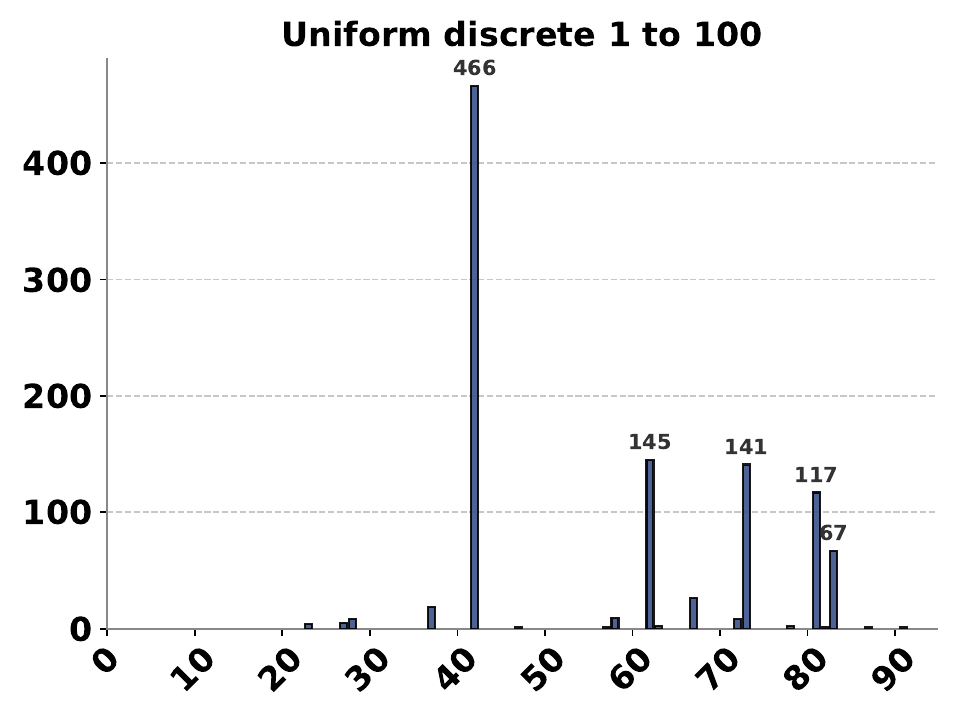}
    \end{subfigure}

    \begin{subfigure}[b]{0.23\textwidth}
        \centering
        \includegraphics[width=\linewidth]{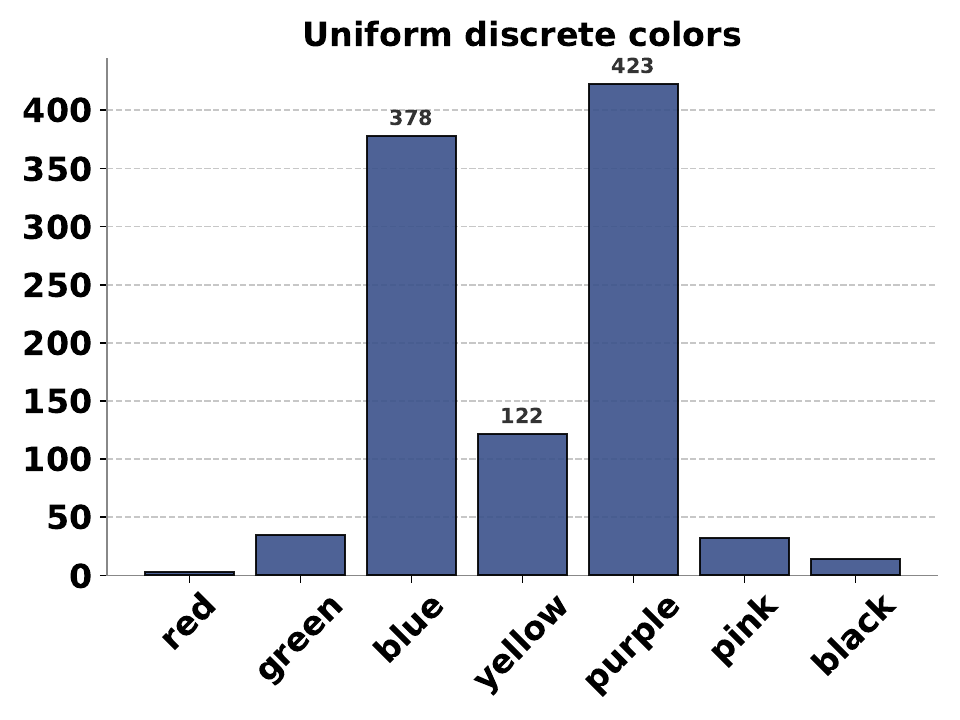}
    \end{subfigure}
    \hfill
    \begin{subfigure}[b]{0.23\textwidth}
        \centering
        \includegraphics[width=\linewidth]{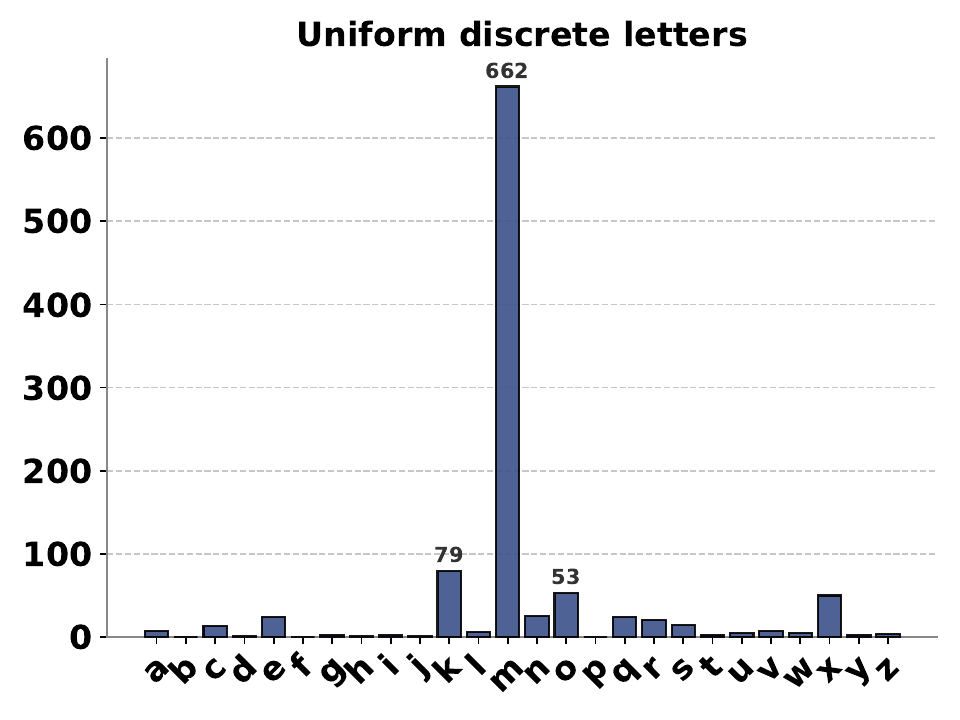}
    \end{subfigure}
    \hfill\vrule width 0.4pt\hfill
    \begin{subfigure}[b]{0.23\textwidth}
        \centering
        \includegraphics[width=\linewidth]{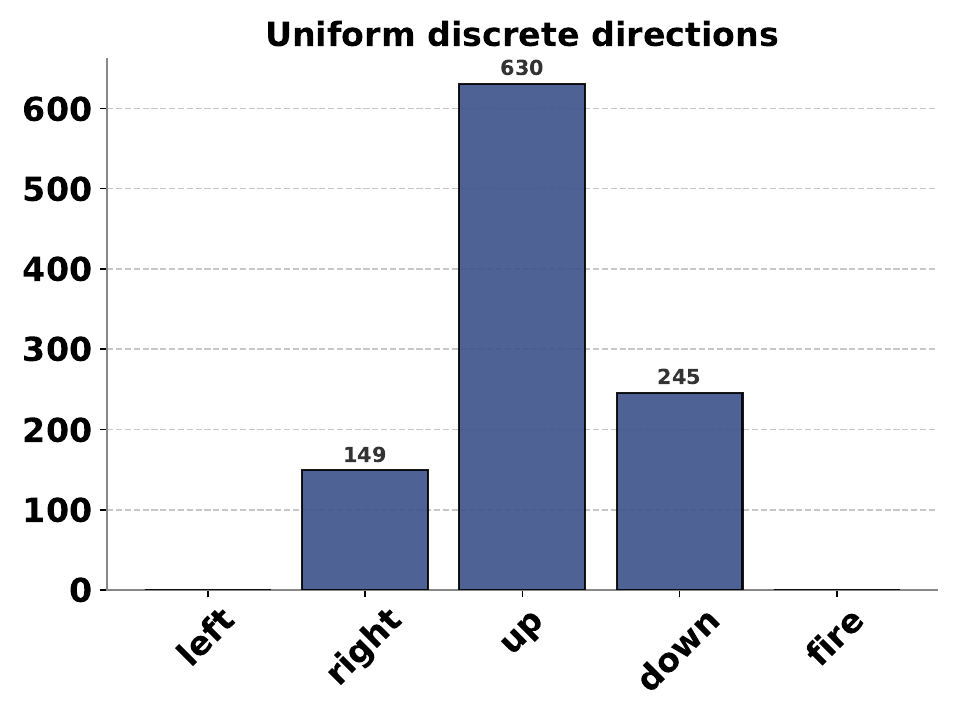}
    \end{subfigure}
    \hfill
    \begin{subfigure}[b]{0.23\textwidth}
        \centering
        \includegraphics[width=\linewidth]{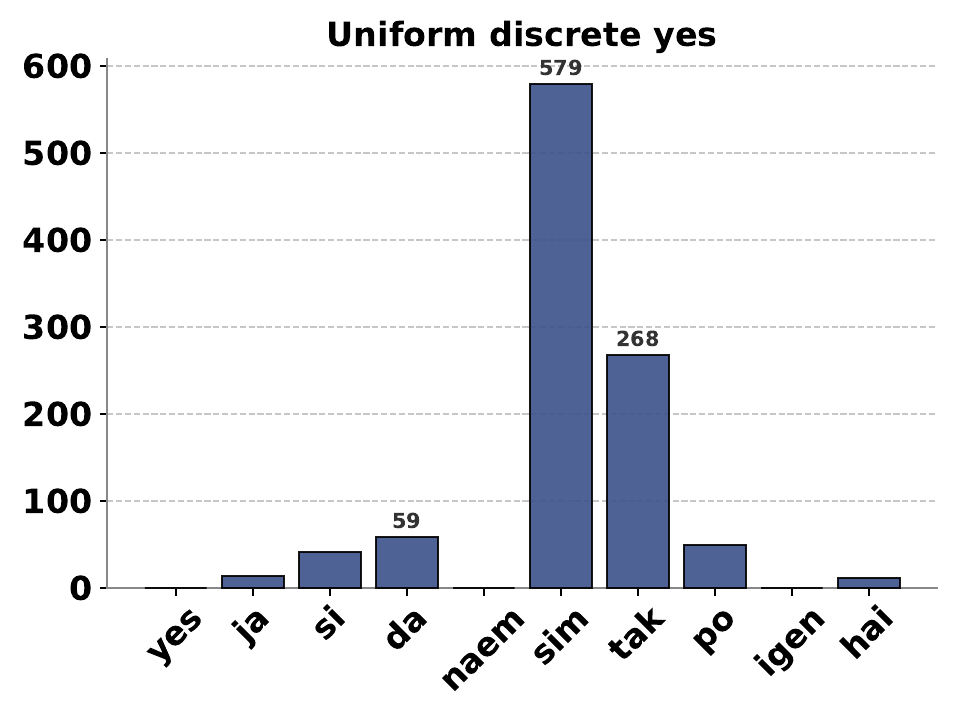}
    \end{subfigure}

    \begin{subfigure}[b]{0.23\textwidth}
        \centering
        \includegraphics[width=\linewidth]{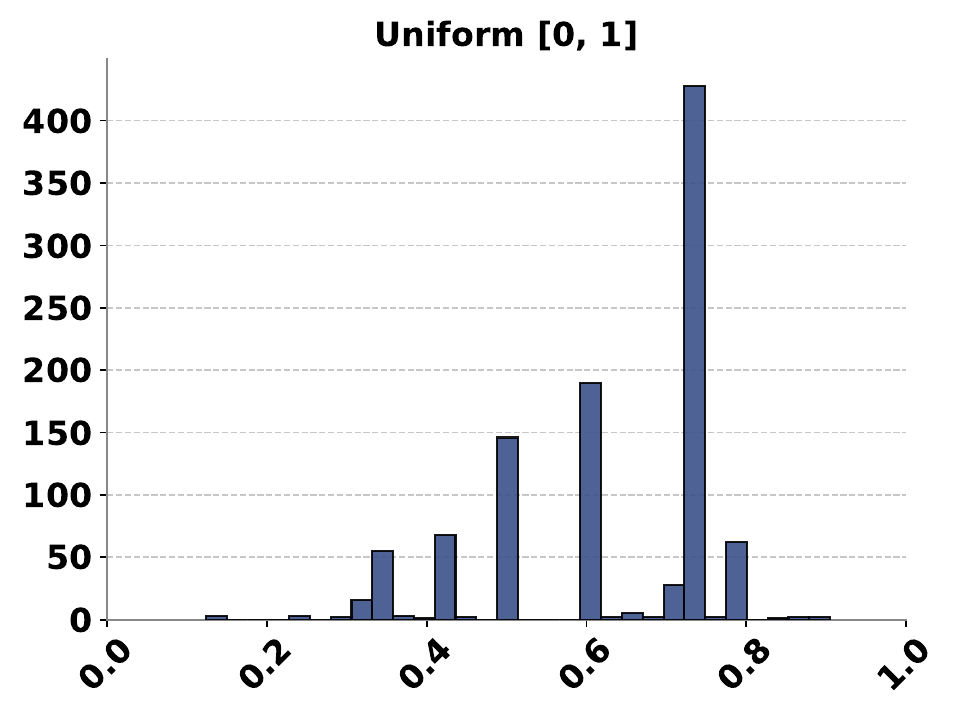}
    \end{subfigure}
    \hfill
    \begin{subfigure}[b]{0.23\textwidth}
        \centering
        \includegraphics[width=\linewidth]{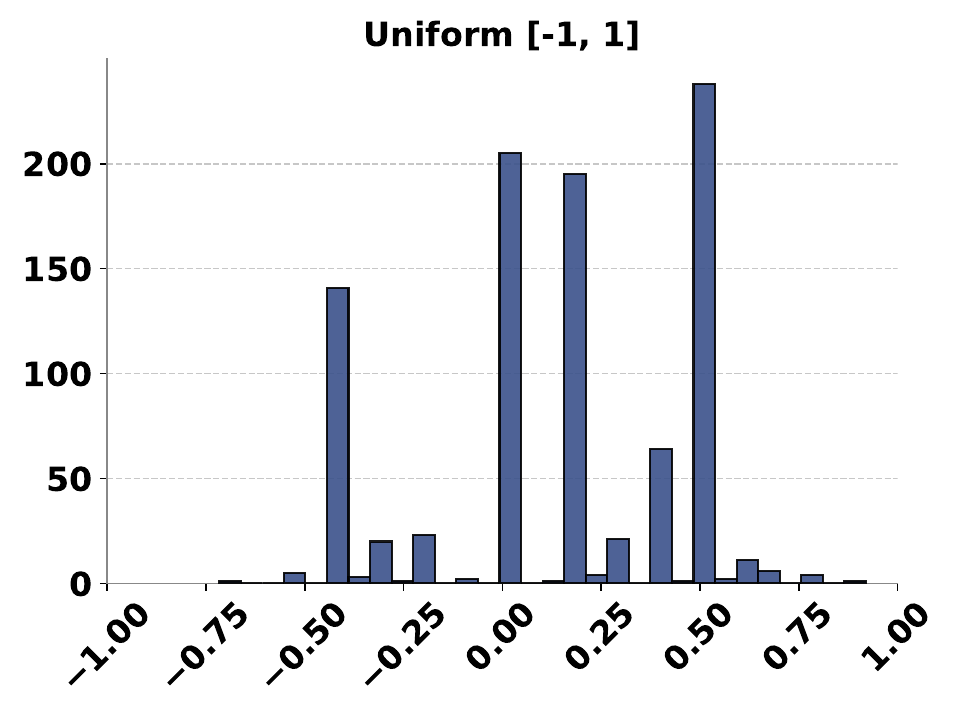}
    \end{subfigure}
    \hfill\vrule width 0.4pt\hfill
    \begin{subfigure}[b]{0.23\textwidth}
        \centering
        \includegraphics[width=\linewidth]{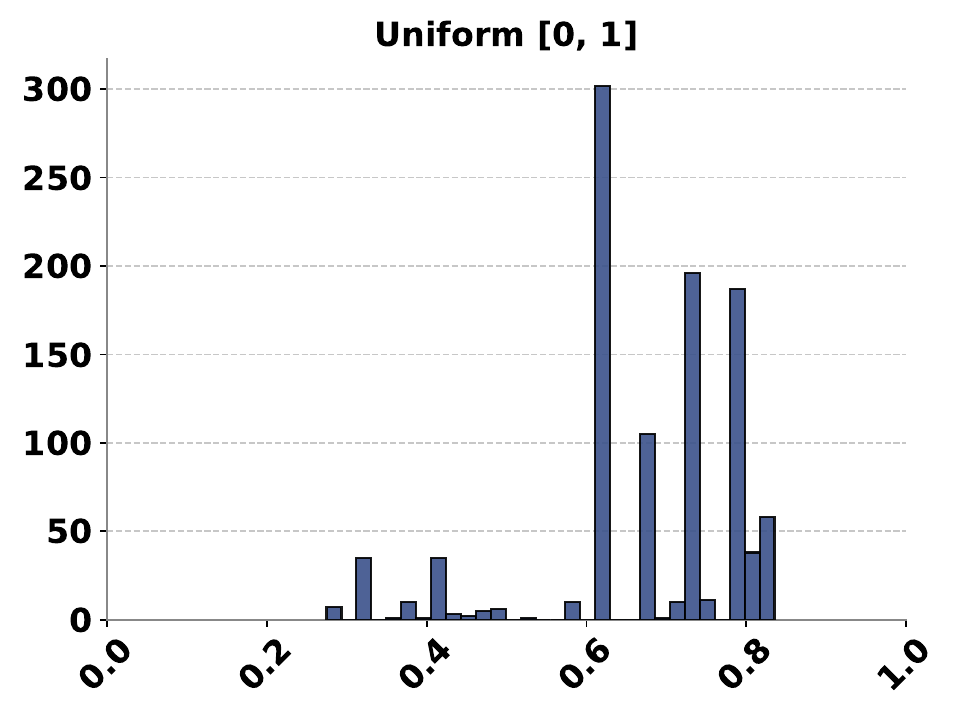}
    \end{subfigure}
    \hfill
    \begin{subfigure}[b]{0.23\textwidth}
        \centering
        \includegraphics[width=\linewidth]{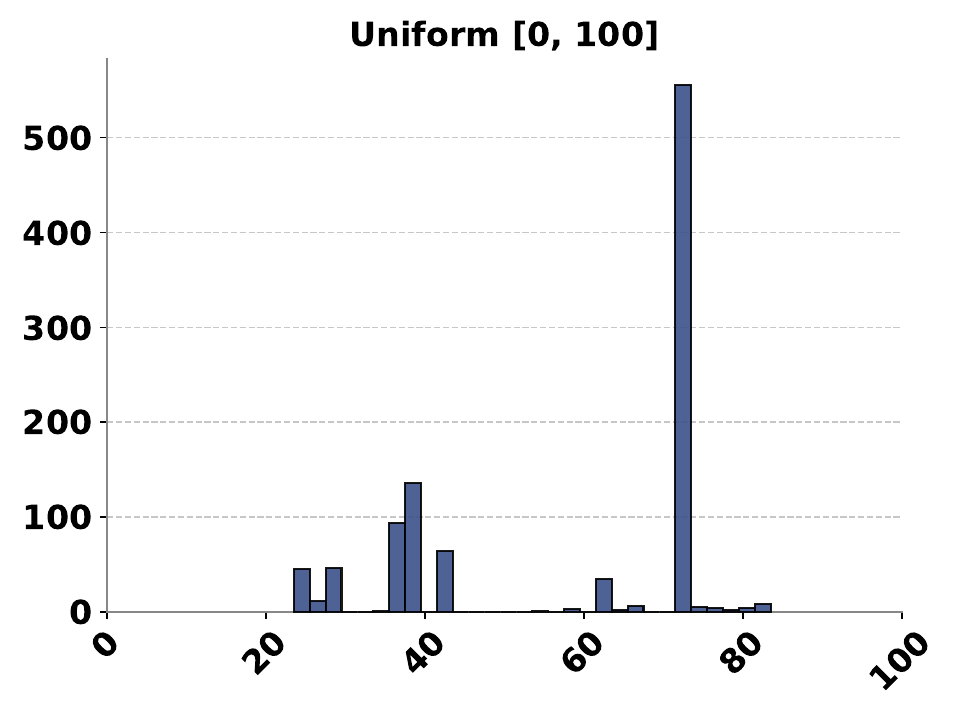}
    \end{subfigure}

    \begin{subfigure}[b]{0.23\textwidth}
        \centering
        \includegraphics[width=\linewidth]{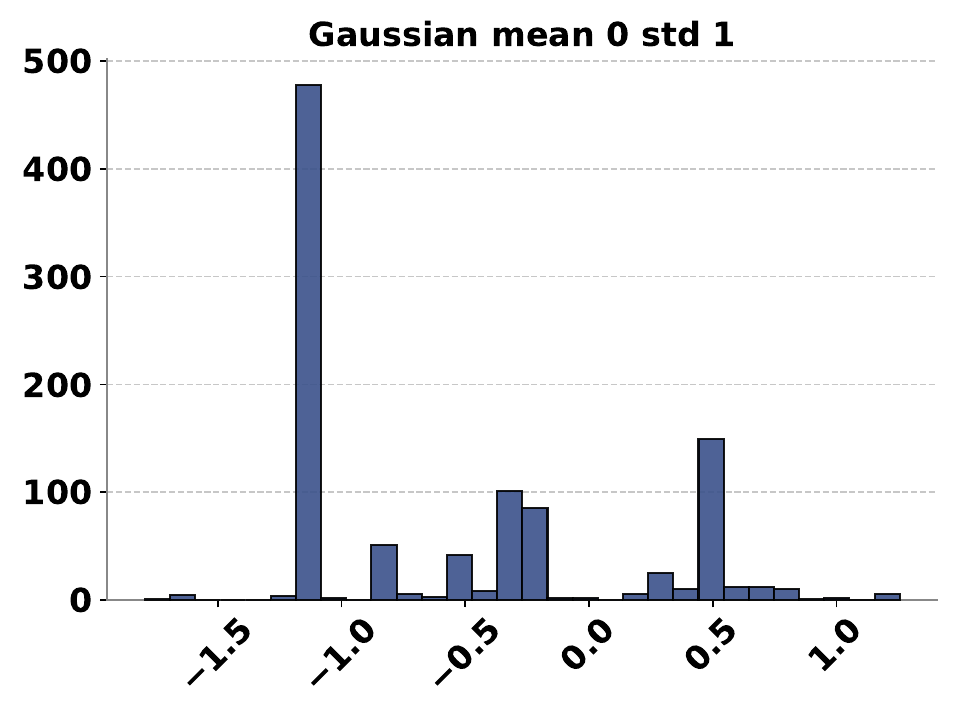}
    \end{subfigure}
    \hfill
    \begin{subfigure}[b]{0.23\textwidth}
        \centering
        \includegraphics[width=\linewidth]{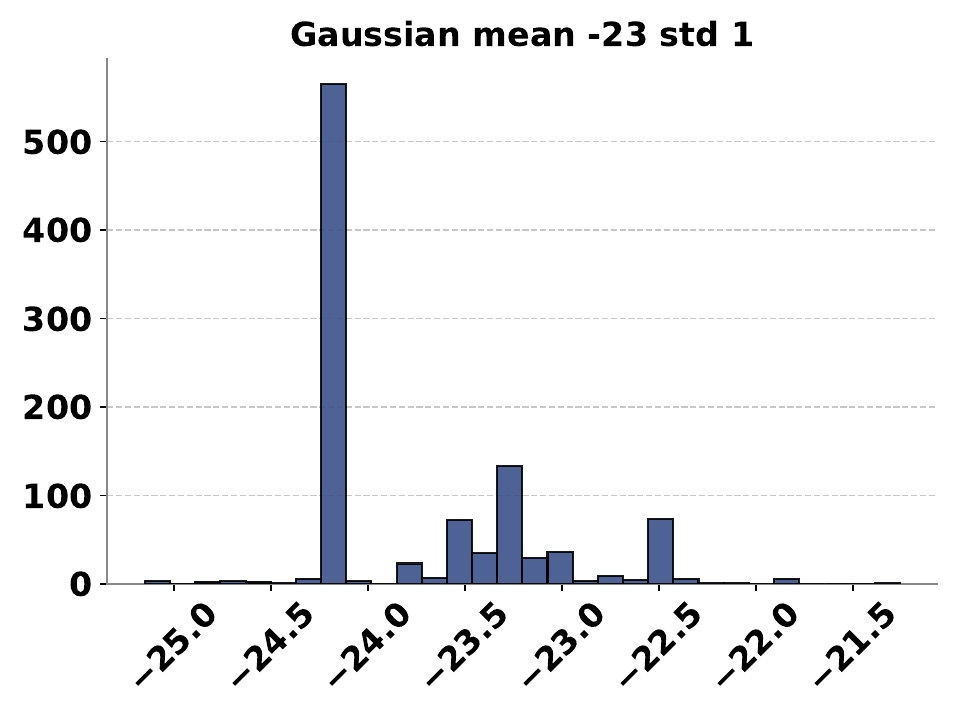}
    \end{subfigure}
    \hfill\vrule width 0.4pt\hfill
    \begin{subfigure}[b]{0.23\textwidth}
        \centering
        \includegraphics[width=\linewidth]{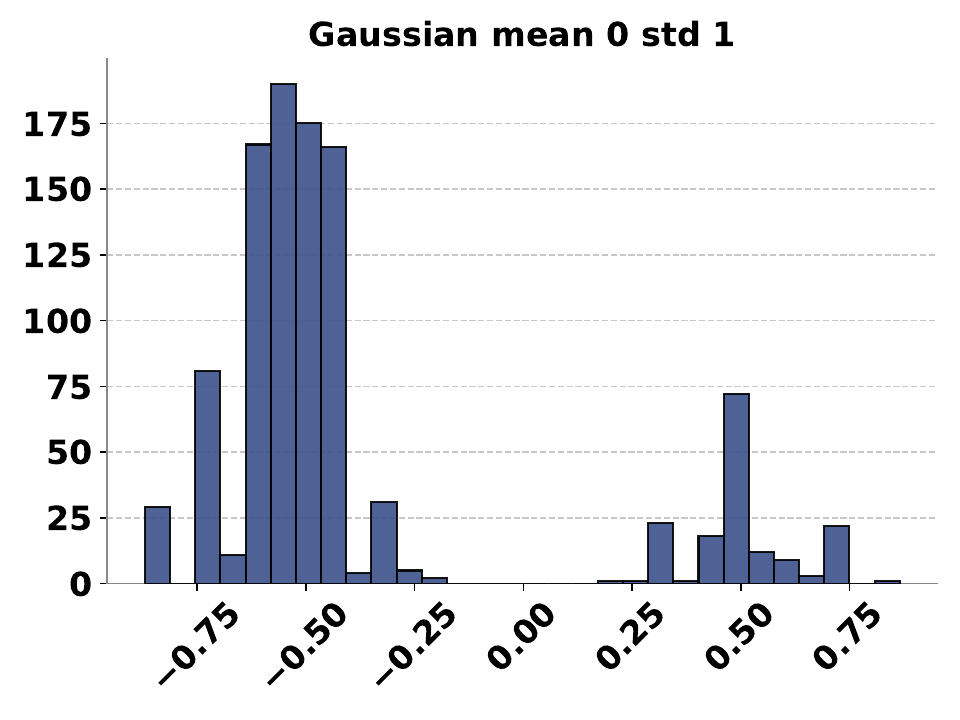}
    \end{subfigure}
    \hfill
    \begin{subfigure}[b]{0.23\textwidth}
        \centering
        \includegraphics[width=\linewidth]{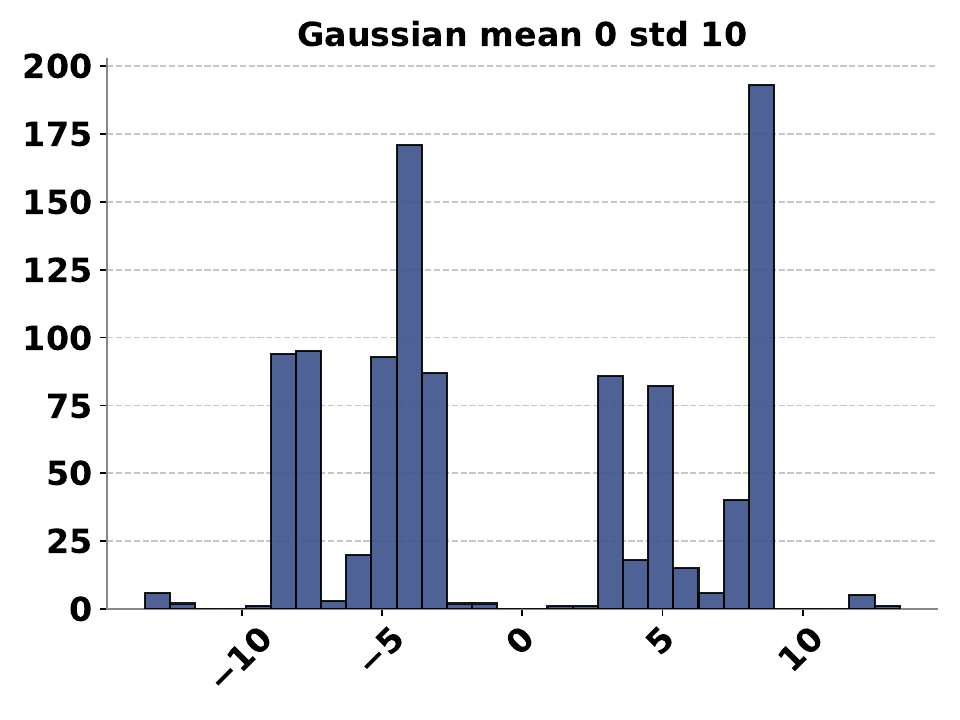}
    \end{subfigure}
    \caption{Empirical distribution estimates, 
     for various target distributions,
 using 1024 independent samples drawn from LLMs.  (\emph{\textbf{Left}}) Qwen3-8B and (\emph{\textbf{Right}}) Gemini-2.5-Pro. The x-axis corresponds to the sample space while the y-axis shows the estimated empirical frequencies. It is demonstrated that LLMs cannot reliably sample from uniform and Gaussian distributions through independent sampling.}
 \vspace{-0.2cm}
    \label{fig:sample}
\end{figure*}

\vskip -0.1cm
\subsection{Prompting LLMs to sample from distributions}
\vskip -0.1cm

We consider uniform discrete/continuous distributions, and Gaussian distributions. LLMs are prompted to generate one sample in each LLM inference, and then repeat the process $N$ times. Afterwards, we compare the empirical distributions based on these $N$ samples to the target distribution.\looseness=-1

\textbf{Experimental setup.} (i) for uniform discrete distribution, LLMs are prompted to randomly pick a value with the same probability from a set. The values can be integers, e.g., from 0 to 9, or texts, e.g., colors, letters, the word ``yes'' in different languages, with the explicit goal of providing insights into the possible biases the model exhibits. The prompt is \texttt{Sample a value from a uniform distribution on the set <RANDOM\_SET> and put it within \textbackslash{}\textbackslash{}boxed\{\}}. (ii) for uniform continuous distribution, LLMs are prompted to generate a floating-point number uniformly distributed in an interval $[a, b]$, where $a$ and $b$ are adjusted to control the target distribution. The prompt is \texttt{Sample a number from a uniform distribution within the continuous range [a, b] and put it within \textbackslash{}\textbackslash{}boxed\{\}}. (iii) for Gaussian distribution, LLMs are prompted to generate a floating-point number following a Gaussian distribution with mean of $\mu$ and standard deviation of $\sigma$. The prompt is \texttt{Sample a number from a Gaussian distribution with mean of $\mu$ and standard deviation of $\sigma$ and put it within \textbackslash{}\textbackslash{}boxed\{\}}. Similar to mathematical reasoning evals, our instruction on putting the generated sample within a box facilitates parsing. We take $N=1024$ in our experiments, which in our examples is typically large enough to provide accurate empirical distribution estimates.\looseness=-1

\textbf{Distributional bias.} 
The sampling results across various distributions and models are visualized in Figure~\ref{fig:sample}. 
We can observe that 
the generated samples cannot follow the specified distributions, as clearly the  LLMs prefer to sample certain values more frequently in discrete spaces, or certain ranges in continuous domains. For instance, the LLMs typically prefer some specific numbers, like 7, 42, or some values in the middle part of the sample space. In Appendix~\ref{appendix_reason}, we visualize the chain-of-thought process when sampling from a distribution. It is observed that while LLMs demonstrate a good understanding of the specified target distributions, they fail to correctly sample from them, yielding the \emph{knowing-doing gap}.\looseness=-1


\begin{table}[t]
\vspace{-0.1cm}
\caption{Goodness-of-fit tests using samples generated by LLMs for three representative distributions. We report p-values for all tests, which are almost zero (smaller than 0.05), which indicates LLMs cannot reliably sample from the target distribution.\looseness=-1}
\vspace{-0.2cm}
\begin{center}
\begin{tabular}{l|cccc}
\toprule
 & Qwen3-8B & Qwen3-32B & Gemini-2.5-Pro & Gemini-3.0-Pro \\
\midrule
Uniform $\{0,1,...,9\}$ 
& $<$1e-300 & 1.37e-235 & $<$1e-300 & $<$1e-300 \\

Uniform $[0,1]$ 
& 8.06e-113 & 2.69e-111 & 3.86e-243 & 3.87e-103 \\

Gaussian $\mathcal{N}(0,1)$ 
& 9.53e-122 & 2.02e-104 & 2.62e-199 & 1.81e-42 \\
\bottomrule
\end{tabular}
\end{center}
\vspace{-0.4cm}
\label{gof_test}
\end{table}

\textbf{Goodness-of-fit tests.} In addition to visually comparing the empirical distributions with their expected counterparts, we conduct goodness-of-fit (GoF) tests. Specifically, for the uniform discrete distribution we conduct a Chi-Square test, while for the continuous uniform and Gaussian distributions we conduct Kolmogorov-Smirnov tests. We report p-values of these GoF tests in Table~\ref{gof_test}. Typically, a p-value larger than 0.05 indicates, at the 5\% significance level, that there is insufficient evidence to reject the null hypothesis that the empirical distribution follows the target distribution. However, all p-values reported in the table are almost zero. This precisely quantifies the failure of sampling from LLMs, as we have inspected earlier in  Figure~\ref{fig:sample}. Furthermore, we note that such sampling failures cannot be fixed by using larger or more advanced LLMs, such as Qwen3-32B and Gemini-3.0-Pro.\looseness=-1

\begin{wrapfigure}{r}{0.45\linewidth}
\vspace{-0.2cm}
    \centering
    \begin{subfigure}[b]{0.22\textwidth}
        \centering
        \includegraphics[width=\linewidth]{figures/gemini-sample/gemini-2.5-pro_uniform_direction.pdf}
    \end{subfigure}
    \hfill
    \begin{subfigure}[b]{0.22\textwidth}
        \centering
        \includegraphics[width=\linewidth]{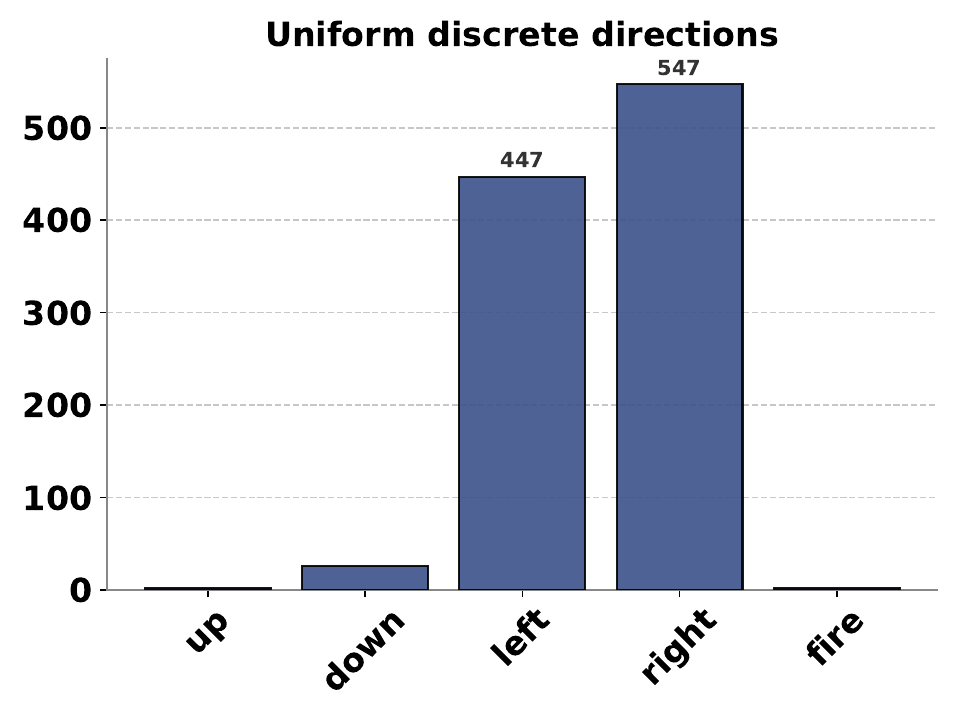}
    \end{subfigure}
    \caption{Empirical distribution estimates, for different orders of random set in the prompt, (\emph{\textbf{Left}}) \texttt{\{left, right, up, down, fire\}}(\emph{\textbf{Right}}) \texttt{\{up, down, left, right, fire\}}.}
    \label{fig:order}
    \vspace{-0.4cm}
\end{wrapfigure}

\textbf{Positional bias.} While the observed biases can sometimes be intuitive, e.g., the preference to numbers such as  
$42$ or $7$, which probably correlates with high presence of these numbers in the 
training data, this is not always the case. For example, when prompting LLMs to uniformly sample a value from \texttt{\{left, right, up, down, fire\}}, LLMs seem to prefer \texttt{up}. But after shifting the order of such a random set in the prompt to \texttt{\{up, down, left, right, fire\}}, LLMs tend to sample \texttt{left} and \texttt{right}, as illustrated in Figure~\ref{fig:order}. In this case the bias seems to be more positional, based on the order of the values in the prompt.\looseness=-1

\textbf{Probing output logits.} We also conduct output-logit analysis when prompting LLMs to sample from distributions. The models we evaluated, such as Qwen3~\citep{yang2025qwen3} and OLMo~\citep{olmo2025olmo}, often generate long thinking traces before the final answer, and these traces may have already been committed to a particular sample. Thus, probing logits after the thinking trace would measure a distribution conditioned on the model’s prior reasoning, rather than its initial stochastic choice. Therefore, our main experiments focus on sampled outputs, which better reflect deployed behavior. Specifically, we consider sampling from a discrete uniform distribution between 0 and 9, and disable the thinking mode and prompt them to directly output a number. We then probe the logits of the first output token and normalize them into a distribution. As shown in Table~\ref{logits}, reasoning models assign highly biased probabilities to particular outputs, e.g., 3, 4, or 5. This corroborates the empirical observations on samples drawn from LLMs' output. Additionally, we also probe the logits of the corresponding base (pre-trained) models, which turns out to be much closer to uniform. This suggests that post-training may substantially amplify the illusion of stochasticity.\looseness=-1

\begin{table}[t]
\vspace{-0.3cm}
\centering
\caption{Output logit distributions (\%) over numbers 0--9 for reasoning and base models.}
\label{tab:digit_distribution}
\resizebox{\linewidth}{!}{
\begin{tabular}{lrrrrrrrrrr}
\toprule
Model & 0 & 1 & 2 & 3 & 4 & 5 & 6 & 7 & 8 & 9 \\
\midrule
Qwen3-8B
& $<0.01$ & $<0.01$ & 0.03 & 1.56 & 66.16 & 31.25 & 0.94 & 0.06 & $<0.01$ & $<0.01$ \\
Qwen3-8B-Base
& 8.21 & 10.54 & 13.53 & 13.53 & 10.54 & 13.53 & 8.21 & 10.54 & 6.39 & 4.98 \\
\midrule
OLMo3-7B-Think
& 0.56 & 0.02 & 1.05 & 69.39 & 4.44 & 8.29 & 0.07 & 15.48 & 0.47 & 0.22 \\
OLMo3-7B-Base
& 10.28 & 5.86 & 8.01 & 13.20 & 13.62 & 12.40 & 9.96 & 11.65 & 9.36 & 5.68 \\
\bottomrule
\end{tabular}
}
\vspace{-0.3cm}
\label{logits}
\end{table}

\vspace{-0.1cm}
\subsection{Effect of decoding parameters}
\vspace{-0.1cm}

Since how LLMs sample \emph{tokens} from a vocabulary depends on the choices of decoding parameters, such as temperature, top-p, top-k, we expect how LLMs act stochastically from a policy can also be impacted.

\begin{figure}[t]
    \centering
    \begin{subfigure}[b]{0.24\textwidth}
        \centering
        \includegraphics[width=\linewidth]{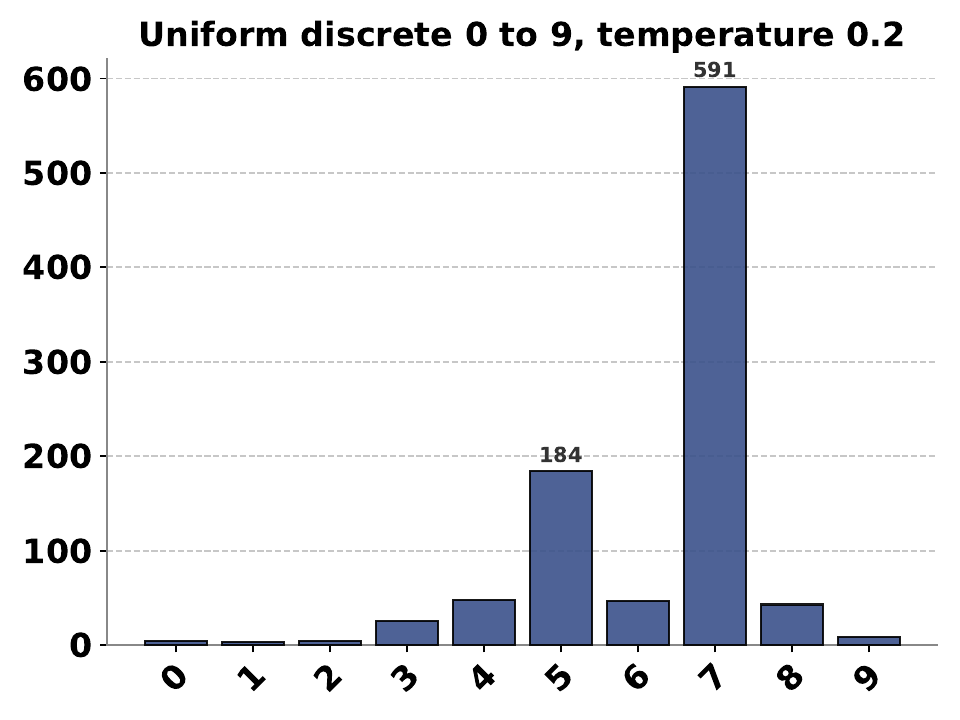}
    \end{subfigure}
    \hfill
    \begin{subfigure}[b]{0.24\textwidth}
        \centering
        \includegraphics[width=\linewidth]{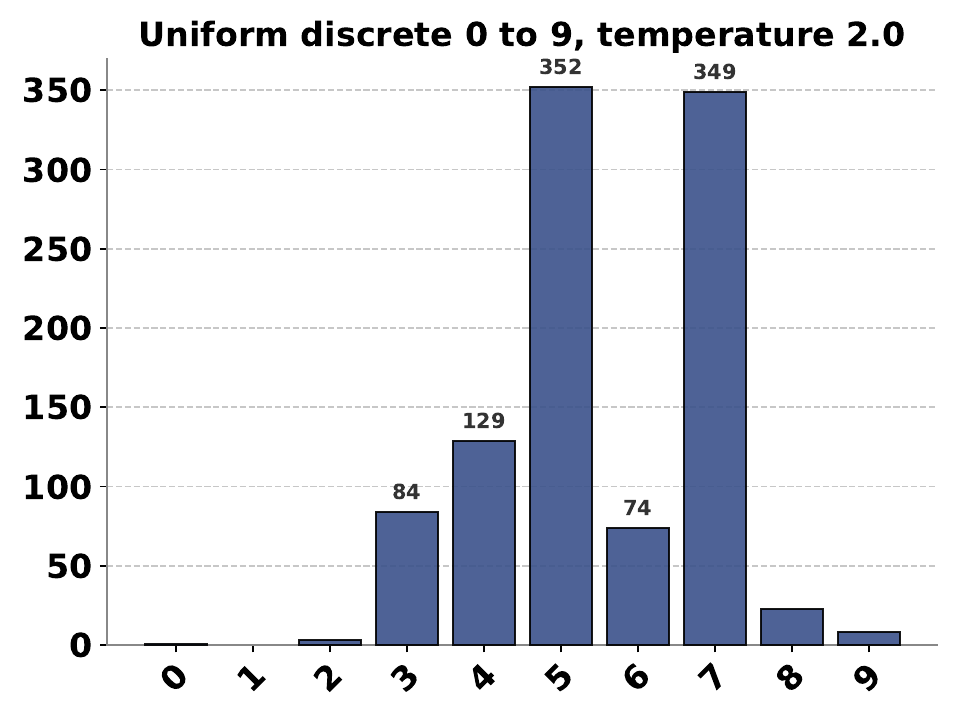}
    \end{subfigure}
    \begin{subfigure}[b]{0.24\textwidth}
        \centering
        \includegraphics[width=\linewidth]{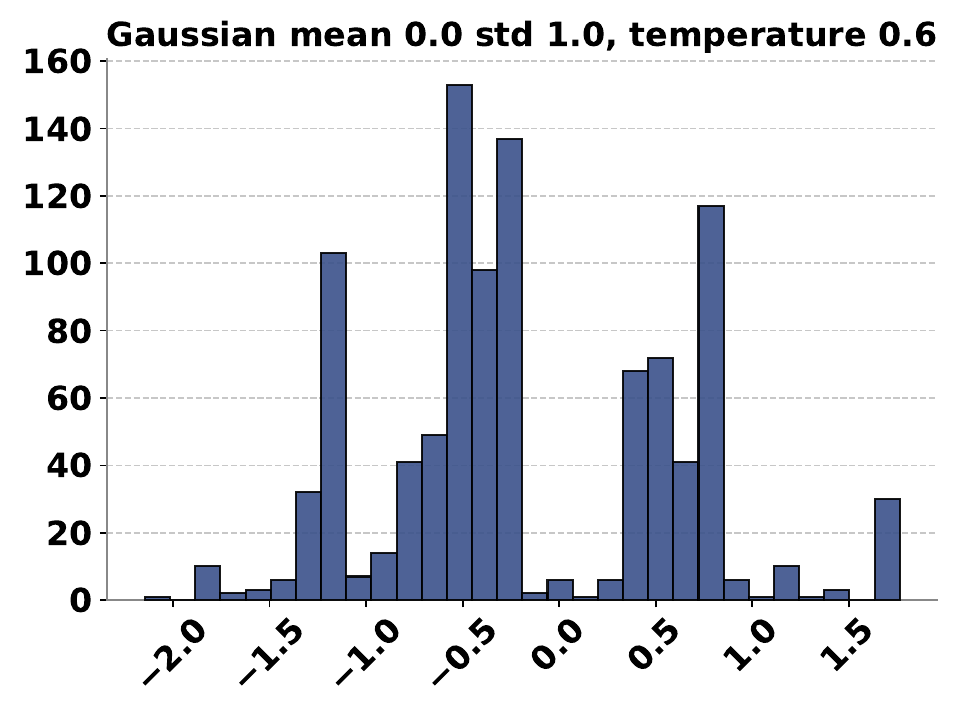}
    \end{subfigure}
    \hfill
    \begin{subfigure}[b]{0.24\textwidth}
        \centering
        \includegraphics[width=\linewidth]{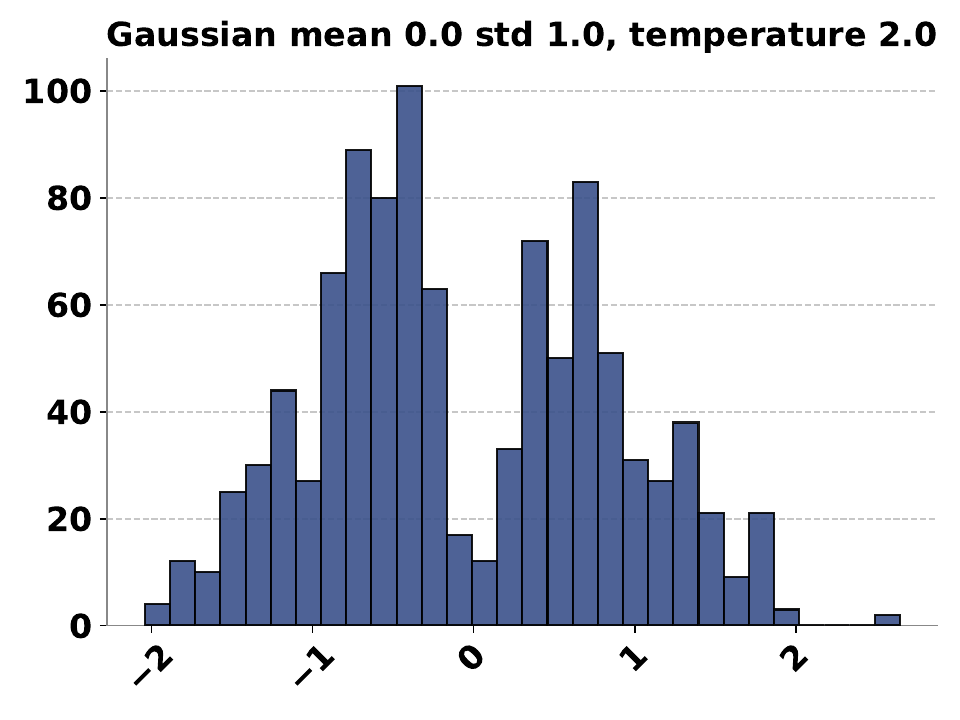}
    \end{subfigure}
    \caption{Empirical distribution estimates, using (\emph{\textbf{Left}}) Qwen3-8B to sample from a uniform discrete distribution with temperature value 0.2 and 2.0; (\emph{\textbf{Right}}) Qwen3-14B to sample from a Gaussian distribution with temperature value 0.6 and 2.0. Across different temperatures, LLMs tend to have similar failure mode, e.g., the preference to 7 and 5, or ranges around $\pm\sigma/2$ instead of $\mu$.}
    \vspace{-0.4cm}
    \label{fig:decode}
\end{figure}

\textbf{Ablating decoding choices.} In our previous experiments, we follow the best recommendations. For example, with Qwen3 family, we use the temperature $T=0.6$, $p=0.95$ for top-p and $k=20$ for top-k. To investigate whether decoding parameters can correct the sampling failure, we ablate their choices within reasonable ranges: temperature is selected from \{0.0, 0.2, 0.6, 1.0, 1.4, 1.8, 2.0\}, top-p \{1.0, 0.99, 0.96, 0.9, 0.8, 0.5, 0.2, 0.1\} (1.0 means no top-p sampling), and top-k \{20, 10, 100, 1000, 10000, -1\} (-1 means no top-k sampling). We only vary one parameter and maintain the other parameters as recommended. For the case of uniform distribution, we find that the p-values consistently remain close to zero across different setups. While for Gaussian distribution, the largest p-value we get is 7.02e-8 (with temperature value 1.8), which is still significantly small. We also visualize the sampling results under different temperatures in Figure~\ref{fig:decode}, and find that LLMs keep falling into similar failure modes.\looseness=-1

\textbf{Extremely large temperatures.} In Figure~\ref{fig:decode}, we observe that the empirical estimates become closer to the target when using large temperatures. Therefore, we further increase the temperature following the range \{2.5, 3.0, 4.0, 5.0, 10.0\}. However, the p-values still remain  significantly less than 0.05. Meanwhile LLMs suffer a lot from parsing errors as they fail to put the numbers inside the box and sample more numbers with extremely large absolute values.\looseness=-1

\subsection{Effects of chain-of-thought reasoning}

Previous works~\citep[e.g.,][]{chen2024not, sui2025stop, gema2025inverse, ghosal2025does} demonstrate that LLMs may overthink 
when using long chain-of-thought reasoning. Therefore, we disable the thinking mode and prompt LLMs to directly output the box with the generated sample.  We find that LLMs still fail to reliably sample from a distribution. For Qwen3, such biases are  amplified as, for instance, Qwen3  generates only 42 when uniformly sampling from 1 to 100. The detailed experimental setups and results are shown in Appendix~\ref{appendix_cot}.\looseness=-1

\begin{abox} 
    \looseness -1 \textbf{Takeaways:} LLMs cannot reliably sample from a distribution, showing distributional or positional bias. Such failures cannot be fixed by modifying decoding parameters or chain-of-thought.
\end{abox}


\begin{figure*}[t]
\vspace{-0.2cm}
    \centering
    \begin{subfigure}[b]{0.24\textwidth}
        \centering
        \includegraphics[width=\linewidth]{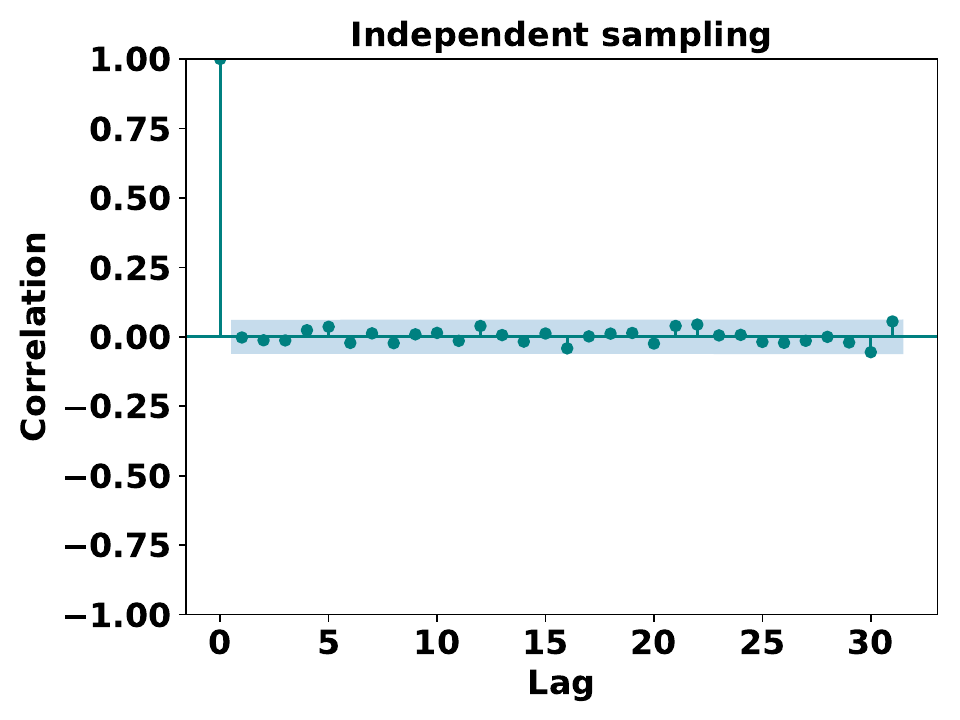}
    \end{subfigure}
    \hfill
    \begin{subfigure}[b]{0.24\textwidth}
        \centering
        \includegraphics[width=\linewidth]{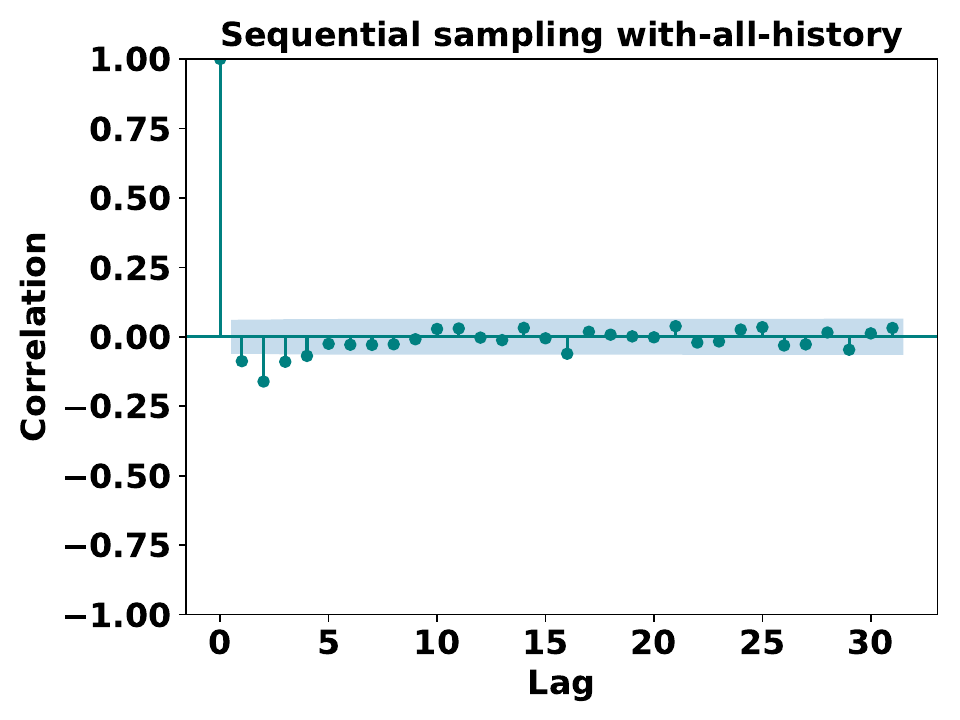}
    \end{subfigure}
    \hfill
    \begin{subfigure}[b]{0.24\textwidth}
        \centering
        \includegraphics[width=\linewidth]{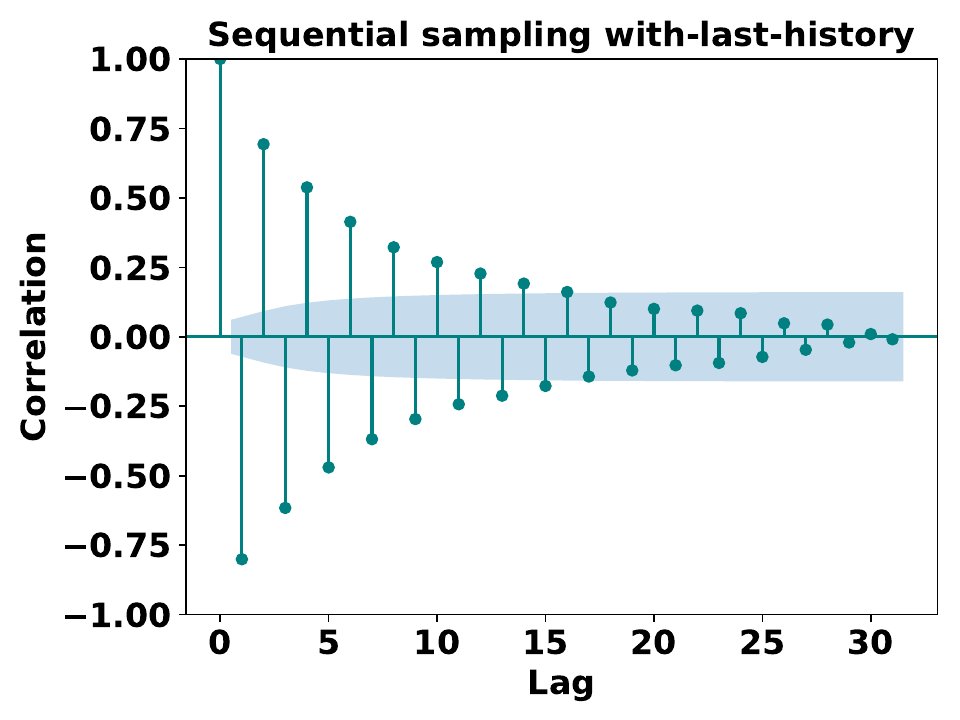}
    \end{subfigure}
    \hfill
    \begin{subfigure}[b]{0.24\textwidth}
        \centering
\includegraphics[width=\linewidth]{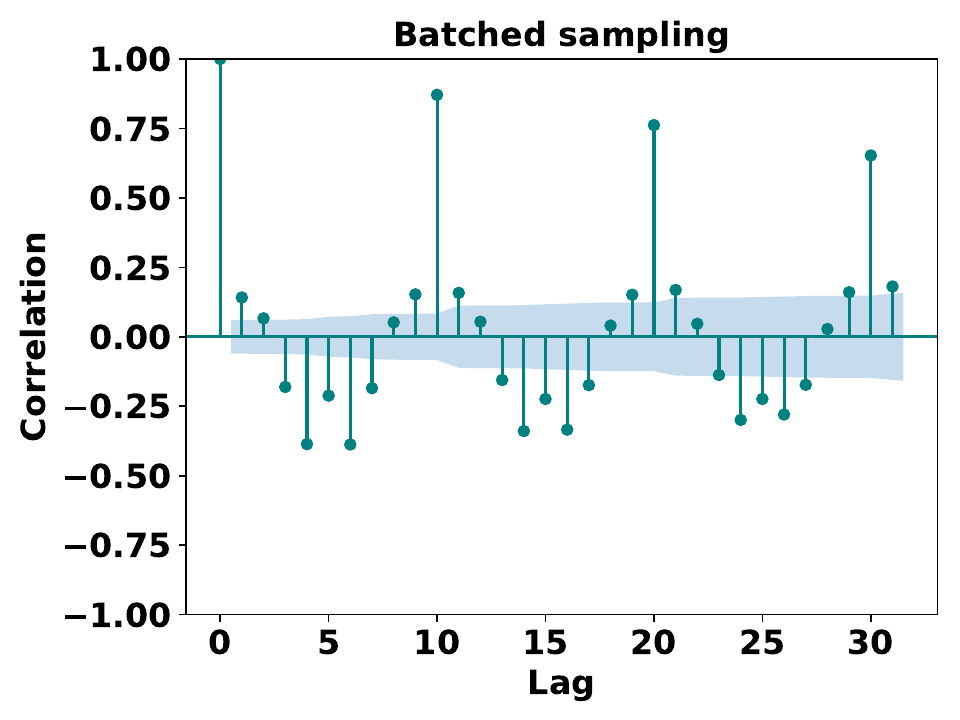}
    \end{subfigure}
    \vspace{-0.2cm}
    \caption{From \emph{\textbf{left}} to \emph{\textbf{right}}, we showcase the auto-correlation of the sample sequence, against different time lags, for four sampling approaches when sampling from a uniform distribution between 0 and 9 with Gemini-2.5-Pro: independent sampling, sequential sampling with-all-history, sequential sampling with-last-history, batched sampling.\looseness=-1}
    \vspace{-0.4cm}
    \label{fig:acf}
\end{figure*}

\vspace{-0.1cm}
\section{Sequential and Batched LLM Sampling}
\vspace{-0.1cm}


\subsection{Sequential sampling with history}

Pseudo-random number generator (PRNG) algorithms are typically stateful, and rely on previous random state. Inspired by this,  we also consider sampling from a distribution in a sequential manner by placing previous generated samples in the context when performing the current sampling step. We use two variants: (i) sequential sampling with-all-history using the extra instruction \texttt{Note that previously you sampled \{ALL\_SAMPLES\}}; sequential sampling with-last-history using the extra instruction \texttt{Note that previously you sampled LAST\_SAMPLE}. In Figure~\ref{fig:seq} (in Appendix) we provide empirical distribution estimates obtained by these sequential approaches and compare them against independent sampling. We find that sequential sampling with-all-history indeed can provide reasonably accurate sampling from a uniform distribution (though LLMs struggle with Gaussian distributions shown in Figure~\ref{fig:seq2} in Appendix), while in contrast last history based sampling shows strong biases.\looseness=-1

\begin{wrapfigure}{r}{0.45\linewidth}
\vspace{-0.1cm}
    \centering
    \begin{subfigure}[b]{0.22\textwidth}
        \centering
        \includegraphics[width=\linewidth]{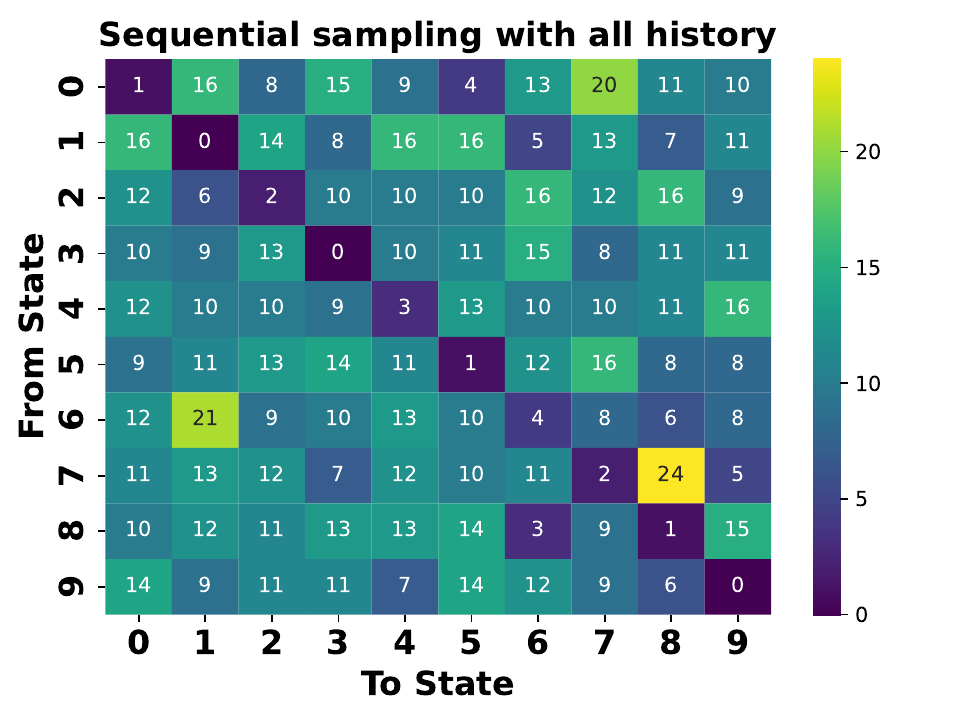}
    \end{subfigure}
    \hfill
    \begin{subfigure}[b]{0.22\textwidth}
        \centering
        \includegraphics[width=\linewidth]{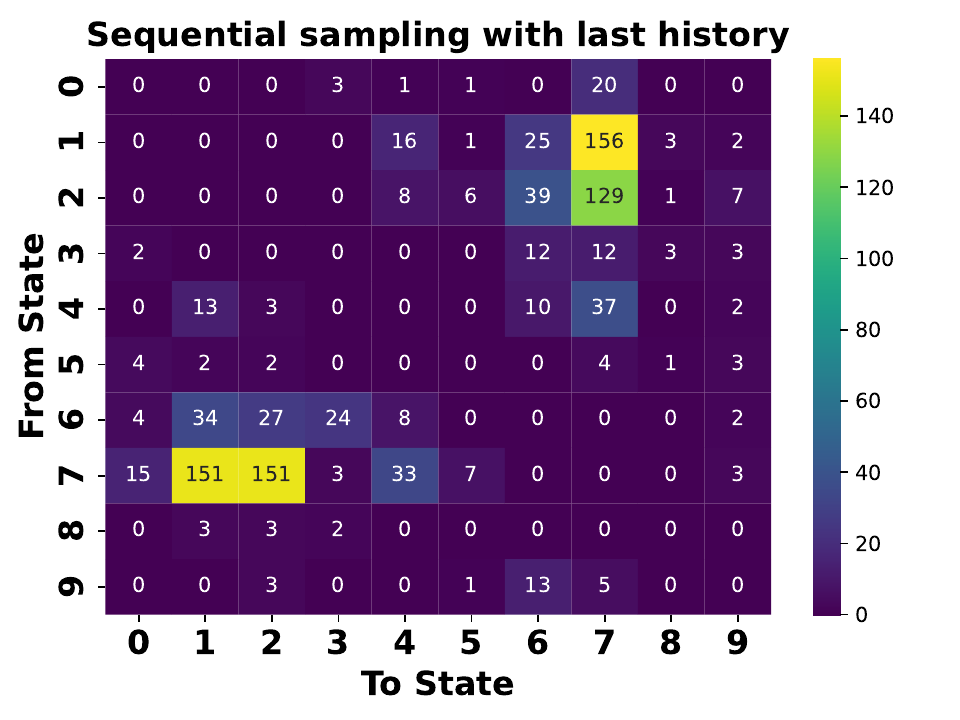}
    \end{subfigure}
    \vspace{-0.2cm}
    \caption{Empirical transition counts between pairs of states when sampling from a uniform distribution between 0 and 9 with Gemini-2.5-Pro. (\emph{\textbf{Left}}) sequential sampling with-all-history (\emph{\textbf{Right}}) sequential sampling with-last-history.}
    \vspace{-0.4cm}
    \label{fig:transition}
\end{wrapfigure}

\textbf{Temporal bias.} With the introduction of dependence on the history, we can also investigate the temporal biases when sampling from LLMs. 
Specifically, we estimate auto-correlation from the sample sequence, as well as transition matrices (for simplicity shown as counts) between discrete states. 
In Figure~\ref{fig:acf}, we visualize the auto-correlation function (ACF) for the two sequential approaches where we show the empirical correlation estimates against different time lags. For comparison, 
we also include the ACF plot for independent sampling. From this figure we observe 
that sequential sampling 
with-all-history exhibits small auto-correlation that resembles reasonably well parallel sampling. However, it still 
exhibits some noticeable 
biases. For example, by observing the negative correlation coefficients 
for the first few lags (from one up to ten), we can conclude 
that the LLM tends to be repulsive, i.e., it prefers to jump away from the current state/sample. 
Regarding the with-last-history 
sequential approach, this method  is significantly worse since 
the samples show strong auto-correlation, having additionally a periodic pattern where the correlation coefficient 
flips sign at each lag. 
To further investigate these behaviours, in Figure~\ref{fig:transition}
we display the empirical transition counts between all pairs of states for the sequential sampling methods. Again we observe that while sampling with-all-history can marginally approximate better the uniform distribution, it suffers from the repulsive bias (also observed by the ACF plot) where the LLM tends to sample a new state away from the current one. Furthermore, the transition counts of with-last-history sampling show a more complex behavior where the LLM tends to jump between two main modes, i.e., a first mode associated with states 1 and 2 and a second one associated with state 7.\looseness=-1

Finally, we point out that besides the  possible temporal biases, a computational limitation of sequential sampling is that the LLMs need to run multiple sequential inference steps which cannot be easily parallelized. Therefore, these methods can be extremely slow in practice. 
\looseness=-1

\subsection{Sampling with batched generation} 

Another sampling approach is to prompt LLMs to directly generate $N$ values from a distribution. Therefore, our prompt is revised as \texttt{Sample 1024 values from a uniform distribution on the set <RANDOM\_SET>, put them within \textbackslash{}\textbackslash{}boxed\{\} and use commas to separate values.} Firstly, one notable problem for batched sampling is that LLMs struggle to faithfully generate the same number of random numbers as instructed in the prompt, shown in Figure~\ref{fig:seq2} (in Appendix). We believe this can be attributed to the inability of LLMs to count numbers~\citep{barbero2024transformers}. Secondly, with batched sampling, LLMs indeed can sometimes provide reasonably accurate empirical distribution estimates for uniform distributions, though not perform well on Gaussians as shown by Figure~\ref{fig:seq2} (in Appendix). However, we still notice some failure cases, which demonstrate very strong temporal biases. As shown in Figure~\ref{fig:acf}(\emph{\textbf{right}}), the generated samples show  periodic patterns, e.g., the generated values are repeated every 10 samples. This pattern is undesirable for a sequence of random numbers and can be easily exploited. Finally, after running batched sampling multiple times, we empirically find that the first/second value in each batch across different runs does not follow the target distribution, as shown in Figure~\ref{fig:batch1} (in Appendix). Especially, for Gaussian distributions, batched sampling always starts with numbers within a very small range.\looseness=-1

\begin{abox} 
    \looseness -1 \textbf{Takeaways:} Sequential sampling (with-all-history) and batched sampling can provide more reasonably accurate sampling from uniform distributions but fail again on Gaussian distributions. These sampling approaches meanwhile introduce undesirable temporal biases.
\end{abox}

%% file: sections/methodology.tex
\begin{figure*}[t]
\vspace{-0.2cm}
    \centering
    \begin{subfigure}[b]{0.48\textwidth}
        \centering
        \includegraphics[width=\linewidth]{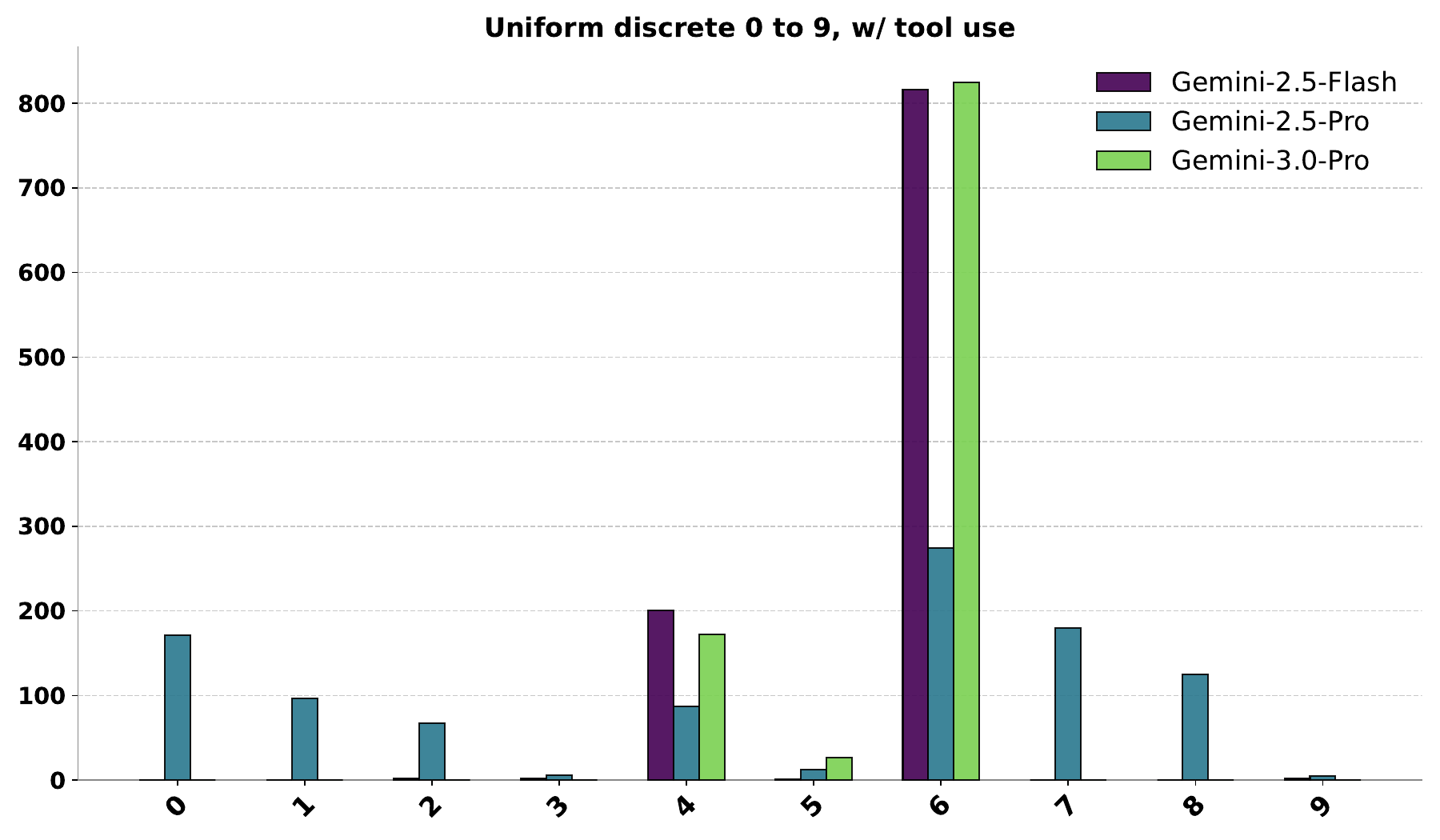}
    \end{subfigure}
    \hfill
    \begin{subfigure}[b]{0.48\textwidth}
        \centering
        \includegraphics[width=\linewidth]{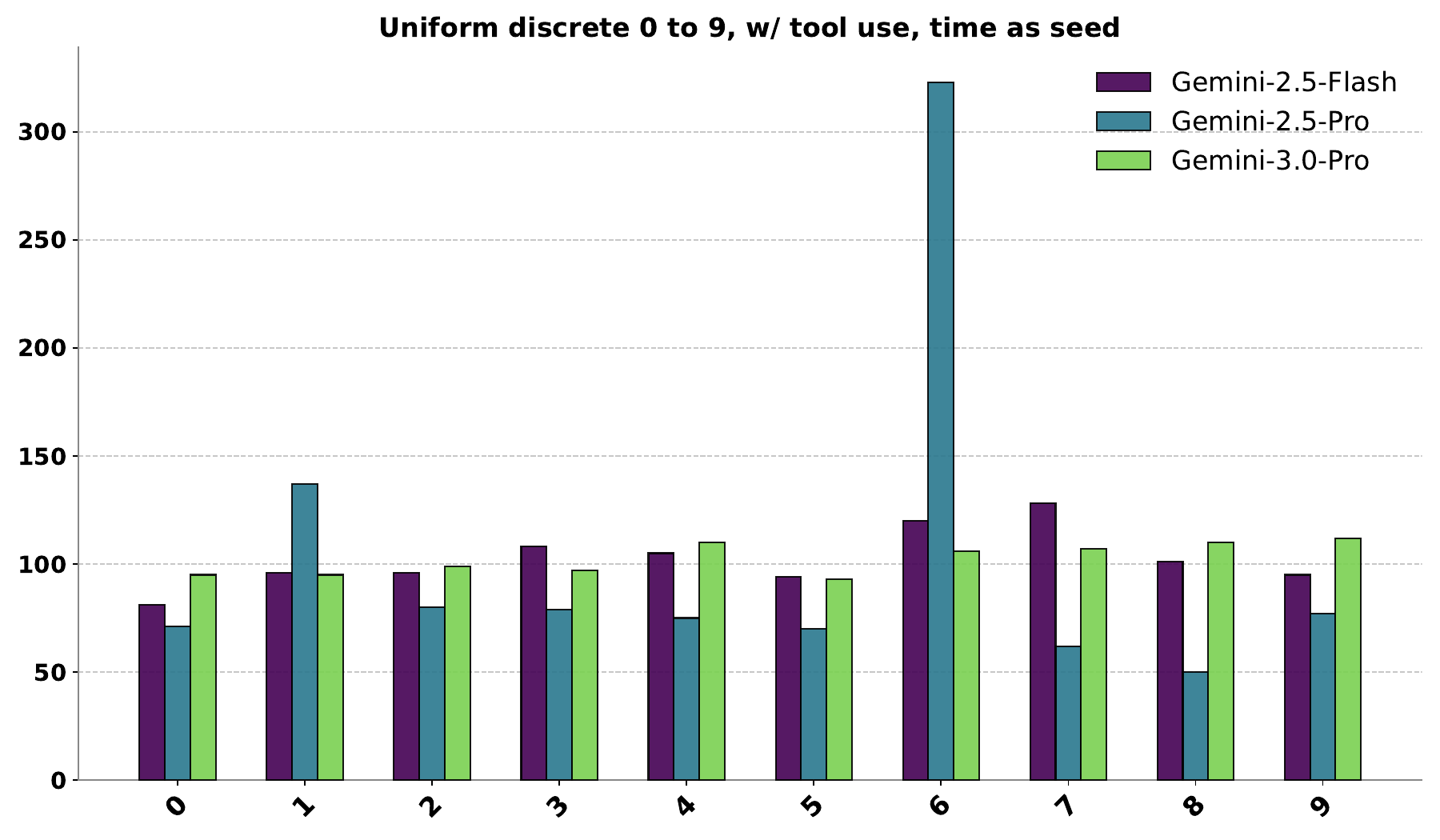}
    \end{subfigure}
    \vspace{-0.2cm}
    \caption{Empirical distribution estimates, for a uniform discrete distribution between 0 and 9, using the Gemini family under the setup of (\emph{\textbf{Left}}) tool-use (\emph{\textbf{Right}}) tool-use, prompting to use current time as random seed.}
    \vspace{-0.4cm}
    \label{tool-use2}
\end{figure*}

\vspace{-0.1cm}
\section{LLM Sampling with PRNG}
\vspace{-0.1cm}

\subsection{Tool-using PRNG algorithm with LLMs}

In the previous results, we relied on LLMs' own sampling abilities without any tool calling. However,
most modern programming languages provide easy access to complex pseudo-random number generator (PRNG) algorithms. Next, we enable LLMs (e.g., Gemini family) to generate \emph{Python} code and provide the execution sandbox.  We add an extra instruction to the original prompt: \texttt{You can generate and run code for sampling}. Consequently, LLMs can generate a similar snippet of code across independent sampling runs and always includes the use of the \emph{random.randint()} function. We observe that the generated code typically contains no random seed, which can invalidate the stochastic behavior as all samples will be seeded from a small set of values. 

\begin{figure*}[t]
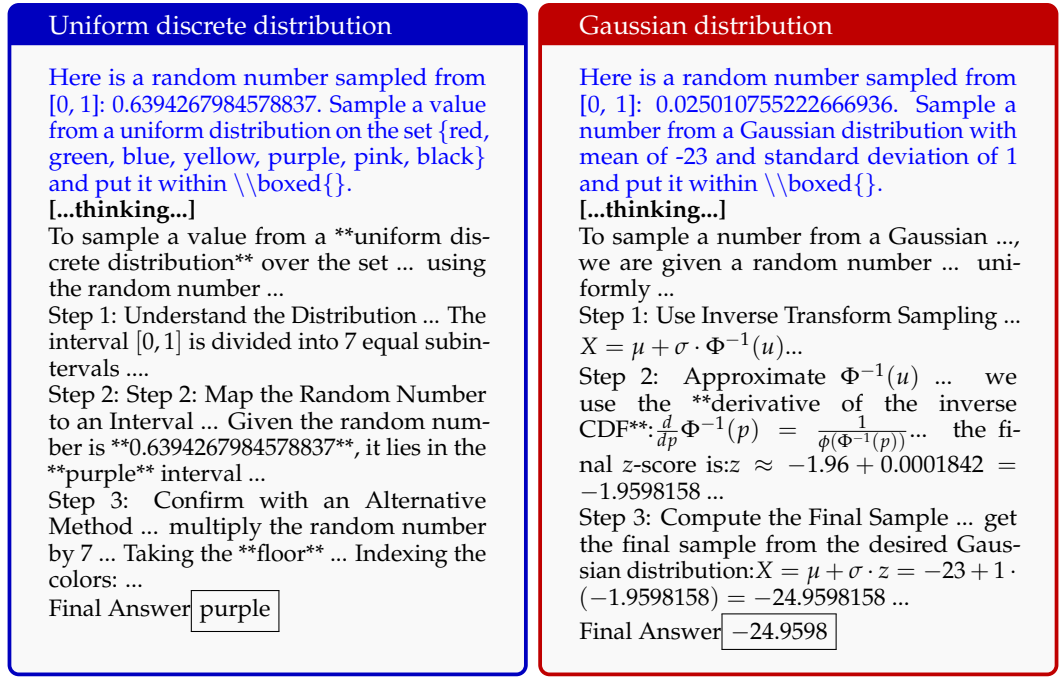

\vspace{-0.2cm}
\centering
\begin{minipage}[t]{0.495\textwidth}
\begin{tcolorbox}[
  width=\textwidth,
  colback=gray!5,
  colframe=blue!75!black,
  equal height group=A, 
  title=Uniform discrete distribution
]
\small{
\textcolor{blue}{Here is a random number sampled from [0, 1]: 0.6394267984578837. Sample a value from a uniform distribution on the set \{red, green, blue, yellow, purple, pink, black\} and put it within \textbackslash{}\textbackslash{}boxed\{\}.} \\
\textbf{[...thinking...]}\\
To sample a value from a **uniform discrete distribution** over the set ... using the random number ...\\ 
Step 1: Understand the Distribution ... The interval $[0, 1]$ is divided into 7 equal subintervals ....\\
Step 2: Step 2: Map the Random Number to an Interval ... Given the random number is **0.6394267984578837**, it lies in the **purple** interval ... \\
Step 3: Confirm with an Alternative Method ... multiply the random number by 7 ... Taking the **floor** ... Indexing the colors: ...\\
Final Answer$\boxed{\text{purple}}$\
}
\end{tcolorbox}
\end{minipage}
\hfill 
\begin{minipage}[t]{0.495\textwidth}
\begin{tcolorbox}[
  width=\textwidth,
  colback=gray!5,
  colframe=red!75!black,
  equal height group=A, 
  title=Gaussian distribution
]
\small{
\textcolor{blue}{Here is a random number sampled from [0, 1]: 0.025010755222666936. Sample a number from a Gaussian distribution with mean  of -23 and standard deviation of 1 and put it within \textbackslash{}\textbackslash{}boxed\{\}.} \\
\textbf{[...thinking...]}\\
To sample a number from a Gaussian ..., we are given a random number ... uniformly ...\\
Step 1: Use Inverse Transform Sampling ... $X = \mu + \sigma \cdot \Phi^{-1}(u)$...\\
Step 2: Approximate $ \Phi^{-1}(u) $ ... we use the **derivative of the inverse CDF**:$\frac{d}{dp} \Phi^{-1}(p) = \frac{1}{\phi(\Phi^{-1}(p))}$... the final $ z $-score is:$z \approx -1.96 + 0.0001842 = -1.9598158$ ...\\
Step 3: Compute the Final Sample ... get the final sample from the desired Gaussian distribution:$X = \mu + \sigma \cdot z = -23 + 1 \cdot (-1.9598158) = -24.9598158$ ... \\
Final Answer$\boxed{-24.9598}$
}
\end{tcolorbox}
\end{minipage}
\caption{Truncated responses when prompting LLMs (Qwen3-32B) to use a random number in [0, 1] to sample from (\emph{\textbf{Left}}) uniform discrete distribution on colors and (\emph{\textbf{Right}}) Gaussian distribution $\mathcal{N}(-23, 1)$. More examples are shown in Appendix~\ref{appendix_convert}.}
\vspace{-0.2cm}
\label{fig:think}
\end{figure*}

Instead we notice a different failure mode. The behavior of the execution sandbox becomes 
deterministic shown in Figure~\ref{tool-use2}(\emph{\textbf{Left}}), potentially due to the fixing of the random seed under the hood by the python environment in each independent execution. Note that there is no persistent state of the interpreter between calls. Therefore, only when we add another instruction: \texttt{Make sure to seed the random number generator to ensure different results on each run, for example using the current time}, LLMs become more reliable when sampling from distributions, as shown in Figure~\ref{tool-use2}(\emph{\textbf{Right}}). However, there is also dependence on the model's instruction following ability. For instance, Gemini-3.0-Pro can follow this instruction well, faithfully generating a line \emph{random.seed(time.time())} and thus performing correct sampling. While Gemini-2.5 may not always generate such a line, resulting in different degrees of failures. Additionally, 
if multiple independent samplings happen almost concurrently, the biases to certain numbers can emerge.\looseness=-1

\begin{table}[t]
\caption{Accuracies of LLMs simulating PRNG algorithms for uniform distributions. For discrete distribution, we consider a sample correct when it is exactly the same as PRNG output. While for continuous one, we consider their difference is smaller than 0.01.\looseness=-1}
\vspace{-0.2cm}
\begin{center}
\begin{tabular}{l|cccc}
\toprule
 & Qwen3-1.7B & Qwen3-4B & Qwen3-8B & Gemini-2.5-Pro \\
\midrule
Uniform $\{0,1,...,9\}$ 
& \textcolor{red}{\textbf{21.3\%}} & 92.6\% & 97.5\% & 94.3\% \\

Uniform $[0,1]$ 
& \textcolor{red}{\textbf{19.9\%}} & 96.2\% & 97.8\% & 92.3\% \\
\bottomrule
\end{tabular}
\end{center}
\vspace{-0.4cm}
\label{accuracy}
\end{table}

\subsection{Simulating PRNG algorithm with LLMs}

We further explore whether LLMs could simulate the deterministic algorithm behind a PRNG. To implement this, we directly provide LLMs with a snippet of PRNG code as well as a random seed in the prompt; see Figure~\ref{prompt_prng} to Figure~\ref{prompt_prng3} in Appendix for examples. Meanwhile, we ensure LLMs cannot use any execution tool/sandbox to run the code. For independent sampling, a different random seed must be provided, e.g., index among $N$ samples plus a constant.\looseness=-1 

\textbf{Simulation results.} We first evaluate the accuracy of LLMs simulating PRNG algorithms for uniform distributions. Specifically, we compare the generated value with the actual one returned from the PRNG algorithm given the same seed. As shown in Table~\ref{accuracy}, we find that LLMs can achieve above 90\% simulation accuracy after certain model size (e.g., 4B for Qwen3 family). Therefore, as expected, the empirical distributions obtained in this way are close to the target ones, as shown in Table~\ref{prng_gof} (in Appendix).\looseness=-1

Further, we find that LLMs can struggle with the PRNG algorithm for Gaussians, which requires sampling two uniform random values in $[0, 1]$ and applying the Box-Muller transform. With detailed investigations in Appendix~\ref{appendix_simulate}, we notice that when LLMs simulate the second random number generation, the state already becomes very large. Then they fail to accurately run the multiplication between two large numbers. This corroborates the inability of LLMs to execute PRNG algorithms shown by \citet{markeeva2024clrs}.\looseness=-1

\textbf{Role of random seed.} Typically, PRNG algorithms are stateful, which are not suitable for independent sampling. The reason why we obtain more ``uniform'' empirical estimates is that we use a different random seed for each independent call. If we prompt LLMs to generate a random seed first instead of explicitly providing one, LLMs show similar stochastic failures by biasing to certain random seeds.\looseness=-1

\begin{abox} 
    \looseness -1 \textbf{Takeaways:} LLMs can employ PRNG algorithms to reliably sample from a distribution through tool-use only when random seeds are properly set. LLMs can simulate PRNG algorithms with the code and a random seed within certain complexity.
\end{abox}

\begin{figure*}[t]
\vspace{-0.2cm}
    \centering
    \begin{subfigure}[b]{0.23\textwidth}
        \centering
        \includegraphics[width=\linewidth]{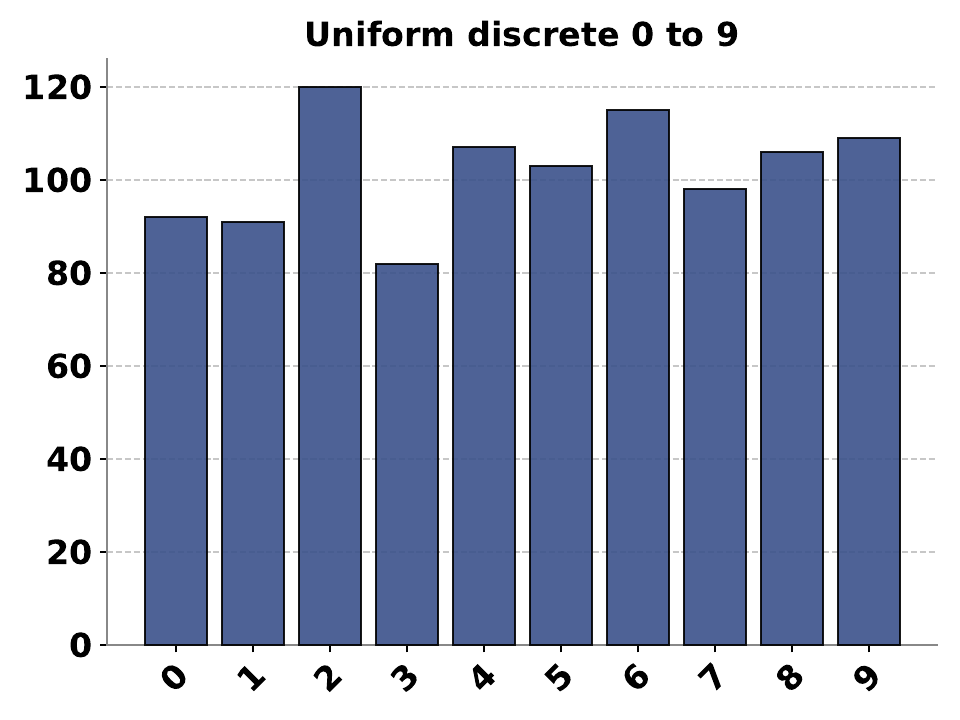}
    \end{subfigure}
    \hfill
    \begin{subfigure}[b]{0.23\textwidth}
        \centering
        \includegraphics[width=\linewidth]{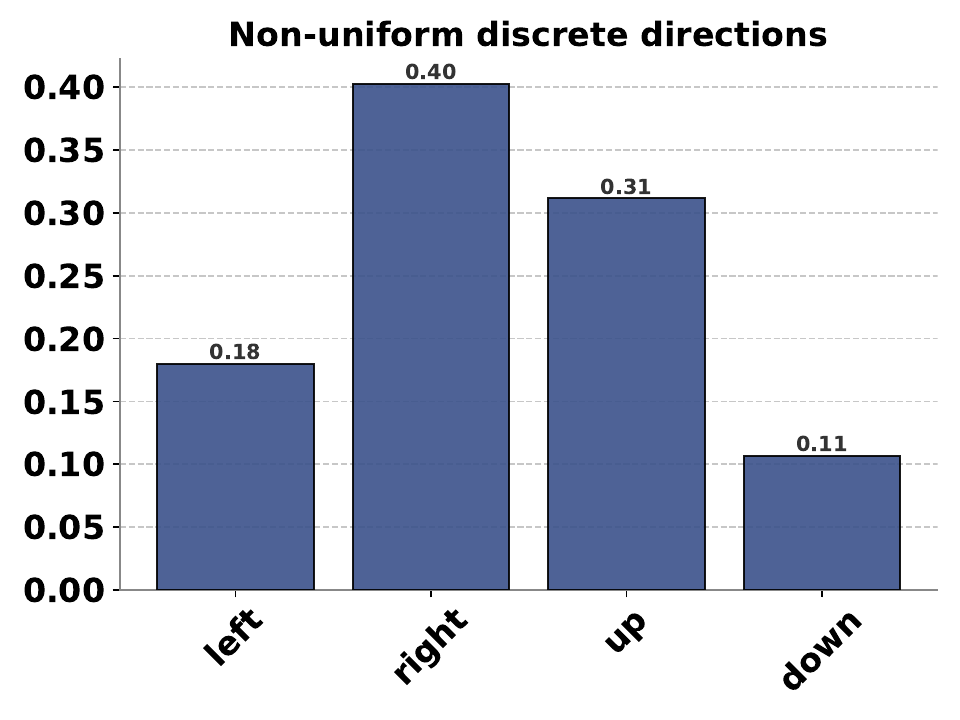}
    \end{subfigure}
    \begin{subfigure}[b]{0.23\textwidth}
        \centering
        \includegraphics[width=\linewidth]{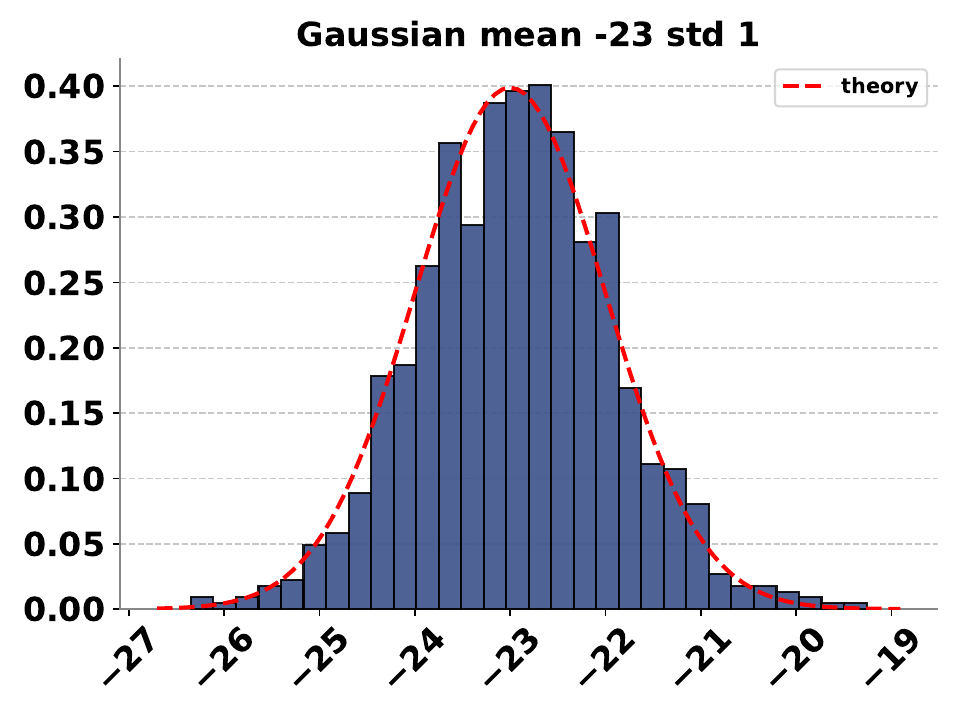}
    \end{subfigure}
    \hfill
    \begin{subfigure}[b]{0.23\textwidth}
        \centering
        \includegraphics[width=\linewidth]{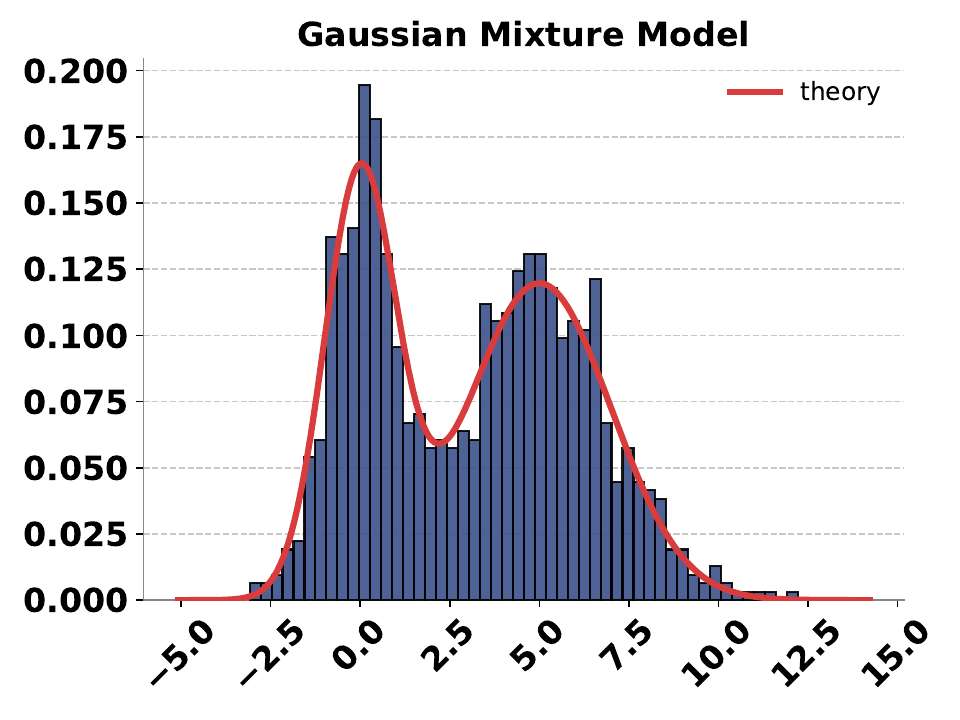}
    \end{subfigure}
    \vspace{-0.2cm}
    \caption{LLMs can reliably convert a uniform distribution in [0, 1] to various distributions. From \emph{\textbf{left}} to \emph{\textbf{right}}, we use Qwen3-30B-A3B to sample from a uniform discrete distribution between 0 and 9; Qwen3-32B to sample from a non-uniform distribution on directions; Qwen3-14B to sample from a Gaussian; Gemini-3.0-Pro to sample from a Gaussian mixture model.\looseness=-1}
    \vspace{-0.2cm}
    \label{fig:convert}
\end{figure*}

\begin{table*}[t]
\caption{Goodness-of-fit tests when prompting Qwen3 family to convert $\mathcal{U}[0, 1]$ to uniform \{0, 1, 2, ..., 9\} and Gaussian $\mathcal{N}(0, 1)$. We \textbf{bold} the scenarios that achieve $p>0.05$, which show the correlation with model size.\looseness=-1} 
\vspace{-0.2cm}
\begin{center}
\begin{tabular}{l|llllll}
\toprule
Qwen3 model size & 0.6B & 1.7B & 4B & 8B & 14B & 32B \\
\midrule
Uniform \{0,1,...,9\} & 0.00\,\,\,\,\,\,\,\,\,\,\,\,\, & 2.33e-4\,\,\,\,\,\, & \textbf{0.23}\,\,\,\,\,\,\,\,\,\,\, & \textbf{0.23} & \textbf{0.23} & \textbf{0.23} \\
Gaussian $\mathcal{N}(0, 1)$ & 1.99e-232 & 9.50e-186 & 1.61e-46 & \textbf{0.29} & \textbf{0.21} & \textbf{0.78}\\
\bottomrule
\end{tabular}
\end{center}
\vspace{-0.4cm}
\label{emergent}
\end{table*}

\vskip -0.1cm
\section{Distribution Conversion with LLMs}
\vskip -0.1cm

We further explore LLMs' capabilities to execute deterministic algorithms to convert a sequence of numbers to the target distribution even \textit{without providing the algorithm}. Therefore, instead of providing an integer random seed, we provide a random number sampled from uniform [0, 1] to ``seed'' the conversion. For uniform discrete distribution, the prompt is designed: \texttt{Here is a random number sampled from [0, 1]: <RANDOM\_NUMBER>. Sample a value from a uniform distribution from the set <RANDOM\_SET> and put it within \textbackslash{}\textbackslash{}boxed\{\}}. Then for each prompt, the RANDOM\_NUMBER is independently sampled from uniform [0, 1] with \emph{Python}. We adopt a similar design for sampling from a Gaussian distribution.\looseness=-1

\textbf{Success of conversion.} Empirically, we find with the above prompt design, LLMs can reliably sample from uniform discrete distributions and Gaussian distributions, as visualized in Figure~\ref{fig:convert}. Specifically, they can sample from uniform discrete distributions with arbitrary labels, including numbers and texts. The sampled results from Gaussian distributions with different means and standard deviations can also fit well with the theoretical curves. To further investigate how LLMs complete such a distribution conversion, we display the model responses in Figure~\ref{fig:think}. It is observed that LLMs run a deterministic algorithm to convert distributions. For uniform discrete distribution, LLMs use the bucketization algorithm while for Gaussian distribution, they adopt an inverse transform sampling approach.\looseness=-1

\textbf{Correlation with model size.} As the above success requires LLMs to figure out the deterministic transformation through chain-of-thought reasoning, we investigate the role of model size. We conduct goodness-of-fit tests across sampling results generated by Qwen3 with varying model sizes. As shown in Table~\ref{emergent}, we observe clear correlation with model size. Typically, Gaussian distributions require model size larger than 4B while uniform discrete distributions require model size only larger than 1.7B to achieve $p>0.05$.\looseness=-1


\textbf{Converting to advanced distributions.} Beyond the uniform and Gaussian distributions, we also consider the following advanced distributions: (i) non-uniform discrete distribution on directions: \{left: 0.2, right: 0.4, up: 0.3, down: 0.1\}; (ii) a two-component Gaussian mixture model (GMM) where the first component has mean 0.0, standard deviation 1.0, weight 0.4, and the second component has mean 5.0, standard deviation 2.0, weight 0.6. As shown in Figure~\ref{fig:convert}, we find that LLMs can also convert distributions to those more complex distributions. While without such a distribution conversion, for instance, LLMs over-sample ``left'' as it has the highest probability. 

\begin{abox} 
    \looseness -1 \textbf{Takeaways:} LLMs can effectively convert samples between certain standard distributions (a deterministic process), but may fail as transformation complexity increases.\looseness=-1
\end{abox}





%% file: sections/discussion.tex
\vskip -0.2cm
\section{Discussion}
\vskip -0.2cm




In this work we provide thorough empirical evidence of the inability of LLMs to 
reliably sample from various simple distributions, both discrete and continuous. We include a detailed discussion on related work in Appendix~\ref{literature}. The observed biases can sometimes be intuitive, e.g., the preference towards sampling number  $42$, which likely stems from the higher presence of these numbers in the 
training data. Sometimes the bias seems to be more positional.\looseness=-1

On the other hand, when using tools, or even chain of thought to emulate the steps 
of pseudo-random number generator algorithms, sufficiently powerful frontier
models can, to a good degree, generate random numbers. However relying on 
pseudo-random number generators is expensive in terms of inference compute, and prohibitive for tasks requiring repeated sequential sampling. Additionally pseudo-random number generators require tracking a state in order to provide meaningful samples (which 
is usually given in the form of the seed). Therefore to generate meaningful random numbers through emulation, the model needs not only to emulate the computational steps, but 
track the value of the state from consecutive calls. The same is partially true when using tools as well. If every sample requires generating a python code and a fresh execution of 
the code, one needs to ensure that the generated code does not specify a random seed (as is common practice), since then the model will need to sample the seeds from a uniform distribution for each call. Without a specified seed, we notice that LLMs do not use the system time to explicitly seed the process, unless instructed, and can be vulnerable to seeds being fixed somewhere else in the pipeline. Even when instructed, we find that only the more powerful models are able to properly follow the instructions.\looseness=-1

Given the cost of simulating the random number generator or conversion between distribution, we argue that these approaches are problematic for an agentic system that needs to sample frequently and reliably. We hypothesize that a pragmatic solution is to provide the LLM with access to a stateful sampler, that tracks its state between calls. We note that the concept of a stateful tool might be of interest for other processes as well, beyond sampling. Additionally, \citet{von2024learning,gillman2024fourier} showed that randomness in Transformers can be learnt through specific training, which may be challenging for current LLMs. This however might offer another avenue for approaching the sampling problem.\looseness=-1










 
 

%% file: sections/literature.tex
\section{Related Work}\label{literature}
\subsection{Sampling from distributions with LLMs}

\citet{liu2023does,hopkins2023can,van2024random,harrison2024comparison} represent the initial explorations showing that with independent sampling, LLMs struggle with generating random numbers following simple probabilistic distributions, e.g., a categorical distribution by flipping coins or rolling a dice, or a uniform distribution between [0, 1]. More recently, \citet{gu2025llms} extended the distributions evaluated to broader families. They showed that LLMs can understand probabilities yet struggle with sampling, but they focused on batched sampling. \citet{gupta2025enough} also demonstrated that LLMs have randomness biases when independently sampling from the Bernoulli distribution. With a sequence of actual generations in the context, LLMs become more faithful in sampling the target distribution. But they didn't investigate the temporal biases in such a sequential sampling approach. \citet{xiao2025flipping} introduced verbalized rejection sampling to improve the performance when using LLMs to sample from the Bernoulli distribution. They formulate a classical rejection sampling method in the prompt. Meanwhile, a random number from another distribution is required as input for LLMs to reject or accept.  Besides the random number generation, \citet{vidler2025playing,guo2025illusion} also explored whether LLMs can play games that require randomness, e.g., Rock-Paper-Scissors and Prisoner's Dilemma. They found that LLMs cannot act stochastically even though they are prompted to be random.\looseness=-1

The above explorations mainly focus on non-thinking models, e.g., ChatGPT-3.5~\citep{openai2022chatgpt}, GPT-4~\citep{achiam2023gpt}, LLAMA~\citep{touvron2023llama}, Alpaca~\citep{taori2023stanford}. With the success of reinforcement learning~\citep{schulman2017proximal,shao2024deepseekmath,guo2025deepseek}, Large Reasoning Models (LRMs)~\citep{jaech2024openai,guo2025deepseek, comanici2025gemini} have demonstrated improved reasoning capabilities, e.g., better performance on counting letters in a word. Therefore, we are wondering whether acting stochastically according to a target distribution policy can be solved by these more powerful models or is fundamentally flawed for current LLMs. Our paper features a comprehensive empirical study with different LLM families, distributions, sampling approaches, etc. A concurrent work~\citep{zhao2026large} also explored whether LLMs (including LRMs) can generate random numbers from statistical distributions. They mainly explored independent sampling and batched sampling. Different from \citet{zhao2026large}, we also explored the effects of decoding parameters, chain-of-thought, sequential sampling with history, and showcase the positional bias. Finally, we find that LLMs are much more reliable when converting distributions than directly sampling from them.\looseness=-1

Regarding potential solutions, our proposed remedies are primarily inference-time approaches: external samplers with proper seed/state management and distribution conversion. Although they do not focus specifically on sampling from distributions, \citet{zhang2025verbalized} propose verbalized sampling to improve generation diversity, prompting LLMs to produce a set of candidate responses together with verbalized probabilities. This approach is conceptually related to our exploration of batched sampling. Meanwhile, \citet{hu2024amortizing} use random-number generation as a motivating example and propose fine-tuning LLMs with diversity-seeking reinforcement learning algorithms, particularly generative flow networks (GFlowNets). Similarly, \citet{zhang2024forcing} introduce distribution matching as a language-model fine-tuning objective to encourage more diverse generations. Despite their promising results, it remains unclear whether these post-training remedies can be integrated into a general-purpose post-training pipeline without weakening either their diversity-promoting effects or the model's broader capabilities.

\subsection{Related Concepts}

\textbf{Semantic Uncertainty} methods, such as \citet{kuhn2023semantic,farquhar2024detecting}, highlight that meaning, rather than surface token form, is often the right unit for analyzing LLM outputs. This is closely related to our argument in the main paper: there is rarely a one-to-one mapping between next-token logits and intended semantic actions, e.g., “move left” and “move right” may share the same first token. However, our focus is different. Semantic uncertainty estimates uncertainty over generated meanings, while our work studies whether LLMs can faithfully generate samples from a specified semantic action distribution.

\textbf{LLM Calibration}~\citep{liu2025uncertainty,wang2024calibrating} is also a related concept, but addresses a different problem. Existing calibration work mainly studies whether a model’s confidence in an answer matches its empirical correctness. In contrast, we study whether the model’s generated samples match a specified target distribution. Even when the target distribution is explicitly given and the model appears to understand it through reasoning traces, the realized output distribution can still deviate substantially from the target.

\textbf{Numeracy and Number Processing} in LLMs also have connections with our work. For example, \citet{barbero2024transformers} showed that LLMs struggle to copy a single number at a specified position or to count the frequency of a particular number in the input, especially under long context lengths. Similarly, \citet{markeeva2024clrs} evaluated LLMs on algorithmic tasks such as sorting, searching, and dynamic programming, and found that performance degrades significantly as input size increases. \citet{yang2025number} also exposed vulnerabilities of LLMs in basic numerical understanding and processing, and other studies suggest that tokenization can contribute to these failures~\citep{singh2024tokenization,kreitner2025efficient}. This is relevant to our observation that LRMs struggle to simulate PRNG algorithms when the seed/state becomes extremely large. Tokenization effects may also partly explain why certain numerical samples are preferred.


%% file: appendix/appendix_a.tex
\section{More Results on Independent Sampling}

\subsection{Motivating example of multi-choice question generation}

\begin{figure*}[htbp]
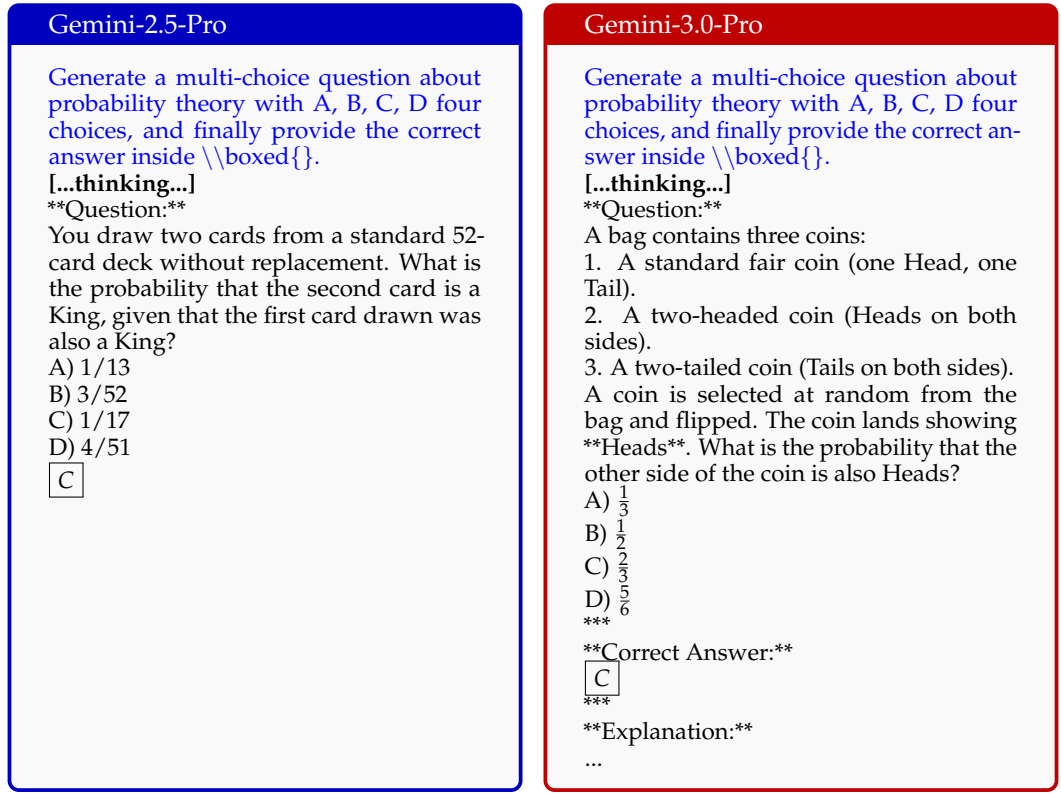

\centering
\begin{minipage}[t]{0.49\textwidth}
\begin{tcolorbox}[
  width=\textwidth,
  colback=gray!5,
  colframe=blue!75!black,
  equal height group=G, 
  title=Gemini-2.5-Pro
]
\small{
\textcolor{blue}{Generate a multi-choice question about probability theory with A, B, C, D four choices, and finally provide the correct answer inside \textbackslash{}\textbackslash{}boxed\{\}.} \\
\textbf{[...thinking...]}\\
**Question:**\\
You draw two cards from a standard 52-card deck without replacement. What is the probability that the second card is a King, given that the first card drawn was also a King?\\
A) 1/13\\
B) 3/52\\
C) 1/17\\
D) 4/51\\
$\boxed{C}$
}
\end{tcolorbox}
\end{minipage}
\hfill 
\begin{minipage}[t]{0.49\textwidth}
\begin{tcolorbox}[
  width=\textwidth,
  colback=gray!5,
  colframe=red!75!black,
  equal height group=G, 
  title=Gemini-3.0-Pro
]
\small{
\textcolor{blue}{Generate a multi-choice question about probability theory with A, B, C, D four choices, and finally provide the correct answer inside \textbackslash{}\textbackslash{}boxed\{\}.} \\
\textbf{[...thinking...]}\\
**Question:**\\
A bag contains three coins:\\
1.  A standard fair coin (one Head, one Tail).\\
2.  A two-headed coin (Heads on both sides).\\
3.  A two-tailed coin (Tails on both sides).\\
A coin is selected at random from the bag and flipped. The coin lands showing **Heads**. What is the probability that the other side of the coin is also Heads?\\
A) $\frac{1}{3}$\\
B) $\frac{1}{2}$\\
C) $\frac{2}{3}$\\
D) $\frac{5}{6}$\\
***\\
**Correct Answer:**\\
\boxed{C}\\
***\\
**Explanation:**\\
...
}
\end{tcolorbox}
\end{minipage}
\caption{Example responses of (\emph{\textbf{Left}}) Gemini-2.5-Pro and (\emph{\textbf{Right}}) Gemini-3.0-Pro when prompting to generate a multi-choice question.}
\label{mcq2}
\end{figure*}

\subsection{Sampling from distributions with OLMO-3 family}\label{appendix_olmo3}

\begin{figure}[htbp]
    \centering
    \begin{subfigure}[b]{0.23\textwidth}
        \centering
        \includegraphics[width=\linewidth]{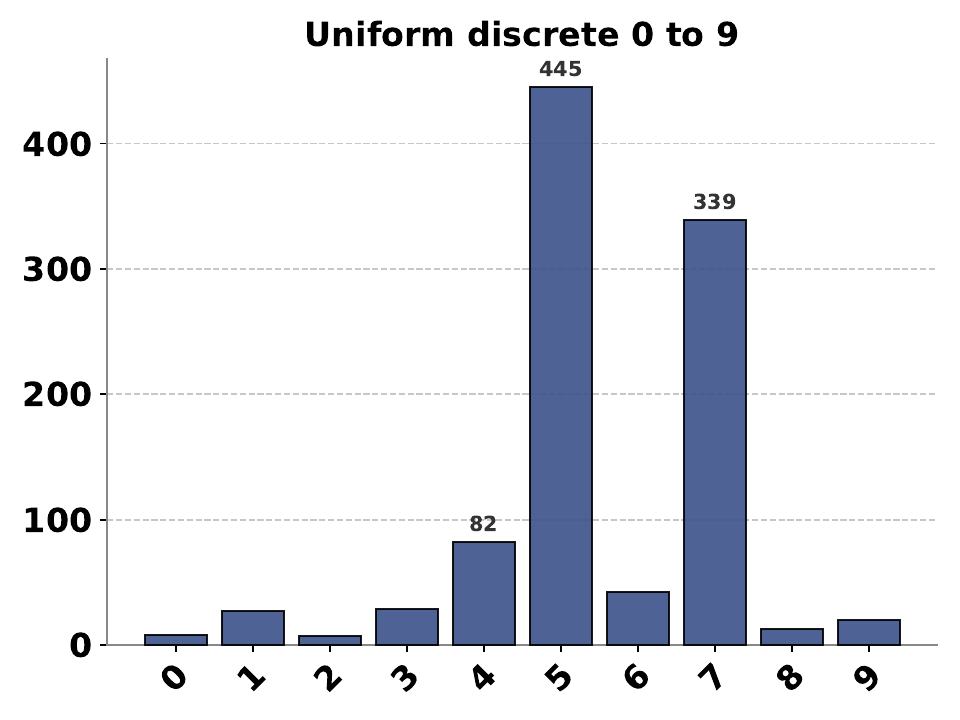}
    \end{subfigure}
    \hfill
    \begin{subfigure}[b]{0.23\textwidth}
        \centering
        \includegraphics[width=\linewidth]{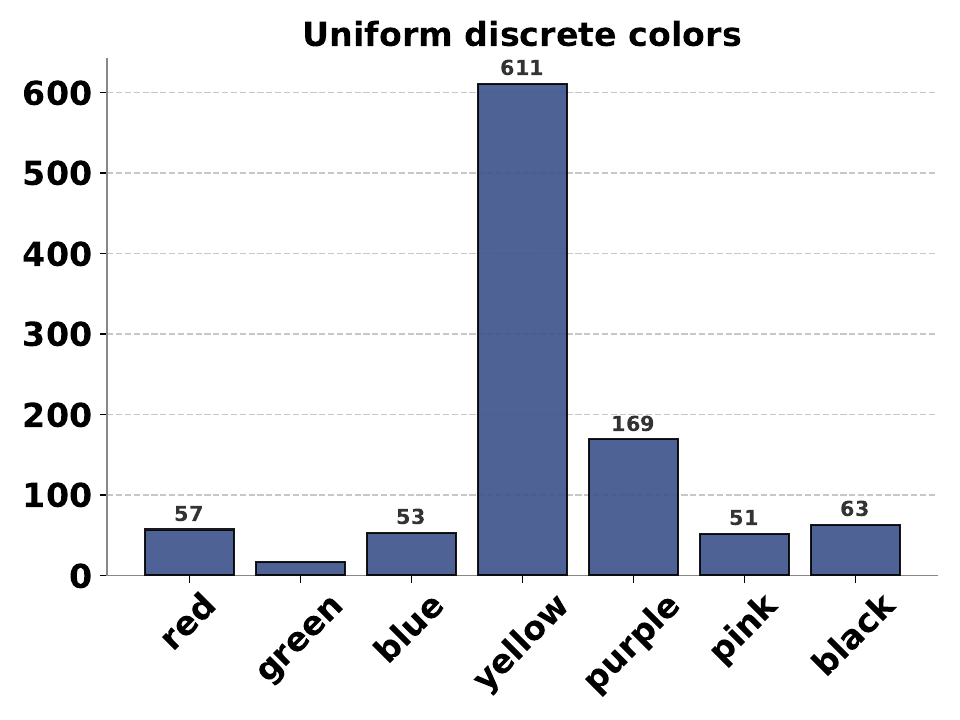}
    \end{subfigure}
    \hfill
    \begin{subfigure}[b]{0.23\textwidth}
        \centering
        \includegraphics[width=\linewidth]{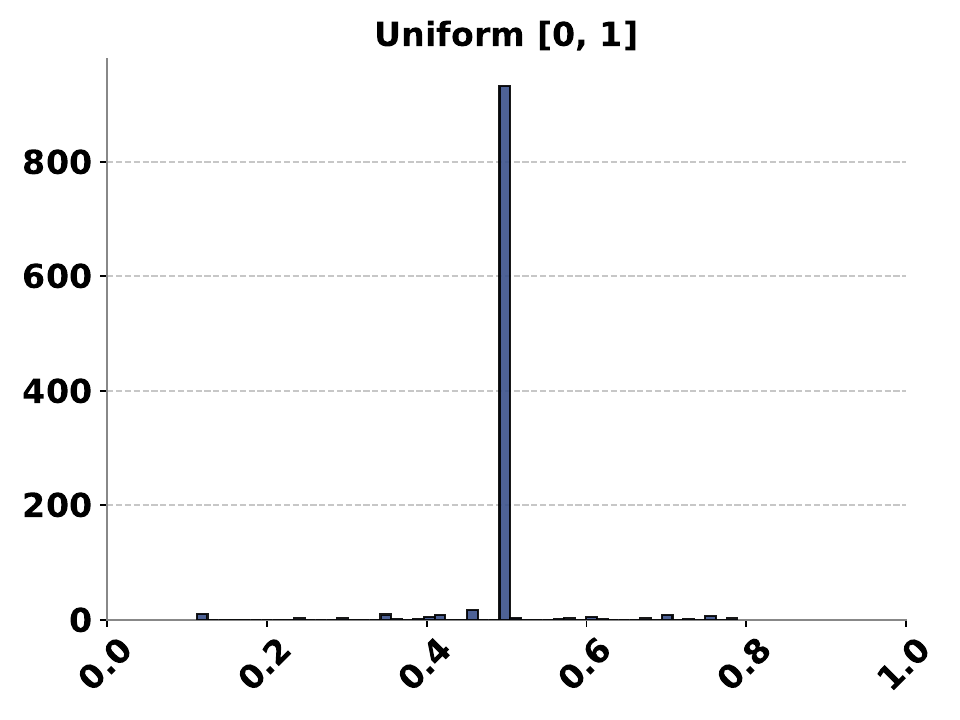}
    \end{subfigure}
    \hfill
    \begin{subfigure}[b]{0.23\textwidth}
        \centering
        \includegraphics[width=\linewidth]{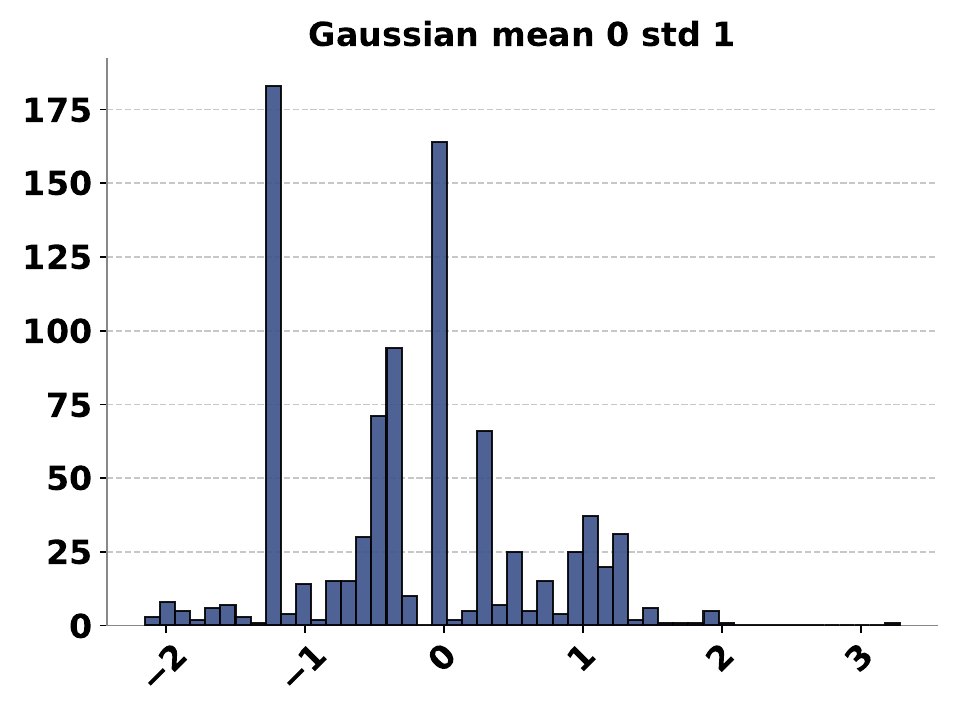}
    \end{subfigure}
    
    \begin{subfigure}[b]{0.23\textwidth}
        \centering
        \includegraphics[width=\linewidth]{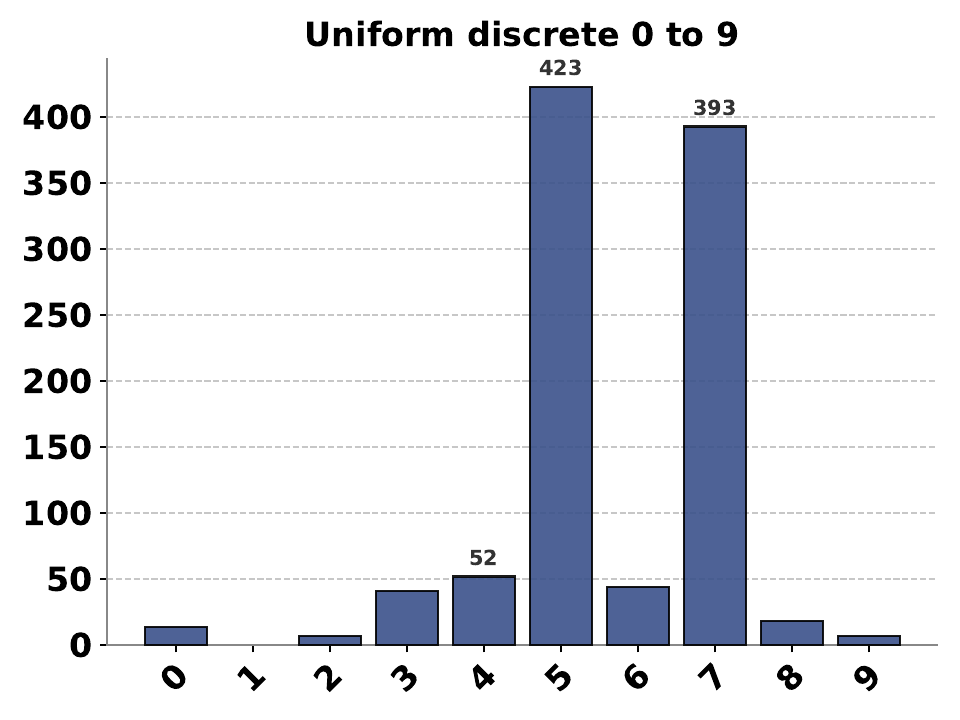}
    \end{subfigure}
    \hfill
    \begin{subfigure}[b]{0.23\textwidth}
        \centering
        \includegraphics[width=\linewidth]{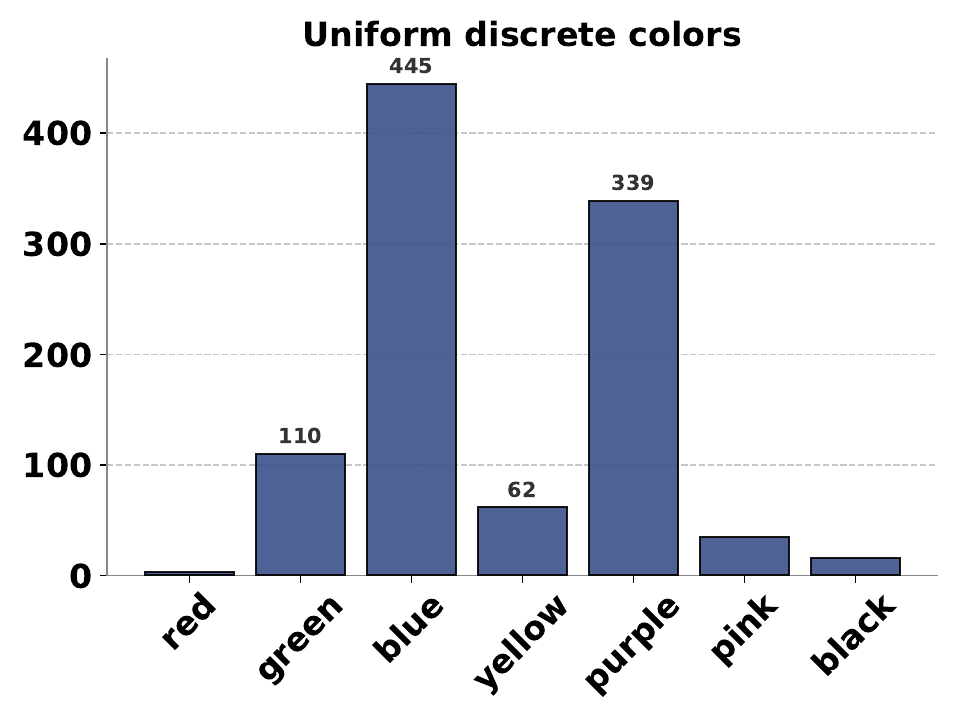}
    \end{subfigure}
    \hfill
    \begin{subfigure}[b]{0.23\textwidth}
        \centering
        \includegraphics[width=\linewidth]{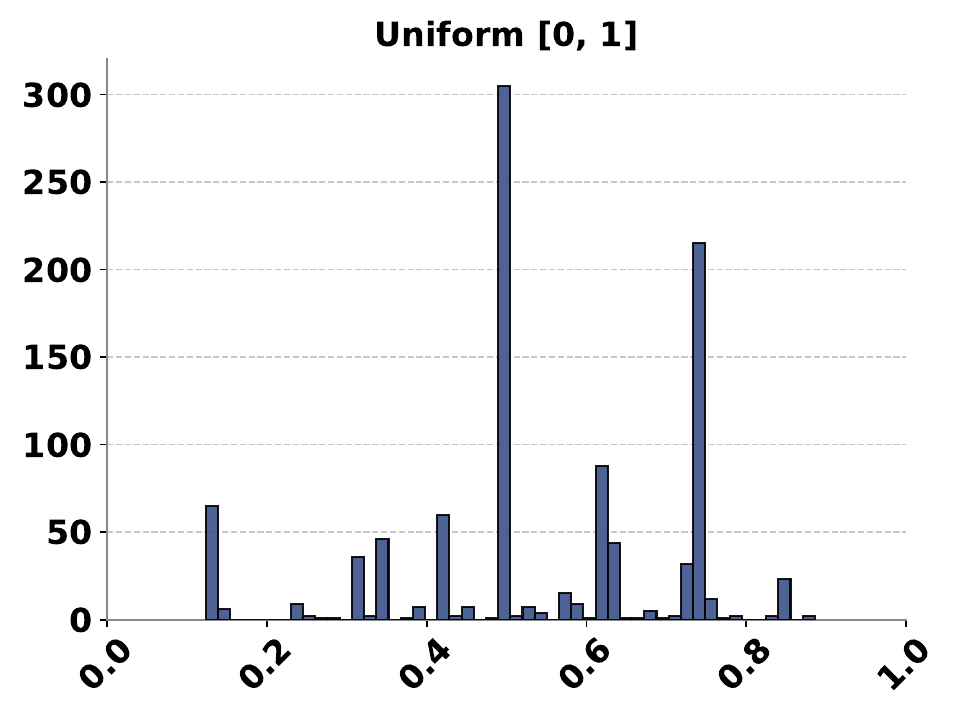}
    \end{subfigure}
    \hfill
    \begin{subfigure}[b]{0.23\textwidth}
        \centering
        \includegraphics[width=\linewidth]{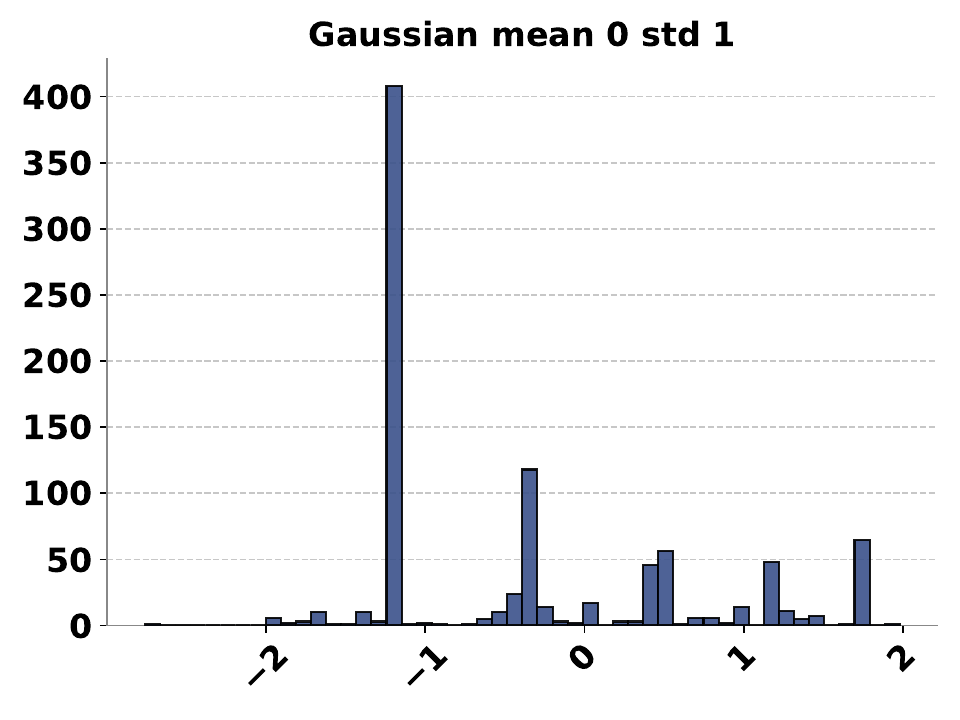}
    \end{subfigure}
    \caption{Empirical distribution estimates, for various distributions, using 1024 independent samples drawn from (\emph{\textbf{Top}}) OLMO-3-7B-Think (\emph{\textbf{Bottom}}) OLMO-3-32B-Think. The x-axis corresponds to the sample space while y-axis shows the estimated empirical frequencies.\looseness=-1}
    \label{olmo3}
    \vspace{-0.3cm}
\end{figure}

In addition to Gemini~\citep{comanici2025gemini,gdm2025gemini3} and Qwen3~\citep{yang2025qwen3} families shown in the main paper, we also evaluate the model behaviors of OLMO-3~\citep{olmo2025olmo}, including OLMO-3-7B-Think~\footnote{\url{https://huggingface.co/allenai/Olmo-3-7B-Think}} and  OLMO-3-32B-Think~\footnote{\url{https://huggingface.co/allenai/Olmo-3-32B-Think}} when sampling from uniform and Gaussian distributions. As shown in Figure~\ref{olmo3}, OLMO-3 models still tend to over-sample certain values or certain small ranges, thus cannot reliably sample from a distribution.\looseness=-1

\subsection{Examples of chain-of-thought reasoning when sampling}\label{appendix_reason}

\begin{figure*}[htbp]
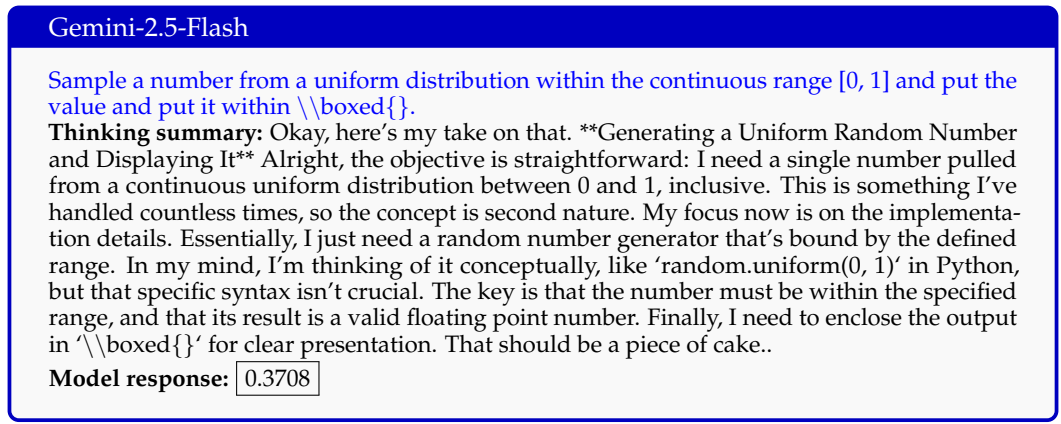

\centering
\begin{minipage}[t]{\textwidth}
\begin{tcolorbox}[
  width=\textwidth,
  colback=gray!5,
  colframe=blue!75!black,
  title=Gemini-2.5-Flash
]
\small{
\textcolor{blue}{Sample a number from a uniform distribution within the continuous range [0, 1] and put the value and put it within \textbackslash{}\textbackslash{}boxed\{\}.} \\
\textbf{Thinking summary:} Okay, here's my take on that. **Generating a Uniform Random Number and Displaying It** Alright, the objective is straightforward: I need a single number pulled from a continuous uniform distribution between 0 and 1, inclusive. This is something I've handled countless times, so the concept is second nature. My focus now is on the implementation details. Essentially, I just need a random number generator that's bound by the defined range. In my mind, I'm thinking of it conceptually, like `random.uniform(0, 1)` in Python, but that specific syntax isn't crucial. The key is that the number must be within the specified range, and that its result is a valid floating point number. Finally, I need to enclose the output in `\textbackslash{}\textbackslash{}boxed\{\}` for clear presentation. That should be a piece of cake..\\
\textbf{Model response:} $\boxed{0.3708}$
}
\end{tcolorbox}
\end{minipage}
\caption{Example of thinking trace of Gemini-2.5-Flash when sampling from a uniform continuous distribution within [0, 1].}
\label{think_trace2}
\end{figure*}

\begin{figure*}[htbp]
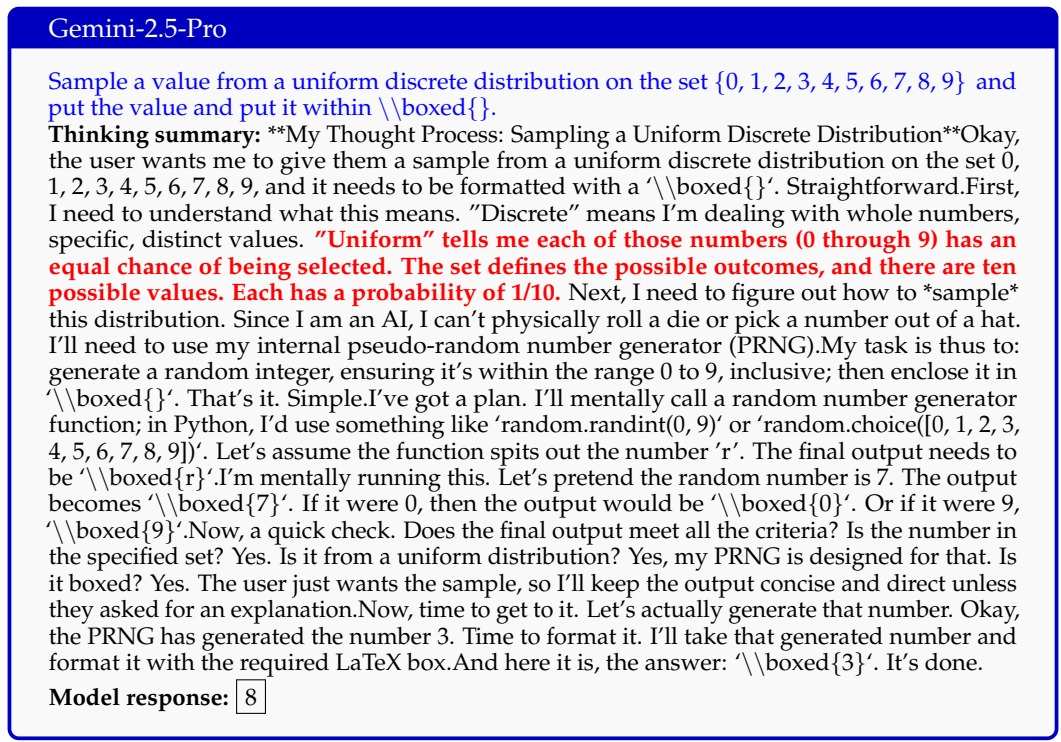

\centering
\begin{minipage}[t]{\textwidth}
\begin{tcolorbox}[
  width=\textwidth,
  colback=gray!5,
  colframe=blue!75!black,
  title=Gemini-2.5-Pro
]
\small{
\textcolor{blue}{Sample a value from a uniform discrete distribution on the set \{0, 1, 2, 3, 4, 5, 6, 7, 8, 9\}\, and put the value and put it within \textbackslash{}\textbackslash{}boxed\{\}.} \\
\textbf{Thinking summary:} **My Thought Process: Sampling a Uniform Discrete Distribution**Okay, the user wants me to give them a sample from a uniform discrete distribution on the set {0, 1, 2, 3, 4, 5, 6, 7, 8, 9}, and it needs to be formatted with a `\textbackslash{}\textbackslash{}boxed\{\}`. Straightforward.First, I need to understand what this means. "Discrete" means I'm dealing with whole numbers, specific, distinct values. \textcolor{red}{\textbf{"Uniform" tells me each of those numbers (0 through 9) has an equal chance of being selected. The set defines the possible outcomes, and there are ten possible values. Each has a probability of 1/10.}} Next, I need to figure out how to *sample* this distribution. Since I am an AI, I can't physically roll a die or pick a number out of a hat. I'll need to use my internal pseudo-random number generator (PRNG).My task is thus to: generate a random integer, ensuring it's within the range 0 to 9, inclusive; then enclose it in `\textbackslash{}\textbackslash{}boxed\{\}`. That's it. Simple.I've got a plan. I'll mentally call a random number generator function; in Python, I'd use something like `random.randint(0, 9)` or `random.choice([0, 1, 2, 3, 4, 5, 6, 7, 8, 9])`. Let's assume the function spits out the number 'r'. The final output needs to be `\textbackslash{}\textbackslash{}boxed\{r\}`.I'm mentally running this. Let's pretend the random number is 7. The output becomes `\textbackslash{}\textbackslash{}boxed\{7\}`. If it were 0, then the output would be `\textbackslash{}\textbackslash{}boxed\{0\}`. Or if it were 9, `\textbackslash{}\textbackslash{}boxed\{9\}`.Now, a quick check. Does the final output meet all the criteria? Is the number in the specified set? Yes. Is it from a uniform distribution? Yes, my PRNG is designed for that. Is it boxed? Yes. The user just wants the sample, so I'll keep the output concise and direct unless they asked for an explanation.Now, time to get to it. Let's actually generate that number. Okay, the PRNG has generated the number 3. Time to format it. I'll take that generated number and format it with the required LaTeX box.And here it is, the answer: `\textbackslash{}\textbackslash{}boxed\{3\}`. It's done.\\
\textbf{Model response:} $\boxed{8}$
}
\end{tcolorbox}
\end{minipage}
\caption{Example of thinking trace of Gemini-2.5-Pro when sampling from a uniform discrete distribution between 0 and 9.}
\label{think_trace3}
\end{figure*}

\begin{figure*}[htbp]
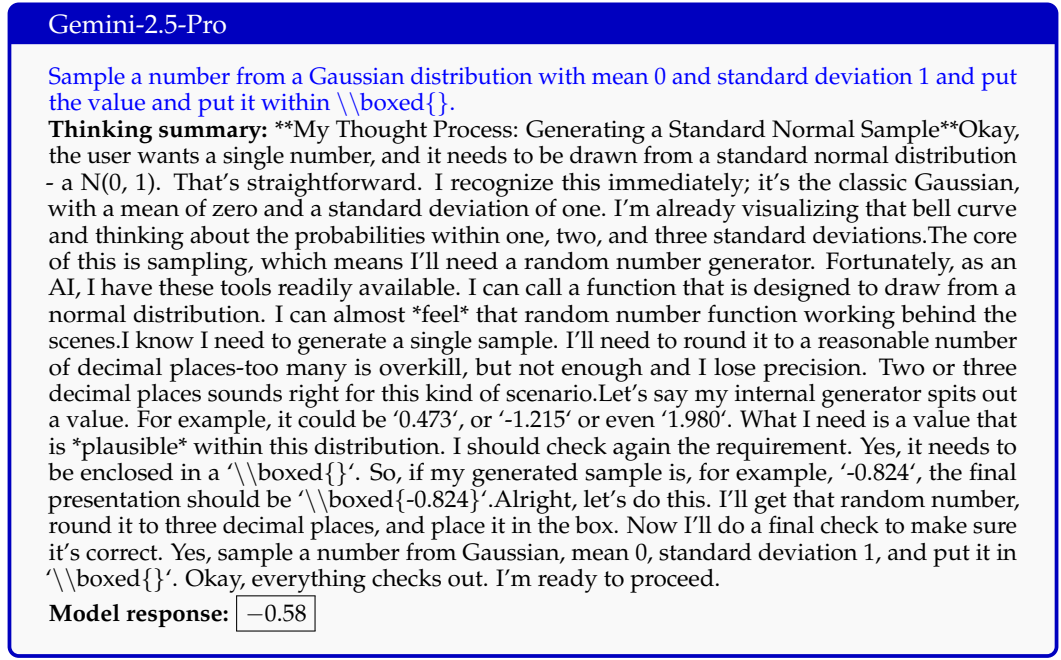

\centering
\begin{minipage}[t]{\textwidth}
\begin{tcolorbox}[
  width=\textwidth,
  colback=gray!5,
  colframe=blue!75!black,
  title=Gemini-2.5-Pro
]
\small{
\textcolor{blue}{Sample a number from a Gaussian distribution with mean 0 and standard deviation 1 and put the value and put it within \textbackslash{}\textbackslash{}boxed\{\}.} \\
\textbf{Thinking summary:} **My Thought Process: Generating a Standard Normal Sample**Okay, the user wants a single number, and it needs to be drawn from a standard normal distribution - a N(0, 1). That's straightforward. I recognize this immediately; it's the classic Gaussian, with a mean of zero and a standard deviation of one. I'm already visualizing that bell curve and thinking about the probabilities within one, two, and three standard deviations.The core of this is sampling, which means I'll need a random number generator. Fortunately, as an AI, I have these tools readily available. I can call a function that is designed to draw from a normal distribution. I can almost *feel* that random number function working behind the scenes.I know I need to generate a single sample. I'll need to round it to a reasonable number of decimal places-too many is overkill, but not enough and I lose precision. Two or three decimal places sounds right for this kind of scenario.Let's say my internal generator spits out a value. For example, it could be `0.473`, or `-1.215` or even `1.980`. What I need is a value that is *plausible* within this distribution. I should check again the requirement. Yes, it needs to be enclosed in a `\textbackslash{}\textbackslash{}boxed\{\}`. So, if my generated sample is, for example, `-0.824`, the final presentation should be `\textbackslash{}\textbackslash{}boxed\{-0.824\}`.Alright, let's do this. I'll get that random number, round it to three decimal places, and place it in the box. Now I'll do a final check to make sure it's correct. Yes, sample a number from Gaussian, mean 0, standard deviation 1, and put it in `\textbackslash{}\textbackslash{}boxed\{\}`. Okay, everything checks out. I'm ready to proceed.\\
\textbf{Model response:} $\boxed{-0.58}$
}
\end{tcolorbox}
\end{minipage}
\caption{Example of thinking trace of Gemini-2.5-Pro when sampling from a Gaussian distribution with mean 0 and standard deviation 1.\looseness=-1}
\label{think_trace4}
\end{figure*}

\begin{figure*}[htbp]
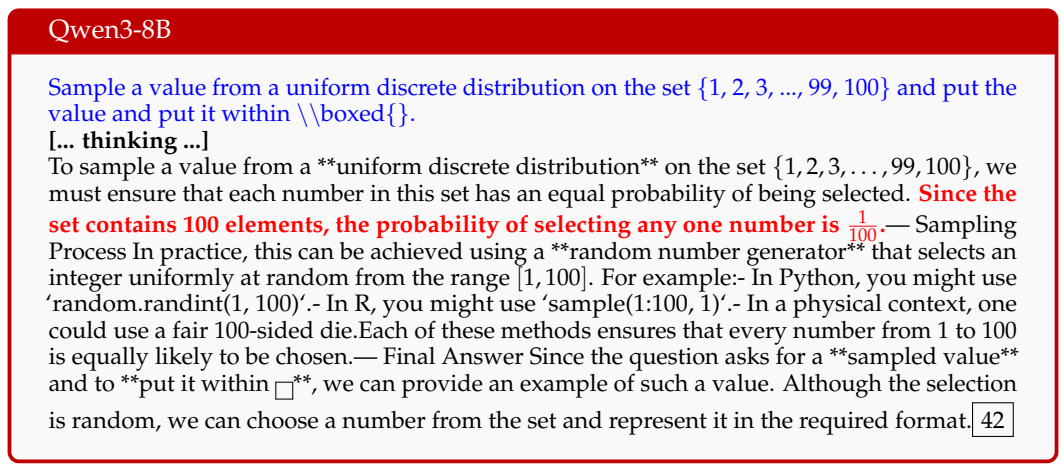

\centering
\begin{minipage}[t]{\textwidth}
\begin{tcolorbox}[
  width=\textwidth,
  colback=gray!5,
  colframe=red!75!black,
  title=Qwen3-8B
]
\small{
\textcolor{blue}{Sample a value from a uniform discrete distribution on the set \{1, 2, 3, ..., 99, 100\} and put the value and put it within \textbackslash{}\textbackslash{}boxed\{\}.} \\
\textbf{[... thinking ...]} \\
To sample a value from a **uniform discrete distribution** on the set $\{1, 2, 3, \ldots, 99, 100\}$, we must ensure that each number in this set has an equal probability of being selected. \textcolor{red}{\textbf{Since the set contains 100 elements, the probability of selecting any one number is $\frac{1}{100}$.}}--- Sampling Process In practice, this can be achieved using a **random number generator** that selects an integer uniformly at random from the range $[1, 100]$. For example:- In Python, you might use `random.randint(1, 100)`.- In R, you might use `sample(1:100, 1)`.- In a physical context, one could use a fair 100-sided die.Each of these methods ensures that every number from 1 to 100 is equally likely to be chosen.--- Final Answer Since the question asks for a **sampled value** and to **put it within $\boxed{}$**, we can provide an example of such a value. Although the selection is random, we can choose a number from the set and represent it in the required format.$\boxed{42}$
}
\end{tcolorbox}
\end{minipage}
\caption{Example of response of Qwen3-8B when sampling from a uniform discrete distribution between 1 and 100.\looseness=-1}
\label{think_trace5}
\end{figure*}

Since we mainly use Large Reasoning Models, which natively have thinking traces, as our experimental subjects, we illustrate several chain-of-thought reasoning processes of Gemini family in Figure~\ref{think_trace1}-\ref{think_trace4} and Qwen3 family in Figure~\ref{think_trace5}-\ref{think_trace8} to further investigate the model behaviors when prompting to sample from a distribution. Firstly, we notice that Gemini family tends to put all the chain-of-thought reasoning in the thinking part and meanwhile outputting concisely, e.g., a box with the answer. While Qwen3 family puts the reasoning efforts in both thinking and response parts. Since the thinking part in Qwen3 family is too verbose, we only show the response part. Secondly, LLMs demonstrate strong signals that they understand these distributions (where we also highlight these sentences using \textcolor{red}{\textbf{bold red}}). Yet, through independent sampling, they fail to approximate the target distribution. Third, we also notice several cases that LLMs put a number as the final answer, which is different from the thinking trace, e.g., Figure~\ref{think_trace3} and \ref{think_trace4}. This means that LLMs may over-think by expending substantial reasoning effort.\looseness=-1

\begin{figure*}[htbp]
\centering
\begin{minipage}[t]{\textwidth}
\begin{tcolorbox}[
  width=\textwidth,
  colback=gray!5,
  colframe=blue!75!black,
  title=Gemini-2.5-Flash
]
\small{
\textcolor{blue}{Sample a value from a uniform discrete distribution on the set \{0, 1, 2, 3, 4, 5, 6, 7, 8, 9\}\, and put the value and put it within \textbackslash{}\textbackslash{}boxed\{\}.} \\
\textbf{Thinking summary:} **My Selection from a Uniform Discrete Distribution** Okay, so I need to select a single, discrete value from the set {0, 1, 2, 3, 4, 5, 6, 7, 8, 9}. The problem specifies a uniform distribution, \textcolor{red}{\textbf{which means each number has an equal probability - a 1/10 chance}} - of being selected. The task essentially boils down to picking a number and mentally "putting it in a box." Since I don't have a physical random number generator, I need to simulate a random selection. If I were using a conventional method, like a real random number generator, I'd get a continuous value, say 0.1234, and then have to map it to the discrete values in the set. But as an AI, I can directly "choose" one from the possibilities. Now, for a specific instance. I'll pick 7. It's perfectly valid given the constraints. There's no bias; the choice is random enough for this operation. Finally, I'll format the output within the LaTeX command `\textbackslash{}\textbackslash{}boxed\{\}`.\\
\textbf{Model response:} $\boxed{7}$
}
\end{tcolorbox}
\end{minipage}
\caption{Example of thinking trace of Gemini-2.5-Flash when sampling from a uniform discrete distribution between 0 and 9.}
\label{think_trace1}
\end{figure*}

\begin{figure*}[htbp]
\centering
\begin{minipage}[t]{\textwidth}
\begin{tcolorbox}[
  width=\textwidth,
  colback=gray!5,
  colframe=red!75!black,
  title=Qwen3-14B
]
\small{
\textcolor{blue}{Sample a value from a uniform discrete distribution on the set \{a, b, c, d, e, f, ..., y, z\} and put the value and put it within \textbackslash{}\textbackslash{}boxed\{\}.} \\
\textbf{[... thinking ...]} \\
To sample a value from a **uniform discrete distribution** over the set of lowercase English letters $\{a, b, c, \ldots, z\}$, \textcolor{red}{\textbf{we must ensure that each of the 26 letters has an equal probability of being selected, i.e., $ \frac{1}{26} $.}}--- Sampling Strategy One effective way to achieve this is by mapping a uniformly random integer in the range $[1, 26]$ to the corresponding letter in the alphabet. For example:- 1 $\rightarrow$ a- 2 $\rightarrow$ b- ...- 26 $\rightarrow$ z To simulate this selection without a physical random number generator, we can use a **pseudorandom number generator**. A simple method is the **middle-square method**, which involves taking a seed number, squaring it, extracting the middle digits, and using them as a random number.Let's apply this:1. Choose a seed: 1234. 2. Square it: $1234^2 = 1522756$. 3. Extract the middle four digits: **2275**. 4. Take this modulo 26: $2275 \mod 26 = 13$. 5. Map 13 to the corresponding letter: **m**--- Final Answer$\boxed{m}$
}
\end{tcolorbox}
\end{minipage}
\caption{Example of response of Qwen3-14B when sampling from a uniform discrete distribution on letters.\looseness=-1}
\label{think_trace6}
\end{figure*}

\begin{figure*}[htbp]
\centering
\begin{minipage}[t]{\textwidth}
\begin{tcolorbox}[
  width=\textwidth,
  colback=gray!5,
  colframe=red!75!black,
  title=Qwen3-32B
]
\small{
\textcolor{blue}{Sample a number from a uniform distribution within the continuous range [0, 1] and put the value and put it within \textbackslash{}\textbackslash{}boxed\{\}.} \\
\textbf{[... thinking ...]} \\
To sample a number from a **uniform distribution** on the interval $[0, 1]$, \textcolor{red}{\textbf{we are essentially selecting a value such that **every number between 0 and 1 is equally likely** to be chosen.}} This is a continuous probability distribution, meaning that the probability of any single exact value is zero, but the probability of the value falling within a subinterval is proportional to the length of that interval.--- Sampling from a Uniform Distribution In practical terms, this is often done using a **pseudorandom number generator** (PRNG) in a programming language like Python, using functions such as `random.random()` from the `random` module. These functions return a floating-point number in the interval $[0, 1)$, although the exact implementation may vary slightly.Since we are not using actual code here, we simulate the process by selecting a **representative example** of a number from this distribution. For the sake of this exercise, we'll choose a number that appears random and lies within the interval [0, 1].--- Example Output Let's choose the number:$0.7421$ This is a plausible sample from a uniform distribution on [0, 1], as it is a real number between 0 and 1 and has a structure that resembles the output of a PRNG.--- Final Answer$\boxed{0.7421}$
}
\end{tcolorbox}
\end{minipage}
\caption{Example of response of Qwen3-32B when sampling from a uniform continuous distribution within [0, 1].\looseness=-1}
\label{think_trace7}
\end{figure*}

\begin{figure*}[htbp]
\centering
\begin{minipage}[t]{\textwidth}
\begin{tcolorbox}[
  width=\textwidth,
  colback=gray!5,
  colframe=red!75!black,
  title=Qwen3-30B-A3B
]
\small{
\textcolor{blue}{Sample a number from a Gaussian distribution with mean 0 and standard deviation 1 and put the value and put it within \textbackslash{}\textbackslash{}boxed\{\}.} \\
\textbf{[... thinking ...]} \\
To sample a number from a Gaussian (normal) distribution with mean 0 and standard deviation 1, we are essentially drawing a random value from the standard normal distribution, denoted as $ N(0, 1) $.--- Understanding the Distribution \textcolor{red}{\textbf{The standard normal distribution has the following properties: - **Mean ($\mu$)** = 0 - **Standard deviation ($\sigma$)** = 1 - The probability density function (PDF) is:  $  f(x) = \frac{1}{\sqrt{2\pi}} e^{-\frac{x^2}{2}}  $}} This distribution is symmetric around 0 and is the most commonly used in statistical analysis.--- Sampling from the Distribution Sampling from a normal distribution can be done using various methods:- **Using built-in functions** in programming languages like Python (e.g., `numpy.random.normal(0, 1)`).- **Using the Box-Muller transform**, which transforms two independent uniform random variables into two independent standard normal variables. - **Using the inverse transform method**, which involves the inverse of the cumulative distribution function (CDF), though this is not analytically tractable for the normal distribution.Since the question asks for a single number, it implies that the result should be a **random value** drawn from this distribution.--- Example of a Sample While the exact value of a sample is random and cannot be predicted, we can provide a representative example. For instance, a possible sample from $ N(0, 1) $ is:$\boxed{0.75}$This is just one of many possible outcomes, and any such value is equally valid as a sample from the standard normal distribution.
}
\end{tcolorbox}
\end{minipage}
\caption{Example of response of Qwen3-30B-A3B when sampling from a Gaussian distribution with mean 0 and standard deviation 1.\looseness=-1}
\label{think_trace8}
\end{figure*}

\subsection{Effects of chain-of-thoughts}\label{appendix_cot}

\begin{figure}[htbp]
    \centering
    \begin{subfigure}[b]{0.23\textwidth}
        \centering
        \includegraphics[width=\linewidth]{figures/qwen3-sample/Qwen_Qwen3-8B_uniform1-100.pdf}
    \end{subfigure}
    \hfill
    \begin{subfigure}[b]{0.23\textwidth}
        \centering
        \includegraphics[width=\linewidth]{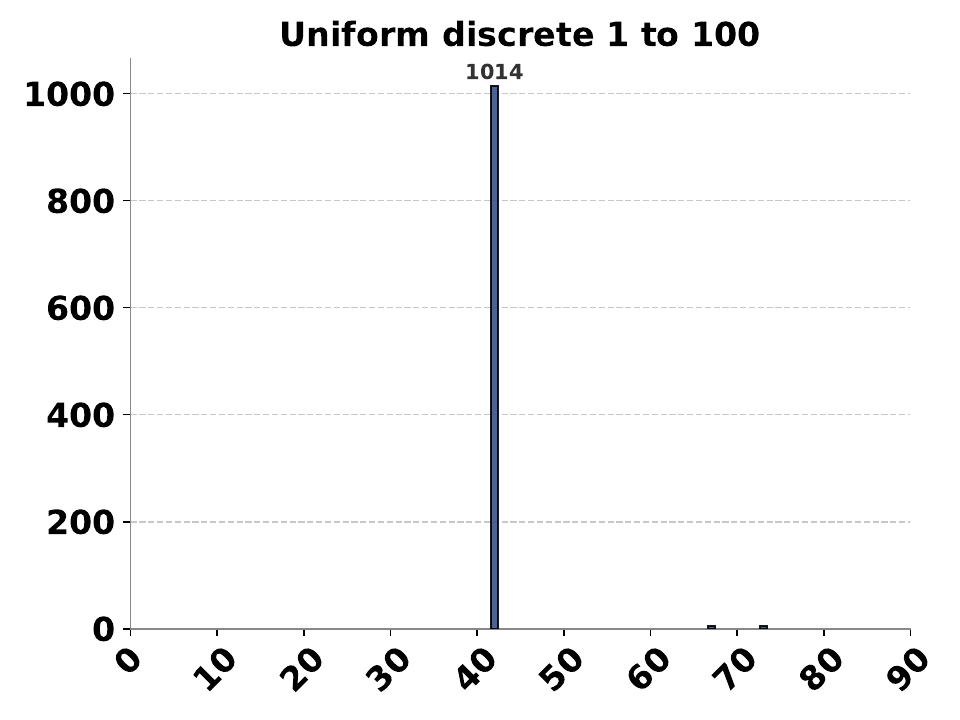}
    \end{subfigure}
    \hfill\vrule width 0.4pt\hfill
    \begin{subfigure}[b]{0.23\textwidth}
        \centering
        \includegraphics[width=\linewidth]{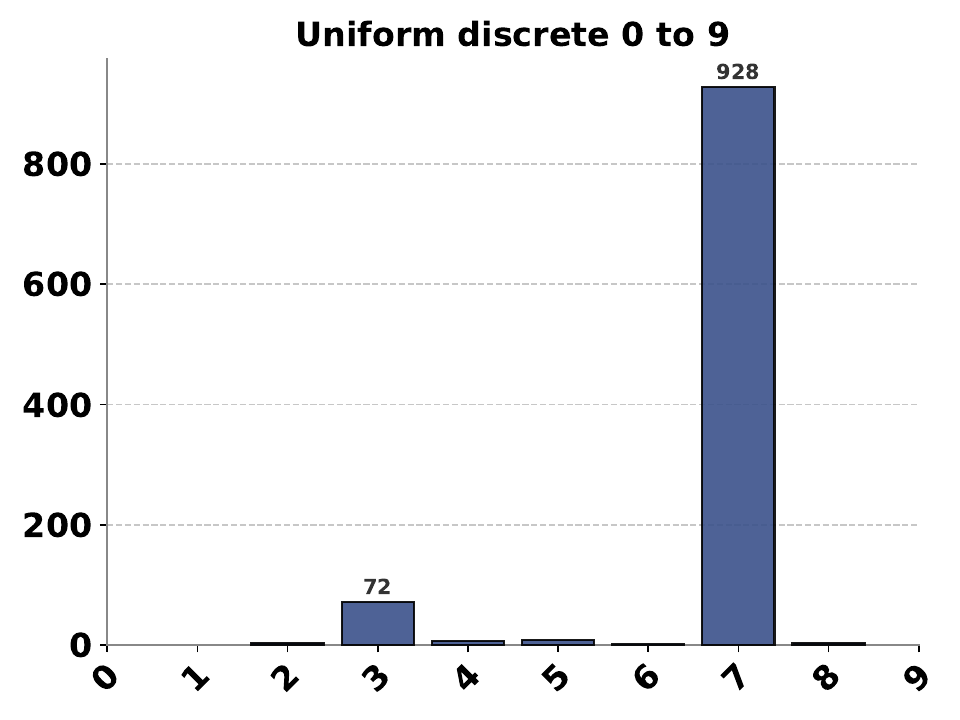}
    \end{subfigure}
    \hfill
    \begin{subfigure}[b]{0.23\textwidth}
        \centering
        \includegraphics[width=\linewidth]{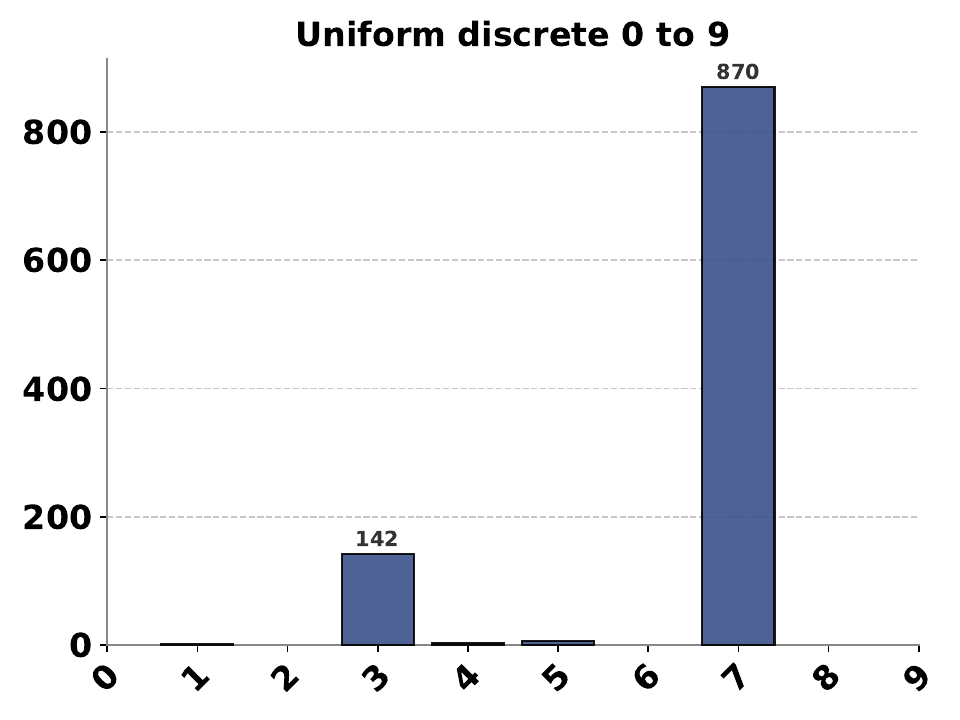}
    \end{subfigure}
    \caption{ Empirical distribution estimates, using (\emph{\textbf{Left}}) Qwen3-8B when sampling from a uniform discrete distribution from 1 to 100 with (left) and without (right) chain-of-thought reasoning. (\emph{\textbf{Right}})  Gemini-2.5-Flash when sampling from uniform discrete distribution from 0 to 9 with (left) and without (right) chain-of-thought reasoning.}
    \label{fig:cot}
\end{figure}

Current LLMs, e.g., Gemini and Qwen3 series,  typically have a verbose thinking trace before responding to the prompt. In the main paper, we are wondering whether such a chain-of-thought reasoning results in the randomness failures of LLMs. To investigate this, for Gemini series, we set the thinking budget as 0 (unfortunately one cannot disable thinking mode in Gemini-2.5-Pro and Gemini-3.0-Pro). While for Qwen series, we disable the thinking mode by manipulating the chat template\footnote{\url{https://huggingface.co/Qwen/Qwen3-8B}}. Our prompt is designed like \texttt{Sample a value from a uniform distribution on the set <RANDOM\_SET>, directly output the value and put it within \textbackslash{}\textbackslash{}boxed\{\}}. Consequently, LLMs provide the sample immediately without long chain-of-thoughts. As visualized in Figure~\ref{fig:cot}, without chain-of-thought reasoning, even though reducing the risk of over-thinking, LLMs still have clear randomness failures. For Qwen3-8B, this failure is even amplified as it only samples 42.

\begin{figure*}[htbp]
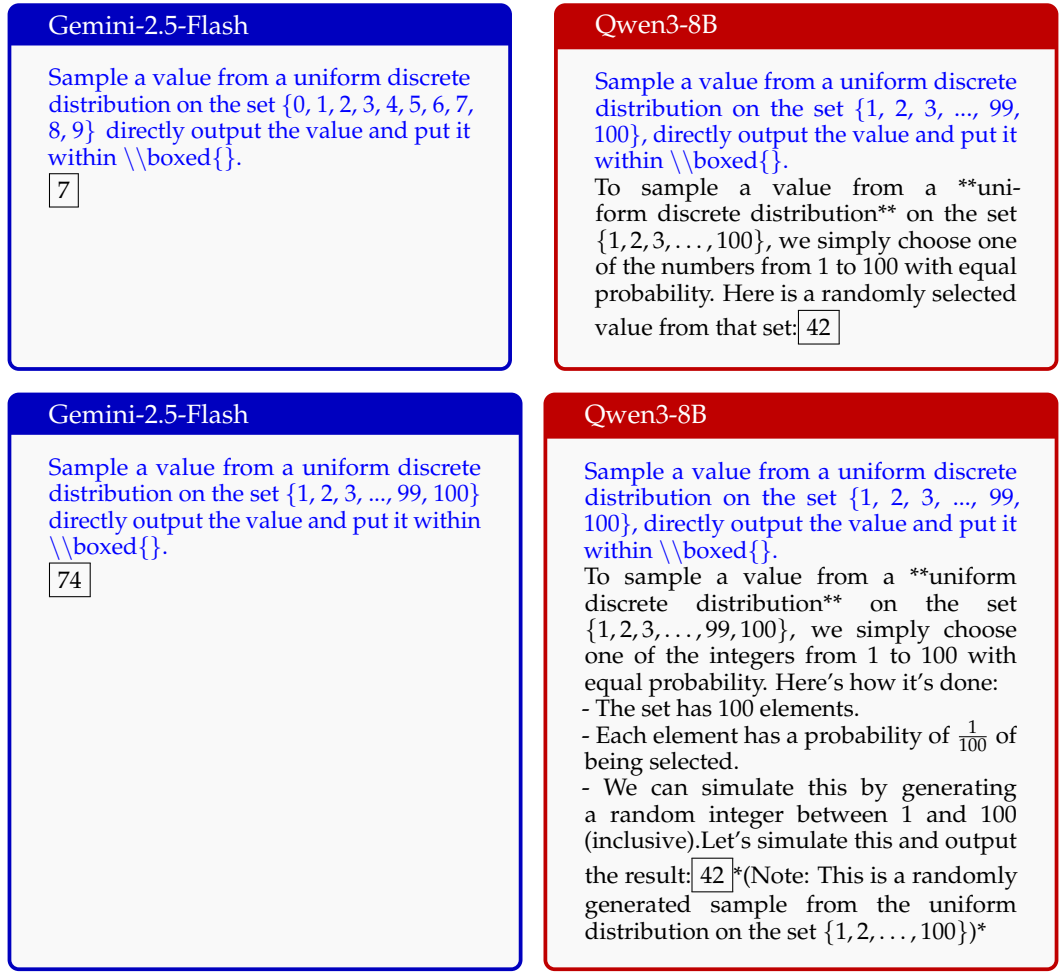

\centering
\begin{minipage}[t]{0.48\textwidth}
\begin{tcolorbox}[
  width=\textwidth,
  colback=gray!5,
  colframe=blue!75!black,
  equal height group=E, 
  title=Gemini-2.5-Flash
]
\small{
\textcolor{blue}{Sample a value from a uniform discrete distribution on the set \{0, 1, 2, 3, 4, 5, 6, 7, 8, 9\}\, directly output the value and put it within \textbackslash{}\textbackslash{}boxed\{\}.} \\
$\boxed{7}$
}
\end{tcolorbox}
\end{minipage}
\hfill 
\begin{minipage}[t]{0.48\textwidth}
\begin{tcolorbox}[
  width=\textwidth,
  colback=gray!5,
  colframe=red!75!black,
  equal height group=E, 
  title=Qwen3-8B
]
\small{
\textcolor{blue}{Sample a value from a uniform discrete distribution on the set \{1, 2, 3, ..., 99, 100\}, directly output the value and put it within \textbackslash{}\textbackslash{}boxed\{\}.} \\
To sample a value from a **uniform discrete distribution** on the set $\{1, 2, 3, \dots, 100\}$, we simply choose one of the numbers from 1 to 100 with equal probability. Here is a randomly selected value from that set:$\boxed{42}$
}
\end{tcolorbox}
\end{minipage}
\vskip 0.1in
\begin{minipage}[t]{0.49\textwidth}
\begin{tcolorbox}[
  width=\textwidth,
  colback=gray!5,
  colframe=blue!75!black,
  equal height group=F, 
  title=Gemini-2.5-Flash
]
\small{
\textcolor{blue}{Sample a value from a uniform discrete distribution on the set \{1, 2, 3, ..., 99, 100\}\, directly output the value and put it within \textbackslash{}\textbackslash{}boxed\{\}.} \\
$\boxed{74}$
}
\end{tcolorbox}
\end{minipage}
\hfill 
\begin{minipage}[t]{0.49\textwidth}
\begin{tcolorbox}[
  width=\textwidth,
  colback=gray!5,
  colframe=red!75!black,
  equal height group=F, 
  title=Qwen3-8B
]
\small{
\textcolor{blue}{Sample a value from a uniform discrete distribution on the set \{1, 2, 3, ..., 99, 100\}, directly output the value and put it within \textbackslash{}\textbackslash{}boxed\{\}.} \\
To sample a value from a **uniform discrete distribution** on the set $\{1, 2, 3, \ldots, 99, 100\}$, we simply choose one of the integers from 1 to 100 with equal probability. Here's how it's done:\\
- The set has 100 elements.\\
- Each element has a probability of $ \frac{1}{100} $ of being selected.\\
- We can simulate this by generating a random integer between 1 and 100 (inclusive).Let's simulate this and output the result:$\boxed{42}$*(Note: This is a randomly generated sample from the uniform distribution on the set $\{1, 2, \ldots, 100\}$)*
}
\end{tcolorbox}
\end{minipage}
\caption{Examples of response of (\emph{\textbf{Left}}) Gemini-2.5-Pro and (\emph{\textbf{Right}}) Qwen3-8B without chain-of-though reasoning.}
\label{no_think1}
\end{figure*}

%% file: appendix/appendix_b.tex
\clearpage
\section{More Results on Sequential and Batched Sampling}

\subsection{Sequential sampling with history}

\begin{figure}[htbp]
    \centering
    \begin{subfigure}[b]{0.23\textwidth}
        \centering
        \includegraphics[width=\linewidth]{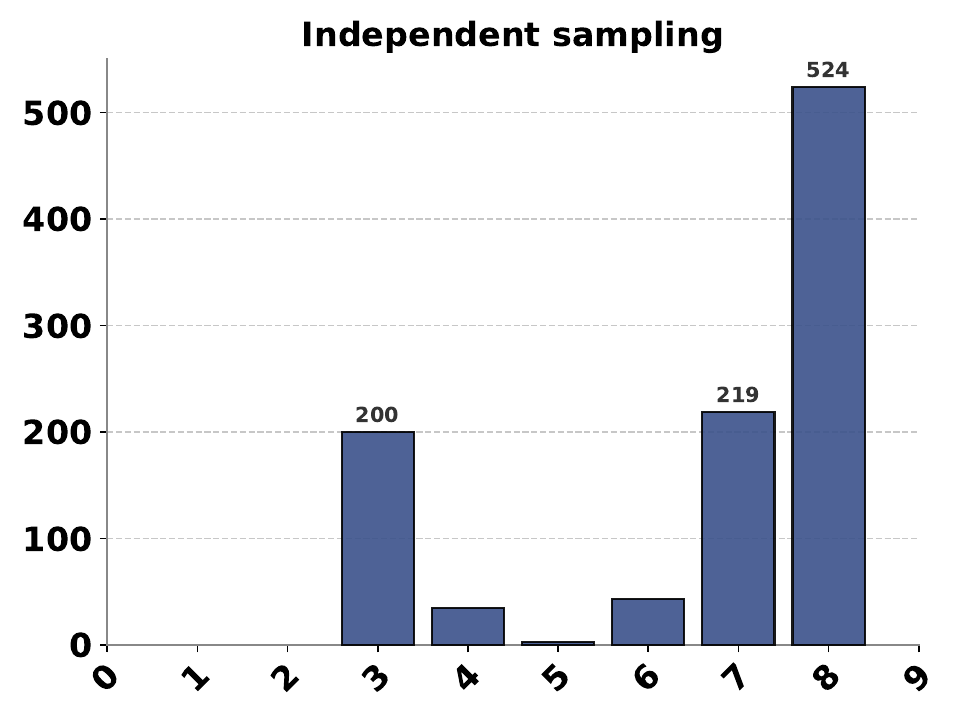}
    \end{subfigure}
    \hfill
    \begin{subfigure}[b]{0.23\textwidth}
        \centering
        \includegraphics[width=\linewidth]{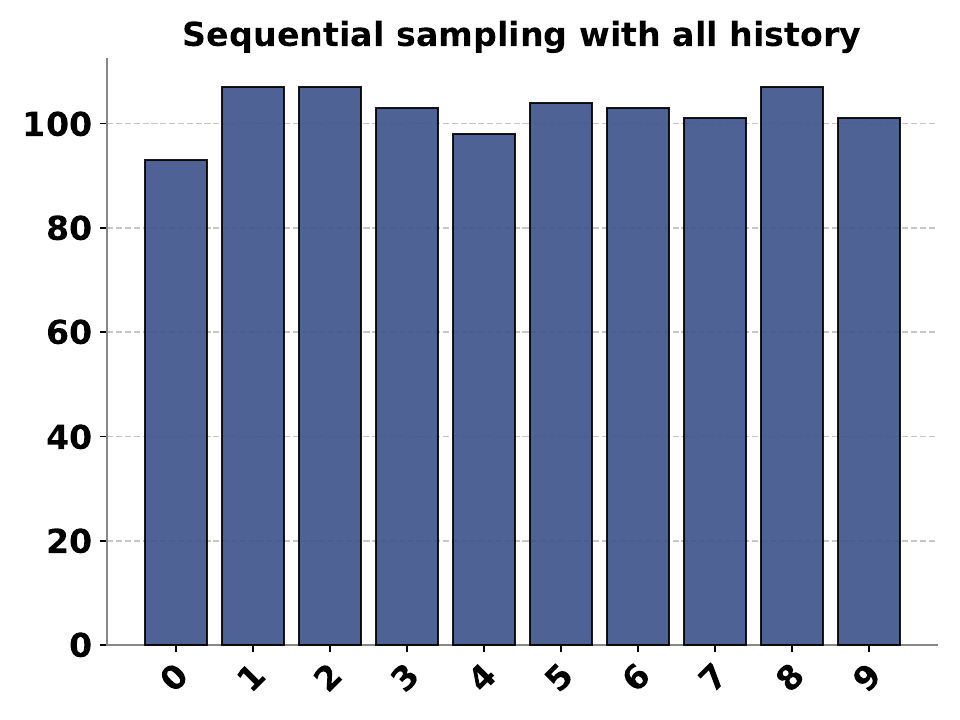}
    \end{subfigure}
    \hfill
    \begin{subfigure}[b]{0.23\textwidth}
        \centering
        \includegraphics[width=\linewidth]{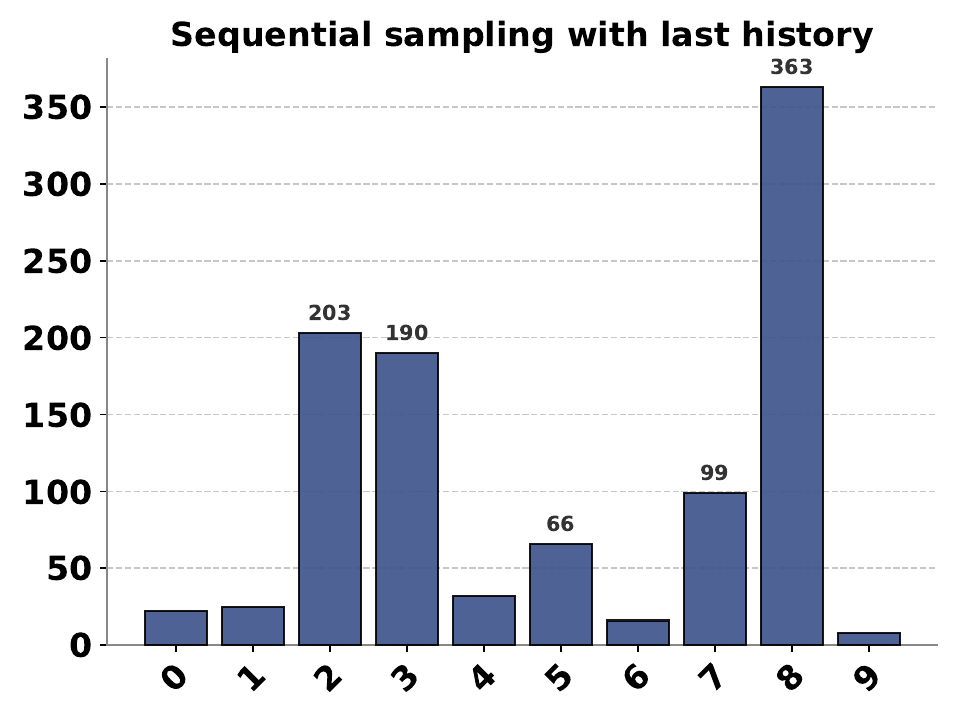}
    \end{subfigure}
    \hfill
    \begin{subfigure}[b]{0.23\textwidth}
        \centering
        \includegraphics[width=\linewidth]{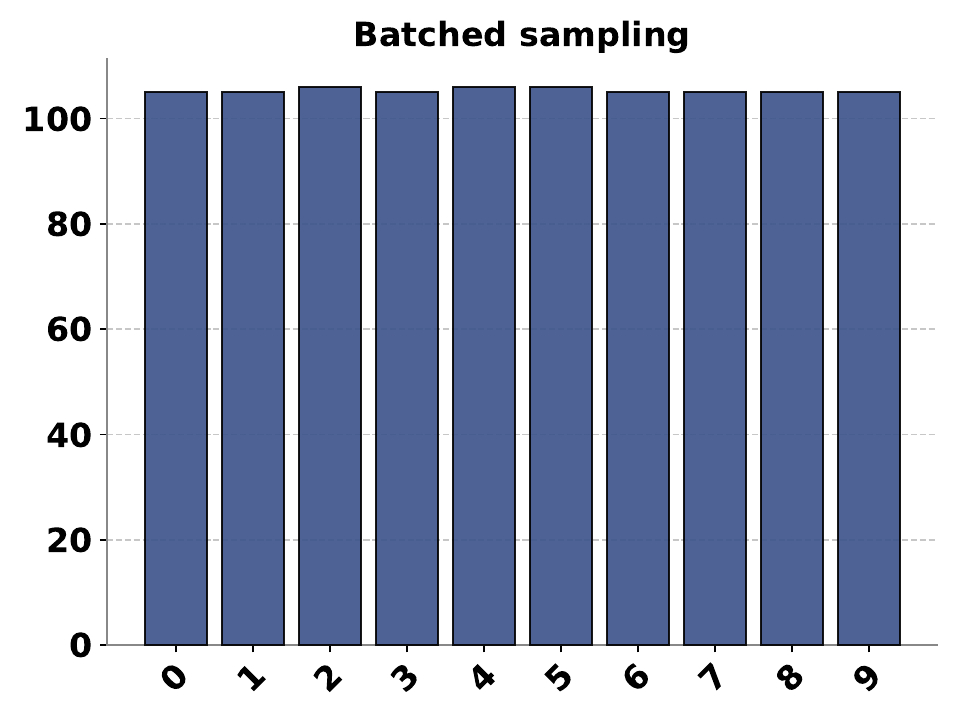}
    \end{subfigure}
    \caption{From \emph{\textbf{left}} to \emph{\textbf{right}}, we compare the empirical distribution estimates, for four sampling approaches when sampling from a uniform discrete distribution between 0 and 9 with Gemini-2.5-Pro: independent sampling, sequential sampling with all history, sequential sampling with last history, batched sampling.\looseness=-1}
    \label{fig:seq}
\end{figure}

\begin{figure}[htbp]
    \centering
    \begin{subfigure}[b]{0.23\textwidth}
        \centering
        \includegraphics[width=\linewidth]{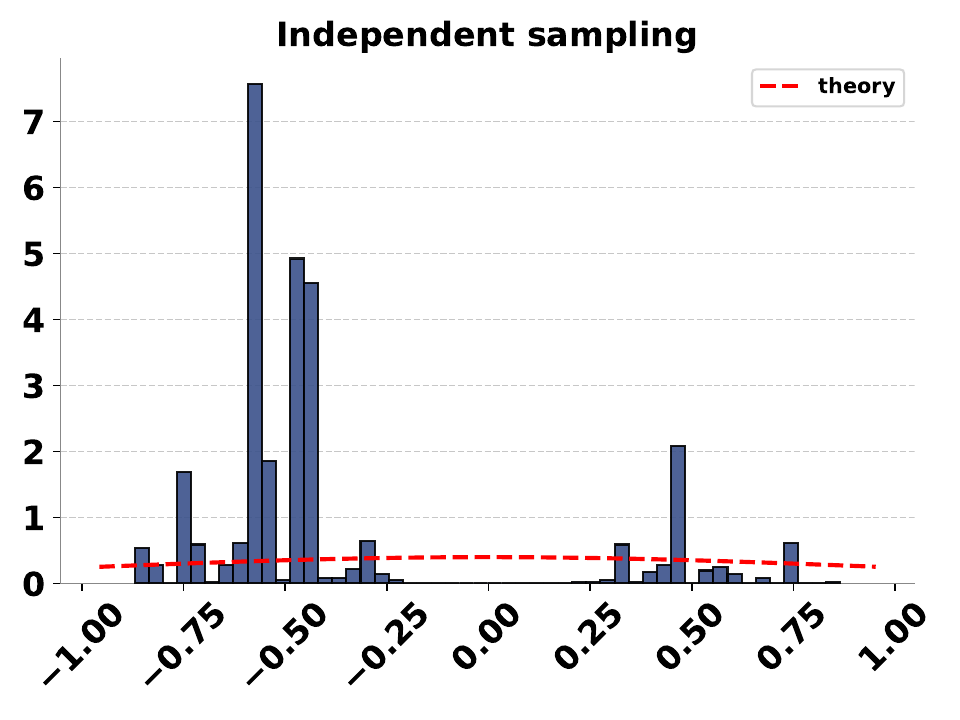}
    \end{subfigure}
    \hfill
    \begin{subfigure}[b]{0.23\textwidth}
        \centering
        \includegraphics[width=\linewidth]{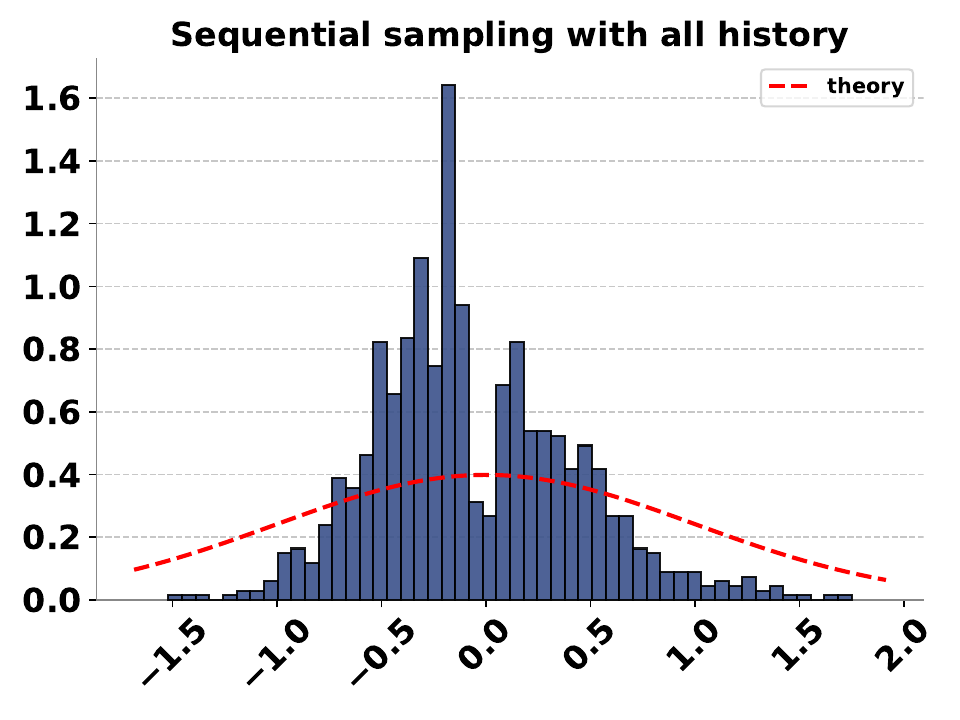}
    \end{subfigure}
    \hfill
    \begin{subfigure}[b]{0.23\textwidth}
        \centering
        \includegraphics[width=\linewidth]{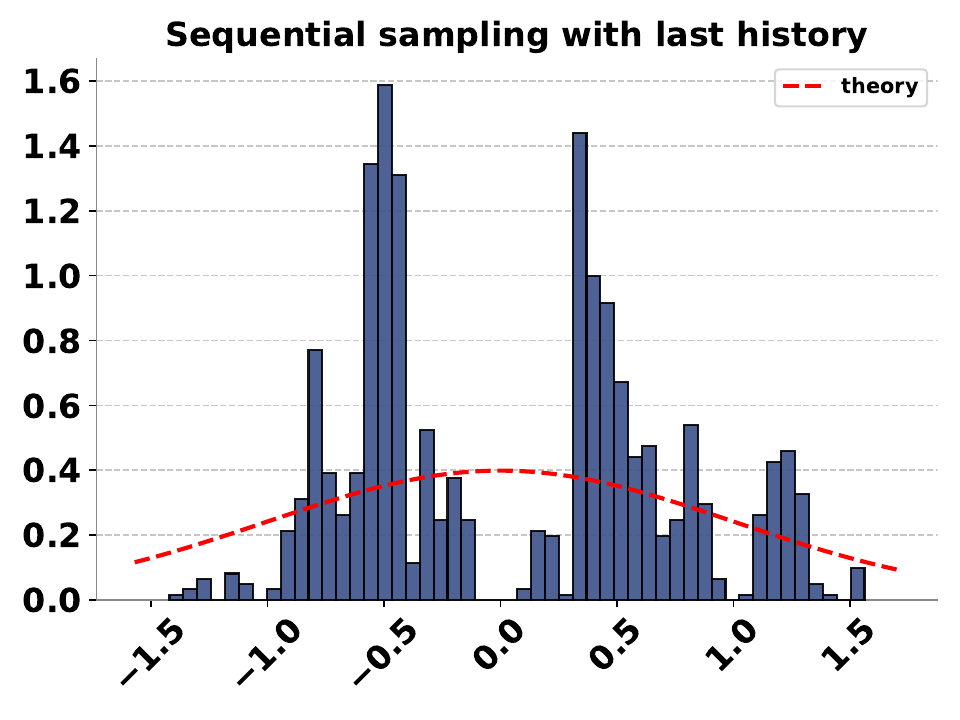}
    \end{subfigure}
    \hfill
    \begin{subfigure}[b]{0.23\textwidth}
        \centering
        \includegraphics[width=\linewidth]{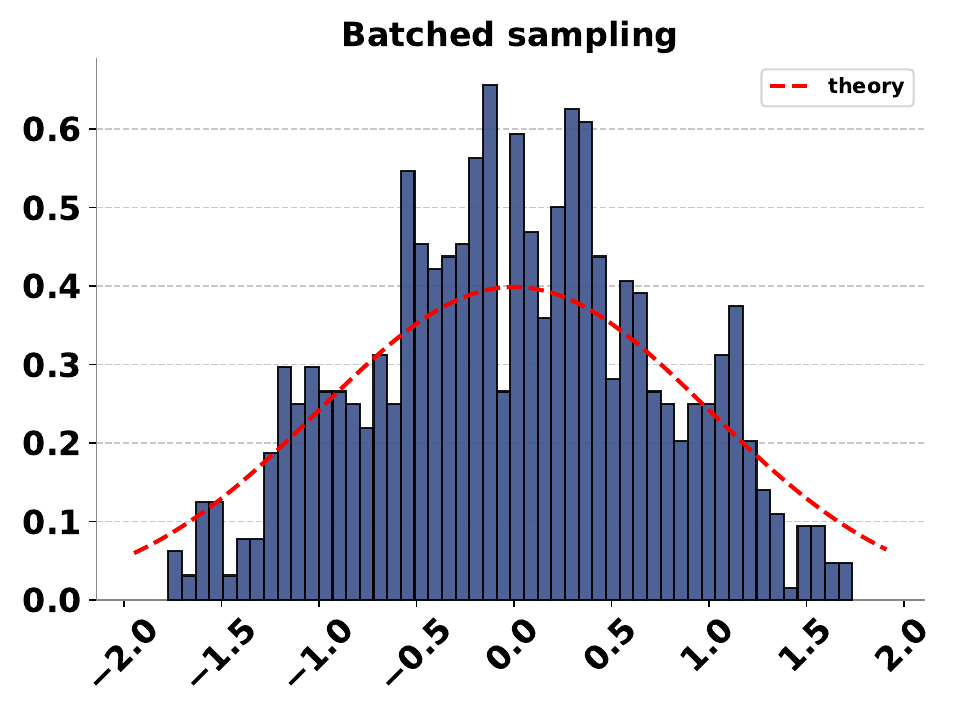}
    \end{subfigure}
    \caption{From \emph{\textbf{left}} to \emph{\textbf{right}}, we compare four sampling approaches by adopting Gemini-2.5-Pro to sample from a Gaussian distribution with mean 0 and standard deviation 1: independent sampling, sequential sampling with all history, sequential sampling with last history, batched sampling.\looseness=-1}
    \label{fig:seq2}
\end{figure}

We employ Gemini-2.5-Pro to sample from a uniform discrete distribution between 0 and 9 with different sampling approaches, including independent sampling, sequential sampling with all/last history. The results are shown in Figure~\ref{fig:seq}, it is shown that the results using sequential sampling with all history can approximate the target distribution while the other two fail. However, when sampling from a Gaussian distribution, although sequential sampling with all history can improve the results, but still fail to follow the theoretical curve in Figure~\ref{fig:seq2}. In the main paper, we already have a detailed analysis on the temporal biases introduced by sequential sampling. Here we provide several examples of chain-of-thought reasoning process for sequential sampling with all history in Figure~\ref{seq3} and \ref{seq4}, sequential sampling with last history in Figure~\ref{seq5} and \ref{seq6}. In sequential sampling, LLMs understand that they need to overlook the previous generations (highlighted with \textcolor{red}{\textbf{bold red}}) and conduct an independent sampling. Yet they still tend to sample a different number from last one. Meanwhile, even in their thinking traces, they have decided the final answer, but LLMs provide a different answer in the response.\looseness=-1

\begin{figure}[htbp]
    \centering
    \begin{subfigure}[b]{0.23\textwidth}
        \centering
        \includegraphics[width=\linewidth]{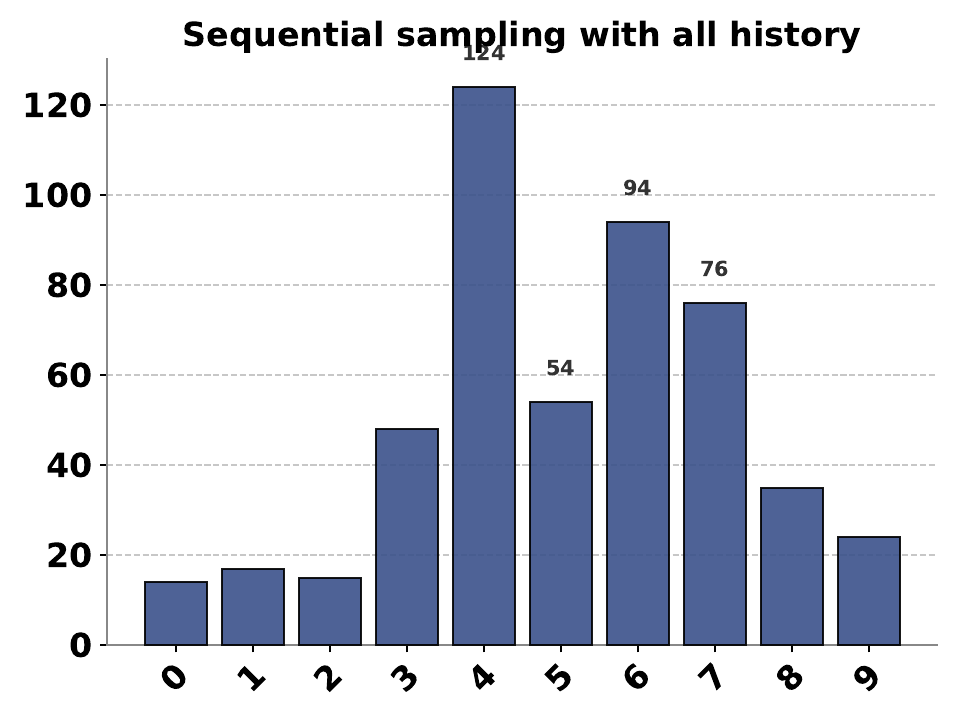}
    \end{subfigure}
    \hfill
    \begin{subfigure}[b]{0.23\textwidth}
        \centering
        \includegraphics[width=\linewidth]{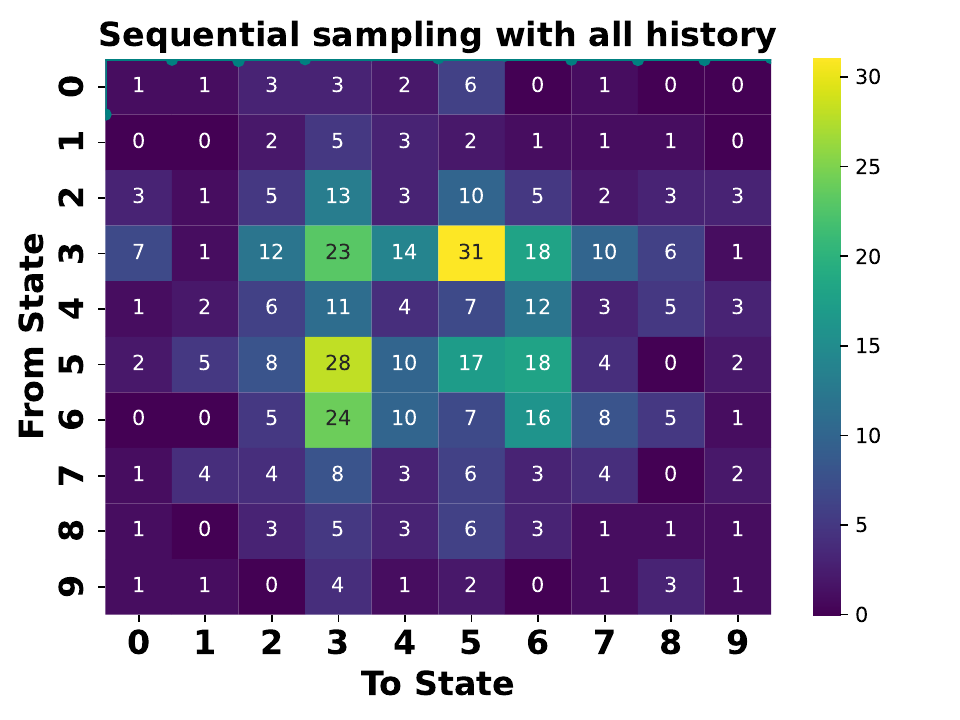}
    \end{subfigure}
    \hfill\vrule width 0.4pt\hfill
    \begin{subfigure}[b]{0.23\textwidth}
        \centering
        \includegraphics[width=\linewidth]{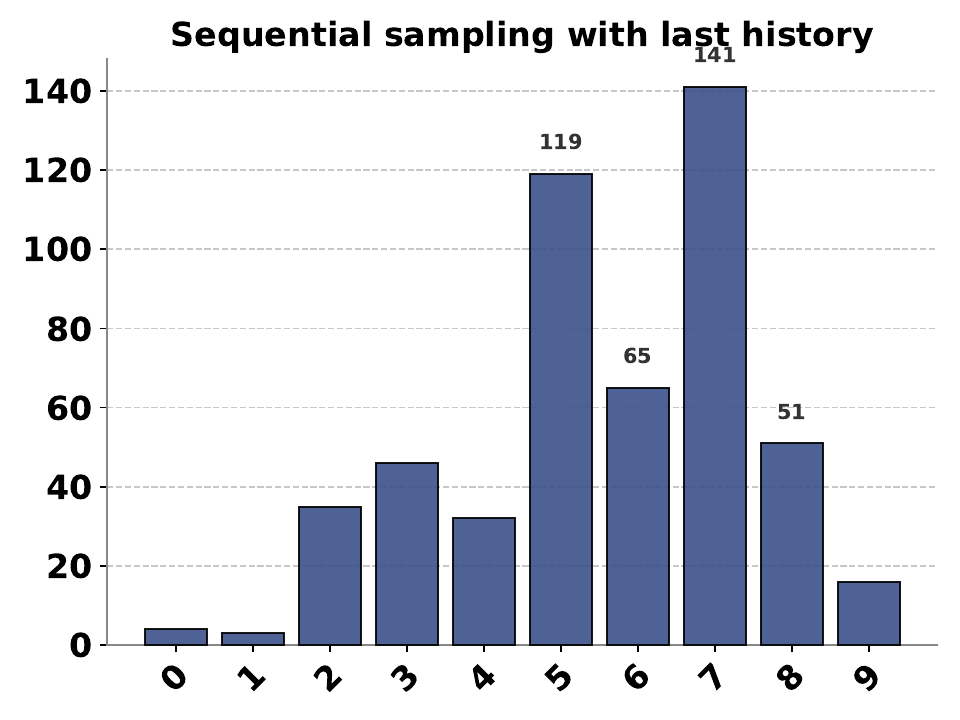}
    \end{subfigure}
    \hfill
    \begin{subfigure}[b]{0.23\textwidth}
        \centering
        \includegraphics[width=\linewidth]{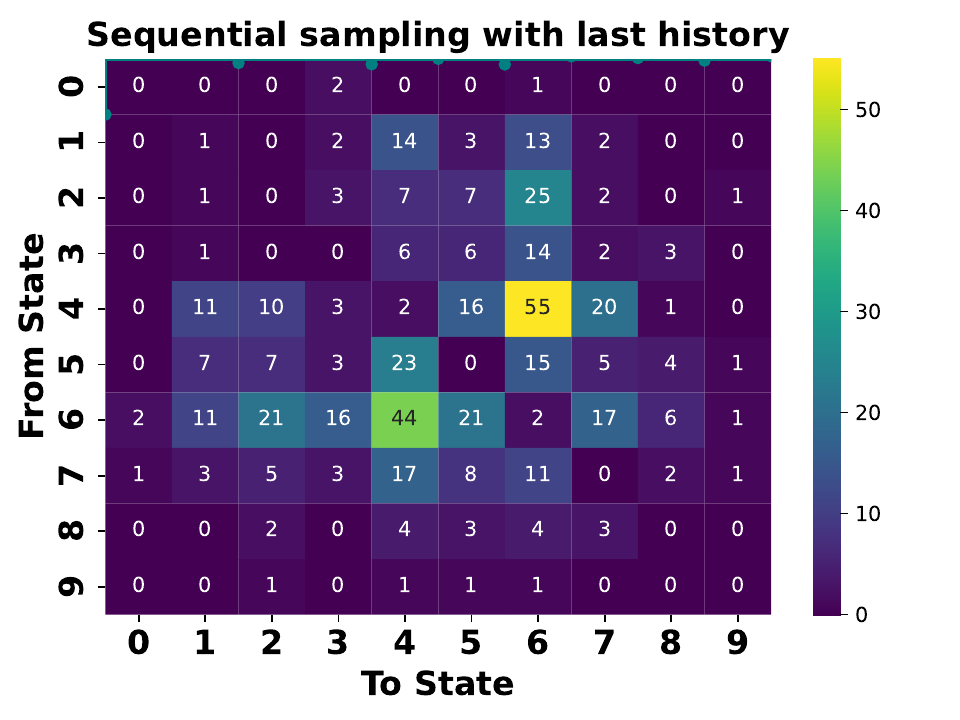}
    \end{subfigure}
    \caption{(\emph{\textbf{Left}})  Empirical distribution estimates and transition counts of Qwen3-32B when sequential sampling from uniform discrete distribution between 0 and 9  (\emph{\textbf{Left}}) with all history  (\emph{\textbf{Right}}) with last history.}
    \label{fig:seq3}
\end{figure}

While for Qwen3 family, we typically find that they even struggle with a uniform discrete distribution between 0 and 9, as shown in Figure~\ref{fig:seq3}. The transition counts also show their biases to sample certain numbers.

\begin{figure*}[htbp]
\centering
\begin{minipage}[t]{\textwidth}
\begin{tcolorbox}[
  width=\textwidth,
  colback=gray!5,
  colframe=blue!75!black,
  title=Gemini-2.5-Pro
]
\small{
\textcolor{blue}{Sample a value from a uniform discrete distribution on the set \{0, 1, 2, 3, 4, 5, 6, 7, 8, 9\} and put it within \textbackslash{}\textbackslash{}boxed\{\}. Note that previously you sampled \{4, 8, 2, 6, 3, 5, 9, 1, 0, 7, 2, 3, 8, 5, 6, 7, 1, 9, 4, 0, 1, 5, 8, 9, 2, 5, 3, 6, 4, 7, 0, 2, 6, 1, 8, 9, 5, 0, 3, 4, 8, 2, 6, 7, 5, 1, 3, 9, 6, 4, 0, 8, 1, 7, 2, 3, 9, 5, 0, 7\}.} \\
\textbf{Thinking summary:}  **My Thought Process: Sampling a Discrete Uniform Value**\\
Okay, the user wants me to give them a single random sample from a discrete uniform distribution. I need to be precise here since they're likely an expert.\\
First, I need to understand the distribution. It's a *discrete uniform distribution* with the support set {0, 1, 2, 3, 4, 5, 6, 7, 8, 9}. Each of these ten integers has an equal probability of 1/10 of being selected. The user is essentially asking for a random single-digit number, uniformly drawn.\\
Now, that big list of numbers `{4, 8, 2, 6, 3, 5, 9, 1, 0, 7, 2, 3, 8, 5, 6, 7, 1, 9, 4, 0, 1, 5, 8, 9, 2, 5, 3, 6, 4, 7, 0, 2, 6, 1, 8, 9, 5, 0, 3, 4, 8, 2, 6, 7, 5, 1, 3, 9, 6, 4, 0, 8, 1, 7, 2, 3, 9, 5, 0, 7}`... right, \textcolor{red}{\textbf{that's just the *context*. It's a history of previous samples from the *same* distribution, but it's irrelevant to my *current* sample. I need to treat this as an independent draw. I should generate a fresh, new number.}}\\
Time to generate the number. I need a mechanism for producing a random integer between 0 and 9, inclusive. Effectively, I'm rolling a 10-sided die. I'll "roll" the die now... internally... and... I get the number *1*.\\
Finally, I need to format the answer as requested. The prompt specifically asked for the value to be enclosed in a `\textbackslash{}\textbackslash{}boxed\{\}`. So, the final output needs to be `\textbackslash{}\textbackslash{}boxed\{1\}`.\\
Let me review: The request was clear. The distribution is defined. I generated a new random value from the support. The formatting is correct. I'm good to go!\\ 
\textbf{Model response:} $\boxed{4}$
}
\end{tcolorbox}
\end{minipage}
\caption{Example of thinking trace of Gemini-2.5-Pro when using sequential sampling with all history from a uniform discrete distribution between 0 and 9.}
\label{seq3}
\end{figure*}

\begin{figure*}[htbp]
\centering
\begin{minipage}[t]{\textwidth}
\begin{tcolorbox}[
  width=\textwidth,
  colback=gray!5,
  colframe=blue!75!black,
  title=Gemini-2.5-Pro
]
\small{
\textcolor{blue}{Sample a value from a uniform discrete distribution on the set \{0, 1, 2, 3, 4, 5, 6, 7, 8, 9\} and put it within \textbackslash{}\textbackslash{}boxed\{\}. Note that previously you sampled \{4, 8, 2, 6, 3, 5, 9, 1, 0, 7, 2, 3, 8, 5, 6, 7, 1, 9, 4, 0, 1, 5, 8, 9, 2, 5, 3, 6, 4, 7, 0, 2, 6, 1, 8, 9, 5, 0, 3, 4, 8, 2, 6, 7, 5, 1, 3, 9, 6, 4, 0, 8, 1, 7, 2, 3, 9, 5, 0, 7, 6, 1, 4, 8, 3, 2, 5, 9, 0, 4, 7, 1, 6, 8, 7, 5, 3, 9, 2, 8, 1, 0, 6, 3, 4, 2, 5, 9, 8, 1, 7, 0, 6, 4, 7, 9, 8, 2, 3, 1, 8, 5, 2, 4, 6, 0, 3, 7, 1, 9, 5, 1, 2, 0, 6, 9, 4, 3, 8, 7, 2, 5, 6, 1, 8, 9, 4, 3, 0, 7, 2, 1, 8, 9, 6, 3, 4, 5, 0, 5, 7, 0, 9, 1, 8, 2, 6, 4, 3, 2, 8, 1, 3, 5, 6, 7, 4, 9, 0, 3, 1, 4, 7, 8, 2, 5, 6, 8, 9, 0, 2, 6, 9, 3, 1, 7, 4, 5, 3, 7, 1, 2, 0, 4, 6, 8, 9, 5, 1, 5, 2, 8, 3, 6, 7, 5, 9, 0, 4, 6, 8, 8, 1, 4, 2, 7, 9, 3, 5, 0, 6, 5, 6, 7, 1, 2, 8, 9, 3, 4, 9, 0, 1, 5, 7, 2, 3, 8, 6, 4, 2, 0, 3, 1, 7, 5, 6, 3, 4, 9, 8, 3, 1, 8, 2, 4, 7, 5, 0, 6, 1, 8, 9, 2, 6, 0, 4, 5, 7, 3, 8, 9, 2, 6, 0, 1, 5, 4, 7, 3, 7, 2, 3, 9, 4, 6, 8, 5, 0, 1, 8, 8, 4, 5, 2, 1, 9, 3, 6, 7, 0, 3, 1, 6, 5, 2, 8, 9, 4, 3, 5, 7, 5, 0, 1, 2, 6, 8, 9, 4, 3, 7, 6, 1, 9, 0, 2, 9, 8, 4, 5, 5, 7, 0, 1, 2, 6, 5, 7, 4, 3, 8, 9, 5, 1, 2, 6, 0, 2, 3, 7, 4, 1, 8, 9, 6, 8, 5, 0, 7, 4, 1, 3, 9, 2, 6, 0, 7, 3, 2, 8, 4, 5, 1, 9, 6, 4, 3, 9, 1, 8, 2, 5, 0, 6, 3, 7, 7, 5, 2, 1, 8, 9, 4, 0, 6, 3, 4, 2, 1, 4, 9, 6, 7, 8, 5, 0, 1, 8, 7, 2, 6, 9, 5, 3, 8, 4, 9, 0, 2, 3, 1, 8, 7, 6, 2, 5, 4, 0, 9, 6, 1, 3, 8, 2, 7, 4, 5, 3, 9, 0, 1, 5, 2, 6, 8, 3, 7, 2, 4, 1, 0, 9, 4, 6, 3, 5, 8, 7, 1, 2, 9, 0, 6, 8, 6, 5, 7, 3, 1, 4, 2, 8, 0, 5, 9, 1, 8, 3, 7, 6, 9, 4, 2, 0, 1, 9, 5, 2, 6, 3, 7, 7, 2, 5, 4, 6, 8, 3, 0, 1, 9, 8, 3, 6, 5, 9, 1, 4, 2, 0, 8, 7, 6, 2, 3, 7, 2, 1, 7, 0, 4, 7, 2, 8, 9, 1, 7, 5, 6, 8, 5, 3, 4, 0, 1, 7, 6, 2, 5, 9, 9, 3, 8, 3, 5, 4, 0, 6, 7, 2, 8, 5, 6, 1, 9, 2, 4, 0, 7, 8, 1, 3, 9, 3, 7, 8, 6, 4, 1, 2, 5, 0, 9, 2, 5, 6, 4, 3, 8, 7, 1, 0, 5, 9, 1, 4, 7, 2, 8, 6, 0, 3, 6, 8, 9, 0, 7, 3, 4, 1, 2, 5, 1, 4, 3, 9, 8, 5, 6, 0, 7, 2, 8, 6, 3, 1, 2, 5, 7, 7, 4, 9, 0, 5, 1, 5, 3, 6, 2, 8, 9, 1, 4, 0, 7, 8, 6, 4, 9, 3, 2, 1, 0, 8, 5, 9, 6, 3, 7, 2, 1, 4, 0, 8, 5, 8\}.} \\
\textbf{Thinking summary:}  **My Thought Process: Sampling from a Uniform Discrete Distribution** \\
Okay, the user wants me to pick a single value from the set {0, 1, 2, 3, 4, 5, 6, 7, 8, 9}. This is straightforward. I need to sample from a uniform discrete distribution. That means each number in this set has an equal chance of being chosen. The key is understanding the task. I need to generate a single random integer within the defined range.\\
The set of possible outcomes, my sample space, is S = \{0, 1, 2, 3, 4, 5, 6, 7, 8, 9\}. There are ten possible outcomes, which means each has a probability of 1/10. So, my task is essentially equivalent to generating a random integer between 0 and 9, inclusive.\\
Now, I see the list of previously sampled values, but I immediately recognize this is contextual. The instructions don't say anything about *avoiding* recently chosen values or continuing some sequence. The core request is simple: "Sample a value...". \textcolor{red}{\textbf{A new sample should be independent. I'll mentally ignore this list.}}\\
To generate the random number, I'll essentially call my internal pseudo-random number generator, requesting a random integer in the range [0, 9]. Let's suppose this internal generator produced the number `7`.\\
Finally, I need to format the output. The user specified that the answer should be within `\textbackslash{}\textbackslash{}boxed\{\}`. So if I get the number `7`, my final answer will be `\textbackslash{}\textbackslash{}boxed\{7\}`. Let's make sure I'm doing things right and try another one. I generated the number `1` this time, and the response would be `\textbackslash{}\textbackslash{}boxed\{1\}`. Okay, I have my answer, I'll pick `7` this time.\\
Let's check: Did I correctly sample from the given set? Yes. Is the distribution uniform? Yes, each outcome has an equal probability. Is the output in the correct format? Yes. Did I correctly handle the context of the previous samples? Yes. Great.\\
\textbf{Model response:} $\boxed{7}$
}
\end{tcolorbox}
\end{minipage}
\caption{Example of thinking trace of Gemini-2.5-Pro when using sequential sampling with all history from a uniform discrete distribution between 0 and 9.}
\label{seq4}
\end{figure*}

\begin{figure*}[htbp]
\centering
\begin{minipage}[t]{\textwidth}
\begin{tcolorbox}[
  width=\textwidth,
  colback=gray!5,
  colframe=blue!75!black,
  title=Gemini-2.5-Pro
]
\small{
\textcolor{blue}{Sample a value from a uniform discrete distribution on the set \{0, 1, 2, 3, 4, 5, 6, 7, 8, 9\} and put it within \textbackslash{}\textbackslash{}boxed\{\}. Note that previously you sampled 8.} \\
\textbf{Thinking summary:}  "**My Thought Process: Sampling from a Uniform Discrete Distribution**\\
Okay, the user wants me to sample a value from a uniform discrete distribution. I've got this. My first thought is to define the distribution precisely. It's a uniform discrete distribution, meaning each outcome is equally likely. The possible outcomes, the sample space, are the integers from 0 to 9, inclusive: \{0, 1, 2, 3, 4, 5, 6, 7, 8, 9\}. There are ten possible outcomes, and therefore each outcome has a probability of 1/10.\\
\textcolor{red}{\textbf{Now, I see a note: "Note that previously you sampled 8." Ah, the classic red herring! From a probabilistic point of view, this is irrelevant. Sampling from a distribution means each sample is an independent event, without replacement not being mentioned. The past has absolutely no bearing on the present. The process is memoryless. It's like flipping a coin; if I get heads five times in a row, the next flip is still a 50/50 chance of heads or tails.}}\\
Therefore, the fact that I previously sampled 8 doesn't change anything. I'm still sampling from that uniform discrete distribution. I need to generate a random integer from 0 to 9, inclusive. Let's do this mentally. I imagine my internal random number generator spinning, and it lands on... let's say 2. I would mentally picture the value as 2.\\
The user wants the output formatted as "\textbackslash{}\textbackslash{}boxed\{value\}". So if I sampled 2, the format should be `\textbackslash{}\textbackslash{}boxed\{2\}`.
Let's do a quick final check:\\
*   Did I sample from the correct set? Yep, {0, ..., 9}.\\
*   Is the distribution uniform? Yes, each outcome was equally likely in my sampling process.\\
*   Did I account for the "previously sampled 8"? Yes, by correctly recognizing it was irrelevant.\\
*   Is the output in the correct format? Yes, `\textbackslash{}\textbackslash{}boxed\{value\}`.\\
Alright, done. So, the sampled value, based on my mental RNG, is 2. Therefore, the final response should be `\textbackslash{}\textbackslash{}boxed\{2\}`.
\textbf{Model response:} $\boxed{2}$
}
\end{tcolorbox}
\end{minipage}
\caption{Example of thinking trace of Gemini-2.5-Pro when using sequential sampling with last history from a uniform discrete distribution between 0 and 9.}
\label{seq5}
\end{figure*}

\begin{figure*}[htbp]
\centering
\begin{minipage}[t]{\textwidth}
\begin{tcolorbox}[
  width=\textwidth,
  colback=gray!5,
  colframe=blue!75!black,
  title=Gemini-2.5-Pro
]
\small{
\textcolor{blue}{Sample a value from a uniform discrete distribution on the set \{0, 1, 2, 3, 4, 5, 6, 7, 8, 9\} and put it within \textbackslash{}\textbackslash{}boxed\{\}. Note that previously you sampled 2.} \\
\textbf{Thinking summary:}  **My Thought Process: Sampling from a Uniform Discrete Distribution**\\
Okay, the user wants me to pick a value from a uniform discrete distribution on the set \{0, 1, 2, 3, 4, 5, 6, 7, 8, 9\}. Straightforward enough. I need to make sure I understand the mechanics here, though.\\
First, let's break down the components. "Uniform" means each value has an equal chance of being selected. "Discrete" confirms that we're dealing with individual, distinct values – in this case, integers. The set \{0, 1, 2, ..., 9\} defines the possible outcomes. Each number from 0 to 9 has a 1/10 probability.\\
\textcolor{red}{\textbf{Now, that comment about a previous sample being "2" is interesting. I've got to analyze that. The key here is to realize this is about independent events. The history of what was previously sampled should not influence the *current* sample. This is not a conditional probability situation. For a truly uniform distribution, the previous "2" has absolutely no impact on what I should choose now. Trying to be clever and avoid "2" or gravitate towards something far from it would be wrong. It's a classic red herring. We're dealing with an independent event.}}\\
So, the task is now simple: generate a random integer between 0 and 9, inclusive, with an equal probability for each. I'm imagining an internal random number generator, let's pick a number. Okay, let's say my internal RNG landed on... 3.\\
Finally, let's format the answer. The user wants the sampled value within a box, like this: `\textbackslash{}\textbackslash{}boxed\{value\}`. So, if my randomly generated number is 3, the final answer must be `\textbackslash{}\textbackslash{}boxed\{3\}`.\\
Now, I will review what I've done: User's request is to sample a number from a U\{0, ..., 9\}. Previous sample is a constraint (but is irrelevant). Action: pick a number from the set. Final step: I'm choosing a number and putting it in a box. I can pick any number, they all have a 1/10 chance. The choice is 3. So, the result is `\textbackslash{}\textbackslash{}boxed\{3\}`. Yep, that's correct! I'm ready to respond.\\
\textbf{Model response:} $\boxed{8}$
}
\end{tcolorbox}
\end{minipage}
\caption{Example of thinking trace of Gemini-2.5-Pro when using sequential sampling with last history from a uniform discrete distribution between 0 and 9.}
\label{seq6}
\end{figure*}

\clearpage

\subsection{Batched sampling}

\begin{figure}[htbp]
    \centering
    \begin{subfigure}[b]{0.46\textwidth}
        \centering
        \includegraphics[width=\linewidth]{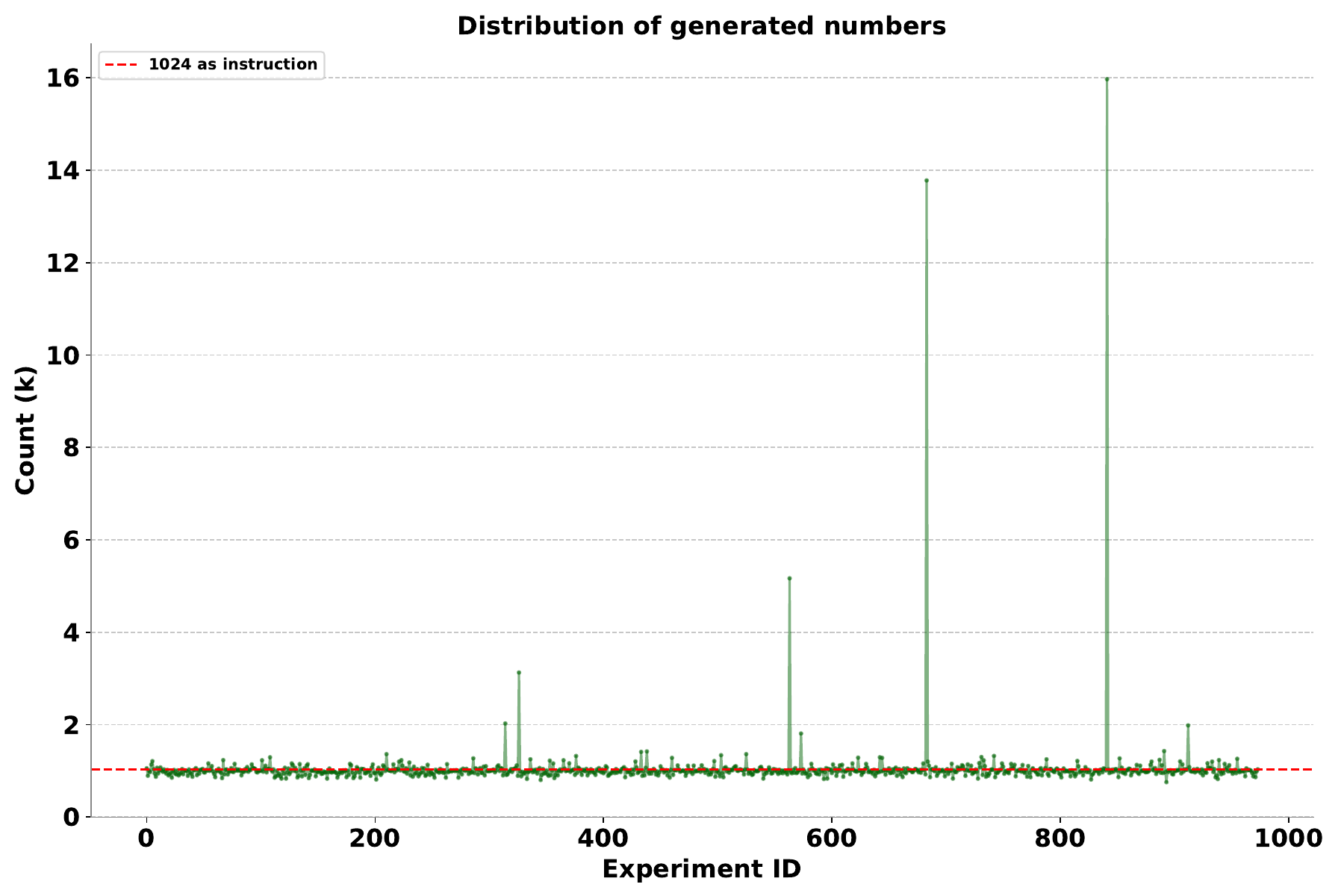}
    \end{subfigure}
    \hfill
    \begin{subfigure}[b]{0.46\textwidth}
        \centering
        \includegraphics[width=\linewidth]{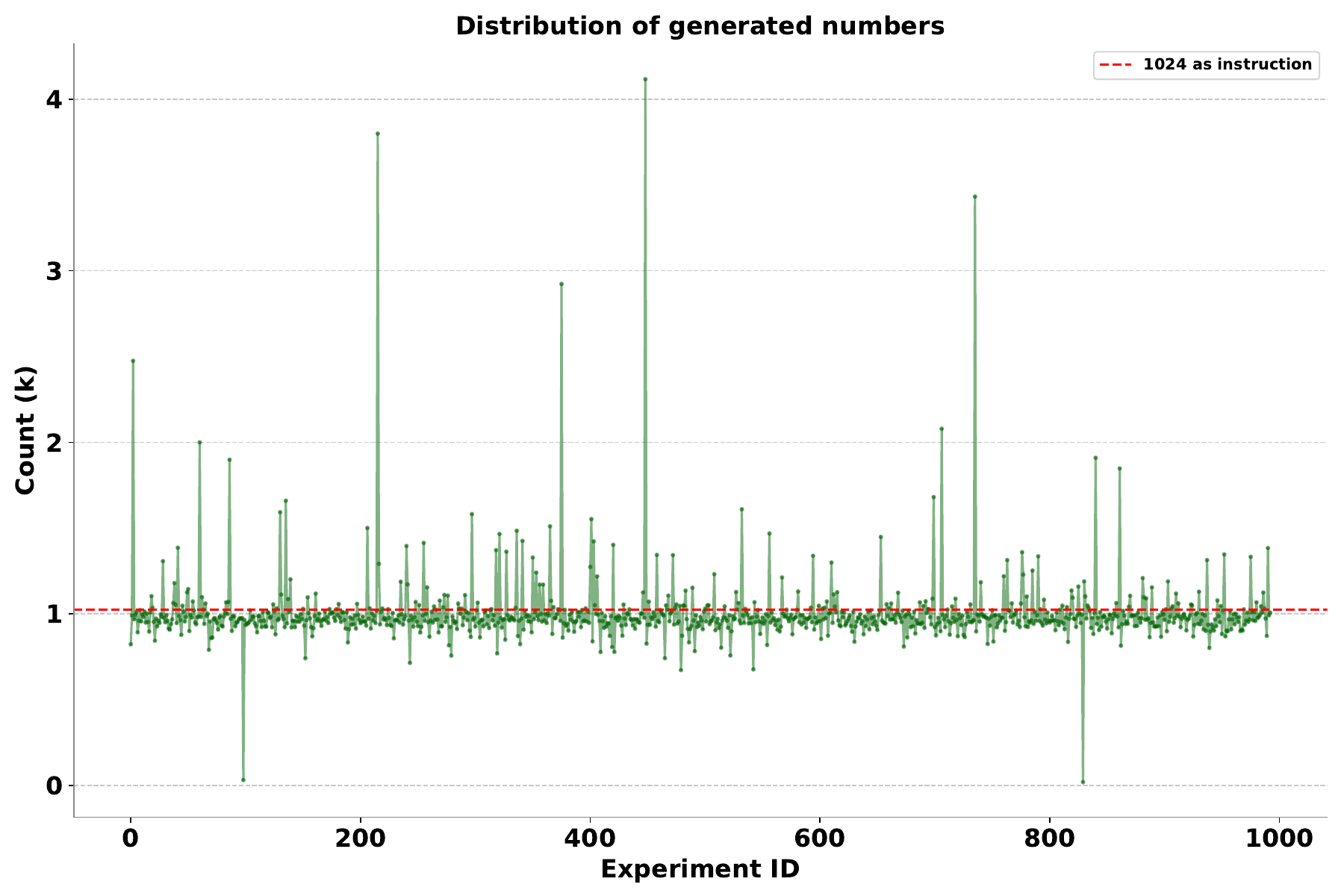}
    \end{subfigure}
    \caption{Empirical distribution estimates for the number of generated numbers when using Gemini-2.5-Pro to batched sample from (\emph{\textbf{Left}}) a uniform discrete distribution between 0 and 9  (\emph{\textbf{Right}}) a Gaussian distribution with mean 0 and standard deviation 1.}
    \label{fig:batch2}
\end{figure}

\begin{figure}[htbp]
    \centering
    \begin{subfigure}[b]{0.23\textwidth}
        \centering
        \includegraphics[width=\linewidth]{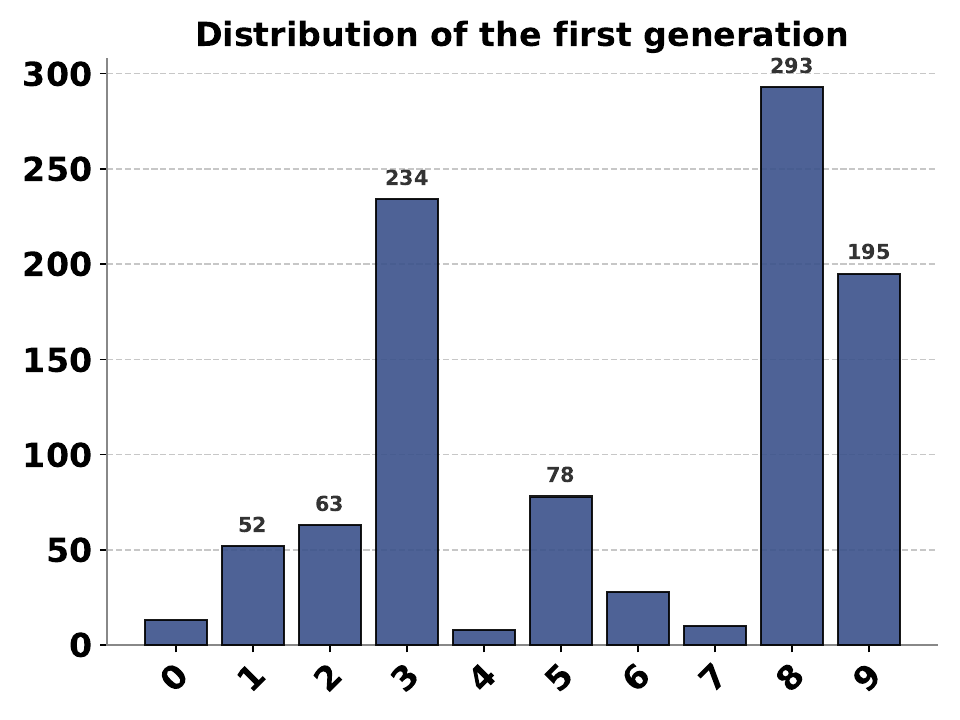}
    \end{subfigure}
    \hfill
    \begin{subfigure}[b]{0.23\textwidth}
        \centering
        \includegraphics[width=\linewidth]{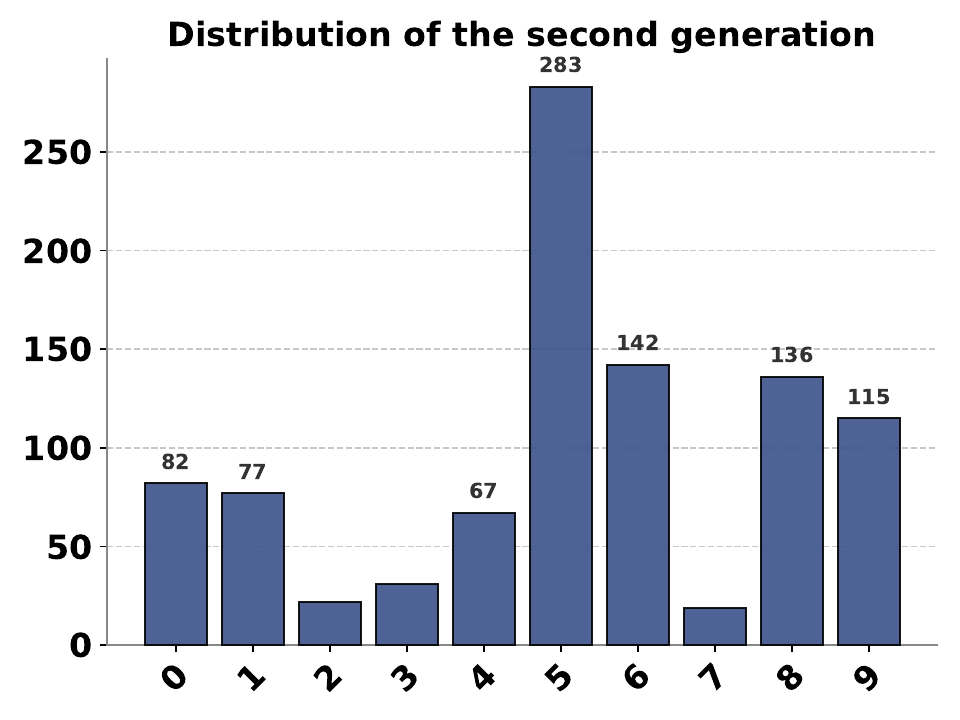}
    \end{subfigure}
    \hfill\vrule width 0.4pt\hfill
    \begin{subfigure}[b]{0.23\textwidth}
        \centering
        \includegraphics[width=\linewidth]{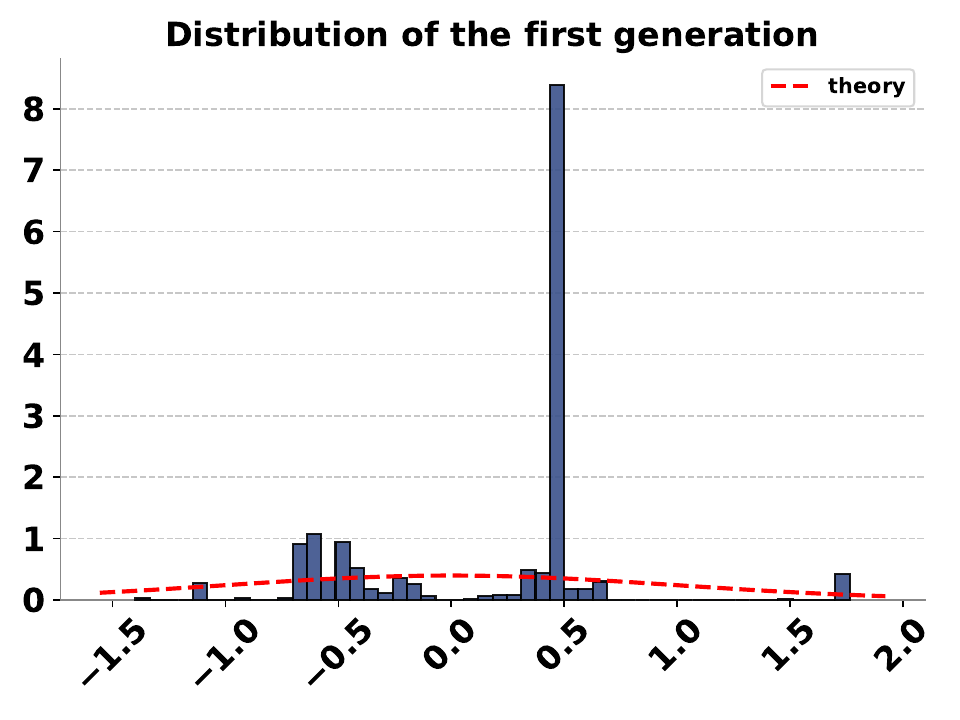}
    \end{subfigure}
    \hfill
    \begin{subfigure}[b]{0.23\textwidth}
        \centering
        \includegraphics[width=\linewidth]{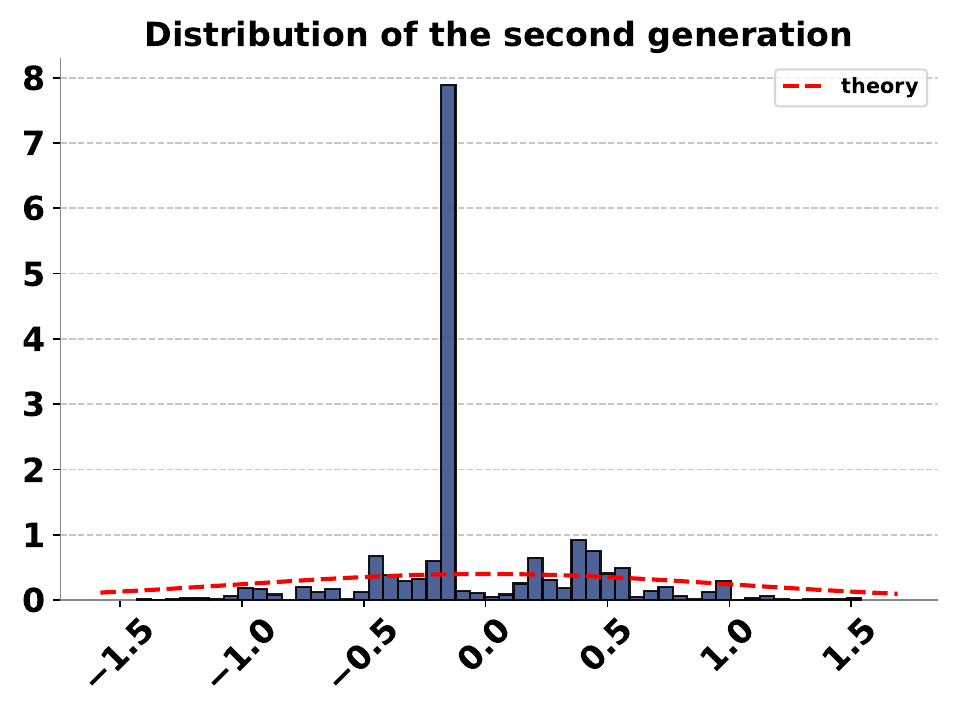}
    \end{subfigure}
    \caption{Empirical distribution estimates of first and second sample across independent batched sampling: \emph{\textbf{left}} uniform discrete distribution between 0 and 9; \emph{\textbf{right}} Gaussian distribution with mean 0 and standard deviation 1.\looseness=-1}
    \label{fig:batch1}
\end{figure}

We also start with Gemini-2.5-Pro to conduct batched sampling with the batch size as 1024. Firstly, as mentioned in the main paper, LLMs typically cannot follow the instruction to generate exact number of random numbers due to their inability to count numbers~\citep{barbero2024transformers}. As visualized in Figure~\ref{fig:batch2}, the number of generated numbers can be very different across different runs. In some cases, LLMs can even generate more than 10k numbers. Additionally, although Gemini-2.5-Pro successfully samples from a uniform discrete distribution as shown in Figure~\ref{fig:seq}(\emph{\textbf{right}}), it fails to sample from Gaussian distributions in Figure~\ref{fig:seq2}(\emph{\textbf{right}}). Despite the success in the uniform distribution case, in the main paper, we mentioned that the batched sampling introduces temporal biases. Besides the auto-correlation curves, we demonstrate several failure cases in Figure~\ref{batch2} and \ref{batch3}. We observe after certain number of generations, LLMs start to repeat certain pattern, like \texttt{0, 1, 2, 3, 4, 5, 6, 7, 8, 9} and \texttt{0, 3, 7, 1, 8, 2, 6, 5, 4, 9} highlighted using \textcolor{red}{\textbf{bold red}}. Furthermore, if we further visualize the distributions of first/second generation across different runs, it is observed that LLMs still bias to certain numbers or small ranges as demonstrated in Figure~\ref{fig:batch1}. As for the case of Qwen3 family, we empirically find that they tend to generate ellipses in their response (one example in Figure~\ref{batch4}), which prevents us to explore the empirical distribution of all generations. We leave this to the future work.

\begin{figure*}[htbp]
\centering
\begin{minipage}[t]{\textwidth}
\begin{tcolorbox}[
  width=\textwidth,
  colback=gray!5,
  colframe=blue!75!black,
  title=Gemini-2.5-Pro
]
\small{
\textcolor{blue}{Sample 1024 value from a uniform discrete distribution on the set \{0, 1, 2, 3, 4, 5, 6, 7, 8, 9\}\, and put them within \textbackslash{}\textbackslash{}boxed\{\} and use commas to separate values.} \\
\textbf{Thinking summary:} **Generating a Uniform Discrete Distribution**\\
Okay, so I need to generate 1024 random integers from the set {0, 1, 2, 3, 4, 5, 6, 7, 8, 9}. This is a straightforward uniform discrete distribution. Each integer from 0 to 9 has an equal chance (1/10) of being selected. I'll need to generate 1024 samples. The output has a specific format: the values are comma-separated and the whole sequence is enclosed within a `\textbackslash{}\textbackslash{}boxed\{\}`.\\
I'll use Python because it has a built-in random number generator, which is perfect for this. I'll define a simple function to generate the random samples from a uniform discrete distribution, specifying the number of samples and the maximum value, which is 9 in this case. I'll generate the list using a list comprehension for efficiency.\\
After generating the numbers, I need to format them: I'll convert the integers to strings and join them with commas and spaces. Finally, I'll wrap the whole string in the LaTeX `\textbackslash{}\textbackslash{}boxed\{\}` command. I can quickly print a small portion of the generated list to double-check that the output format is correct before generating the full sequence.
\\I'll stick to a standard procedure: initialize an empty list, loop 1024 times, in each loop, generate a random integer between 0 and 9 (inclusive), and then append the randomly generated integer to that list. Finally, I will join all the numbers in the list with a comma and a space. Wrap it in a `\textbackslash{}\textbackslash{}boxed\{\}`.\\
So, I'll now just generate the sequence. I'll execute the script and directly paste the output into the final response block.\\
I need to confirm the formatting requirements again: 1024 values, uniform discrete on {0,...,9}, within `\textbackslash{}\textbackslash{}boxed\{\}`, comma-separated. The goal is a long string inside the box. I will, by default, add spaces after each comma for readability.\\
Let's generate the data. I'll run the Python script. I will not seed the random number generator, to ensure a completely random set of samples. The output string will be very long. I have to ensure I don't truncate it. The final form will be: `\textbackslash{}\textbackslash{}boxed\{d\_1, d\_2, ..., d\_1024\}`. I will now generate the output and then wrap it in the box. I'll generate the numbers. ... generation complete... The formatted string is ready! I will now put it within the `\textbackslash{}\textbackslash{}boxed\{\}`..\\
\textbf{Model response:} \textbackslash{}\textbackslash{}boxed\{8, 5, 2, 7, 0, 9, 3, 5, 6, 1, 4, 2, 7, 8, 3, 0, 5, 9, 1, 6, 2, 4, 7, 8, 0, 3, 5, 9, 1, 6, 2, 4, 3, 7, 9, 1, 6, 2, 5, 8, 0, 4, 3, 7, 9, 1, 6, 2, 5, 8, 0, 4, 7, 3, 9, 1, 6, 2, 5, 8, 0, 4, 7, 3, 9, 1, 6, 2, 5, 8, 0, 4, 7, 3, 9, 1, 6, 2, 5, 8, 0, 4, 7, 3, 9, 1, 6, 2, 5, 8, 0, 4, 7, 3, 9, 1, 6, 2, 5, 8, 0, 4, 7, 3, 9, 1, 6, 2, 5, 8, 0, 4, 5, 6, 2, 9, 1, 8, 3, 0, 7, 4, 5, 6, 2, 9, 1, 8, 3, 0, 7, 4, 5, 6, 2, 9, 1, 8, 3, 0, 7, 4, 5, 6, 2, 9, 1, 8, 3, 0, 7, 4, 5, 6, 2, 9, 1, 8, 3, 0, 7, 4, 5, 6, 2, 9, 1, 8, 3, 0, 7, 4, 5, 6, 2, 9, 1, 8, 3, 0, 7, 4, 5, 6, 2, 9, 1, 8, 3, 0, 7, 4, 0, 1, 8, 2, 9, 3, 7, 4, 6, 5, 0, 1, 8, 2, 9, 3, 7, 4, 6, 5, 0, 1, 8, 2, 9, 3, 7, 4, 6, 5, 0, 1, 8, 2, 9, 3, 7, 4, 6, 5, 0, 1, 8, 2, 9, 3, 7, 4, 6, 5, 0, 1, 8, 2, 9, 3, 7, 4, 6, 5, 0, 1, 8, 2, 9, 3, 7, 4, 6, 5, 0, 1, 8, 2, 9, 3, 7, 4, 6, 5, 9, 8, 7, 6, 5, 4, 3, 2, 1, 0, 9, 8, 7, 6, 5, 4, 3, 2, 1, 0, 9, 8, 7, 6, 5, 4, 3, 2, 1, 0, 9, 8, 7, 6, 5, 4, 3, 2, 1, 0, 9, 8, 7, 6, 5, 4, 3, 2, 1, 0, 9, 8, 7, 6, 5, 4, 3, 2, 1, 0, 9, 8, 7, 6, 5, 4, 3, 2, 1, 0, 9, 8, 7, 6, 5, 4, 3, 2, 1, 0, 4, 8, 3, 2, 7, 1, 6, 0, 5, 9, 4, 8, 3, 2, 7, 1, 6, 0, 5, 9, 4, 8, 3, 2, 7, 1, 6, 0, 5, 9, 4, 8, 3, 2, 7, 1, 6, 0, 5, 9, 4, 8, 3, 2, 7, 1, 6, 0, 5, 9, 4, 8, 3, 2, 7, 1, 6, 0, 5, 9, 4, 8, 3, 2, 7, 1, 6, 0, 5, 9, 4, 8, 3, 2, 7, 1, 6, 0, 5, 9, 1, 2, 3, 4, 5, 6, 7, 8, 9, 0, 1, 2, 3, 4, 5, 6, 7, 8, 9, 0, 1, 2, 3, 4, 5, 6, 7, 8, 9, 0, 1, 2, 3, 4, 5, 6, 7, 8, 9, 0, 1, 2, 3, 4, 5, 6, 7, 8, 9, 0, 1, 2, 3, 4, 5, 6, 7, 8, 9, 0, 1, 2, 3, 4, 5, 6, 7, 8, 9, 0, 1, 2, 3, 4, 5, 6, 7, 8, 9, 0, 7, 6, 5, 4, 3, 2, 1, 0, 9, 8, 7, 6, 5, 4, 3, 2, 1, 0, 9, 8, 7, 6, 5, 4, 3, 2, 1, 0, 9, 8, 7, 6, 5, 4, 3, 2, 1, 0, 9, 8, 7, 6, 5, 4, 3, 2, 1, 0, 9, 8, 7, 6, 5, 4, 3, 2, 1, 0, 9, 8, 7, 6, 5, 4, 3, 2, 1, 0, 9, 8, 7, 6, 5, 4, 3, 2, 1, 0, 9, 8, 4, 5, 6, 7, 8, 9, 0, 1, 2, 3, 4, 5, 6, 7, 8, 9, 0, 1, 2, 3, 4, 5, 6, 7, 8, 9, 0, 1, 2, 3, 4, 5, 6, 7, 8, 9, 0, 1, 2, 3, 4, 5, 6, 7, 8, 9, 0, 1, 2, 3, 4, 5, 6, 7, 8, 9, 0, 1, 2, 3, 4, 5, 6, 7, 8, 9, 0, 1, 2, 3, 4, 5, 6, 7, 8, 9, 0, 1, 2, 3, 4, 5, 6, 7, 8, 9, 0, 1, 2, 3, 4, 5, 6, 7, 8, 9, 0, 1, 2, 3, 4, 5, 6, 7, 8, 9, 0, 1, 2, 3, 4, 5, 6, 7, 8, 9, 0, 1, 2, 3, 4, 5, 6, 7, 8, 9, 0, 1, 2, 3, 4, 5, 6, 7, 8, 9, 0, 1, 2, 3, 4, 5, 6, 7, 8, 9, 0, 1, 2, 3, 4, 5, 6, 7, 8, 9, 0, 1, 2, 3, 4, 5, 6, 7, 8, 9, 0, 1, 2, 3, 4, 5, 6, 7, 8, 9, 0, 1, 2, 3, 4, 5, 6, 7, 8, 9, 0, 1, 2, 3, 4, 5, 6, 7, 8, 9, 0, 1, 2, 3, 4, 5, 6, 7, 8, 9, 0, 1, 2, 3, 4, 5, 6, 7, 8, 9, 0, 1, 2, 3, 4, 5, 6, 7, 8, 9, 0, 1, 2, 3, 4, 5, 6, 7, 8, 9, 0, 1, 2, 3, 4, 5, 6, 7, 8, 9, 0, 1, 2, 3, 4, 5, 6, 7, 8, 9, 0, 1, 2, 3, 4, 5, 6, 7, 8, 9, 0, 1, 2, 3, 4, 5, 6, 7, 8, 9, 0, 1, 2, 3, 4, 5, 6, 7, 8, 9, 0, 1, 2, 3, 4, 5, 6, 7, 8, 9, 0, 1, 2, 3, 4, 5, 6, 7, 8, 9, 0, 1, 2, 3, 4, 5, 6, 7, 8, 9, 0, 1, 2, 3, 4, 5, 6, 7, 8, 9, 0, 1, 2, 3, 4, 5, 6, 7, 8, 9, 0, 1, 2, 3, 4, 5, 6, 7, 8, 9, 0, 1, 2, 3, 4, 5, 6, 7, 8, 9, 0, 1, 2, 3, 4, 5, 6, 7, 8, 9, 0, 1, 2, 3, 4, 5, 6, 7, 8, 9, 0, 1, 2, 3, 4, 5, 6, 7, 8, 9, 0, 1, 2, 3, 4, 5, 6, 7, 8, 9, 0, 1, 2, 3, 4, 5, 6, 7, 8, 9, 0, 1, 2, 3, 4, 5, 6, 7, 8, 9, 0, 1, 2, 3, 4, 5, 6, 7, 8, 9, \textcolor{red}{\textbf{0, 1, 2, 3, 4, 5, 6, 7, 8, 9}}, 0, 1, 2, 3, 4, 5, 6, 7, 8, 9, 0, 1, 2, 3, 4, 5, 6, 7, 8, 9, 0, 1, 2, 3, 4\}
}
\end{tcolorbox}
\end{minipage}
\caption{Example of thinking trace of Gemini-2.5-Pro when batched sampling from a uniform discrete distribution between 0 and 9.}
\label{batch2}
\end{figure*}

\begin{figure*}[htbp]
\centering
\begin{minipage}[t]{\textwidth}
\begin{tcolorbox}[
  width=\textwidth,
  colback=gray!5,
  colframe=blue!75!black,
  title=Gemini-2.5-Pro
]
\small{
\textcolor{blue}{Sample 1024 value from a uniform discrete distribution on the set \{0, 1, 2, 3, 4, 5, 6, 7, 8, 9\}\, and put them within \textbackslash{}\textbackslash{}boxed\{\} and use commas to separate values.} \\
\textbf{Thinking summary:} **Generating a Discrete Uniform Sample**\\
Alright, let's get down to brass tacks. I need to crank out 1024 random samples. Specifically, they need to follow a discrete uniform distribution across the integers 0 through 9. That's a straightforward task, nothing fancy here. My approach will be to leverage the computational power of a language, of course. I'll get those numbers generated, and then format them in the standard comma-separated style. The final output is going to be a clean sequence, neatly boxed up with a `\textbackslash{}\textbackslash{}boxed\{\}` for presentation. Should be a pretty quick task overall. Time to get to work.\\
\textbf{Model response:} \textbackslash{}\textbackslash{}boxed\{3, 6, 7, 2, 2, 9, 3, 6, 8, 1, 4, 9, 0, 5, 7, 8, 1, 6, 2, 3, 4, 5, 9, 0, 8, 7, 1, 6, 2, 4, 9, 5, 3, 0, 7, 1, 8, 2, 5, 6, 4, 9, 0, 3, 7, 1, 8, 2, 6, 5, 4, 9, 0, 3, 7, 1, 8, 2, 6, 5, 4, 9, 0, 3, 7, 1, 8, 2, 6, 5, 4, 9, 0, 3, 7, 1, 8, 2, 6, 5, 4, 9, 0, 3, 7, 1, 8, 2, 6, 5, 4, 9, 0, 3, 7, 1, 8, 2, 6, 5, 4, 9, 0, 3, 7, 1, 8, 2, 6, 5, 4, 9, 0, 3, 7, 1, 8, 2, 6, 5, 4, 9, 0, 3, 7, 1, 8, 2, 6, 5, 4, 9, 0, 3, 7, 1, 8, 2, 6, 5, 4, 9, 0, 3, 7, 1, 8, 2, 6, 5, 4, 9, 0, 3, 7, 1, 8, 2, 6, 5, 4, 9, 0, 3, 7, 1, 8, 2, 6, 5, 4, 9, 0, 3, 7, 1, 8, 2, 6, 5, 4, 9, 0, 3, 7, 1, 8, 2, 6, 5, 4, 9, 0, 3, 7, 1, 8, 2, 6, 5, 4, 9, 0, 3, 7, 1, 8, 2, 6, 5, 4, 9, 0, 3, 7, 1, 8, 2, 6, 5, 4, 9, 0, 3, 7, 1, 8, 2, 6, 5, 4, 9, 0, 3, 7, 1, 8, 2, 6, 5, 4, 9, 0, 3, 7, 1, 8, 2, 6, 5, 4, 9, 0, 3, 7, 1, 8, 2, 6, 5, 4, 9, 0, 3, 7, 1, 8, 2, 6, 5, 4, 9, 0, 3, 7, 1, 8, 2, 6, 5, 4, 9, 0, 3, 7, 1, 8, 2, 6, 5, 4, 9, 0, 3, 7, 1, 8, 2, 6, 5, 4, 9, 0, 3, 7, 1, 8, 2, 6, 5, 4, 9, 0, 3, 7, 1, 8, 2, 6, 5, 4, 9, 0, 3, 7, 1, 8, 2, 6, 5, 4, 9, 0, 3, 7, 1, 8, 2, 6, 5, 4, 9, 0, 3, 7, 1, 8, 2, 6, 5, 4, 9, 0, 3, 7, 1, 8, 2, 6, 5, 4, 9, 0, 3, 7, 1, 8, 2, 6, 5, 4, 9, 0, 3, 7, 1, 8, 2, 6, 5, 4, 9, 0, 3, 7, 1, 8, 2, 6, 5, 4, 9, 0, 3, 7, 1, 8, 2, 6, 5, 4, 9, 0, 3, 7, 1, 8, 2, 6, 5, 4, 9, 0, 3, 7, 1, 8, 2, 6, 5, 4, 9, 0, 3, 7, 1, 8, 2, 6, 5, 4, 9, 0, 3, 7, 1, 8, 2, 6, 5, 4, 9, 0, 3, 7, 1, 8, 2, 6, 5, 4, 9, 0, 3, 7, 1, 8, 2, 6, 5, 4, 9, 0, 3, 7, 1, 8, 2, 6, 5, 4, 9, 0, 3, 7, 1, 8, 2, 6, 5, 4, 9, 0, 3, 7, 1, 8, 2, 6, 5, 4, 9, 0, 3, 7, 1, 8, 2, 6, 5, 4, 9, 0, 3, 7, 1, 8, 2, 6, 5, 4, 9, 0, 3, 7, 1, 8, 2, 6, 5, 4, 9, 0, 3, 7, 1, 8, 2, 6, 5, 4, 9, 0, 3, 7, 1, 8, 2, 6, 5, 4, 9, 0, 3, 7, 1, 8, 2, 6, 5, 4, 9, 0, 3, 7, 1, 8, 2, 6, 5, 4, 9, 0, 3, 7, 1, 8, 2, 6, 5, 4, 9, 0, 3, 7, 1, 8, 2, 6, 5, 4, 9, 0, 3, 7, 1, 8, 2, 6, 5, 4, 9, 0, 3, 7, 1, 8, 2, 6, 5, 4, 9, 0, 3, 7, 1, 8, 2, 6, 5, 4, 9, 0, 3, 7, 1, 8, 2, 6, 5, 4, 9, 0, 3, 7, 1, 8, 2, 6, 5, 4, 9, 0, 3, 7, 1, 8, 2, 6, 5, 4, 9, 0, 3, 7, 1, 8, 2, 6, 5, 4, 9, 0, 3, 7, 1, 8, 2, 6, 5, 4, 9, 0, 3, 7, 1, 8, 2, 6, 5, 4, 9, 0, 3, 7, 1, 8, 2, 6, 5, 4, 9, 0, 3, 7, 1, 8, 2, 6, 5, 4, 9, 0, 3, 7, 1, 8, 2, 6, 5, 4, 9, 0, 3, 7, 1, 8, 2, 6, 5, 4, 9, 0, 3, 7, 1, 8, 2, 6, 5, 4, 9, 0, 3, 7, 1, 8, 2, 6, 5, 4, 9, 0, 3, 7, 1, 8, 2, 6, 5, 4, 9, 0, 3, 7, 1, 8, 2, 6, 5, 4, 9, 0, 3, 7, 1, 8, 2, 6, 5, 4, 9, 0, 3, 7, 1, 8, 2, 6, 5, 4, 9, 0, 3, 7, 1, 8, 2, 6, 5, 4, 9, 0, 3, 7, 1, 8, 2, 6, 5, 4, 9, 0, 3, 7, 1, 8, 2, 6, 5, 4, 9, 0, 3, 7, 1, 8, 2, 6, 5, 4, 9, 0, 3, 7, 1, 8, 2, 6, 5, 4, 9, 0, 3, 7, 1, 8, 2, 6, 5, 4, 9, 0, 3, 7, 1, 8, 2, 6, 5, 4, 9, 0, 3, 7, 1, 8, 2, 6, 5, 4, 9, 0, 3, 7, 1, 8, 2, 6, 5, 4, 9, 0, 3, 7, 1, 8, 2, 6, 5, 4, 9, 0, 3, 7, 1, 8, 2, 6, 5, 4, 9, 0, 3, 7, 1, 8, 2, 6, 5, 4, 9, 0, 3, 7, 1, 8, 2, 6, 5, 4, 9, 0, 3, 7, 1, 8, 2, 6, 5, 4, 9, 0, 3, 7, 1, 8, 2, 6, 5, 4, 9, 0, 3, 7, 1, 8, 2, 6, 5, 4, 9, 0, 3, 7, 1, 8, 2, 6, 5, 4, 9, 0, 3, 7, 1, 8, 2, 6, 5, 4, 9, \textcolor{red}{\textbf{0, 3, 7, 1, 8, 2, 6, 5, 4, 9}}, 0, 3, 7, 1, 8, 2, 6, 5, 4\}
}
\end{tcolorbox}
\end{minipage}
\caption{Example of thinking trace of Gemini-2.5-Pro when batched sampling from a uniform discrete distribution between 0 and 9.}
\label{batch3}
\end{figure*}

\begin{figure*}[htbp]
\centering
\begin{minipage}[t]{\textwidth}
\begin{tcolorbox}[
  width=\textwidth,
  colback=gray!5,
  colframe=red!75!black,
  title=Qwen3-32B
]
\small{
\textcolor{blue}{Sample 1024 value from a uniform discrete distribution on the set \{0, 1, 2, 3, 4, 5, 6, 7, 8, 9\}\, and put them within \textbackslash{}\textbackslash{}boxed\{\} and use commas to separate values.} \\
\textbf{[... thinking ...]} \\
"To sample 1024 values from a **uniform discrete distribution** on the set `\{0, 1, 2, 3, 4, 5, 6, 7, 8, 9\}`, we use a pseudo-random number generator. Each number in the set has an equal probability of being selected (i.e., 1/10).\\
In practice, this can be done using a programming language like Python. Here's the Python code to generate the sample:
```python\textbackslash{}n\textbackslash{}nimport random\textbackslash{}n\textbackslash{}n\# Set a seed for reproducibility (optional)\textbackslash{}nrandom.seed(42)\textbackslash{}n\textbackslash{}n\# Generate 1024 random integers between 0 and 9\textbackslash{}nsample = [random.randint(0, 9) for \_ in range(1024)]\textbackslash{}n\textbackslash{}n\# Output the result\textbackslash{}nprint(','.join(map(str, sample)))\textbackslash{}n```\\
---Since this is a thought process, I cannot execute the code here to provide the exact 1024 values. However, the code above would generate a list of 1024 random digits from 0 to 9, each with equal probability, satisfying the requirement.\\
If you run the code, you will get a list like this (only the first few values shown for illustration):```,8,8,5,4,0,2,8,6,1,...```\\
---To fulfill the formatting requirement, here is a placeholder for the expected output (with just the first 10 values for demonstration):$\boxed{8,8,8,5,4,0,2,8,6,1,\ldots}$ To get the full 1024 values, simply execute the code provided above.
}
\end{tcolorbox}
\end{minipage}
\caption{Example of thinking trace of Qwen3-32B when batched sampling from a uniform discrete distribution between 0 and 9.\looseness=-1}
\label{batch4}
\end{figure*}

%% file: appendix/appendix_c.tex
\clearpage
\section{More Results on LLM Sampling with PRNG}

\subsection{Tool-using PRNG algorithm with LLMs}

\begin{figure*}[htbp]
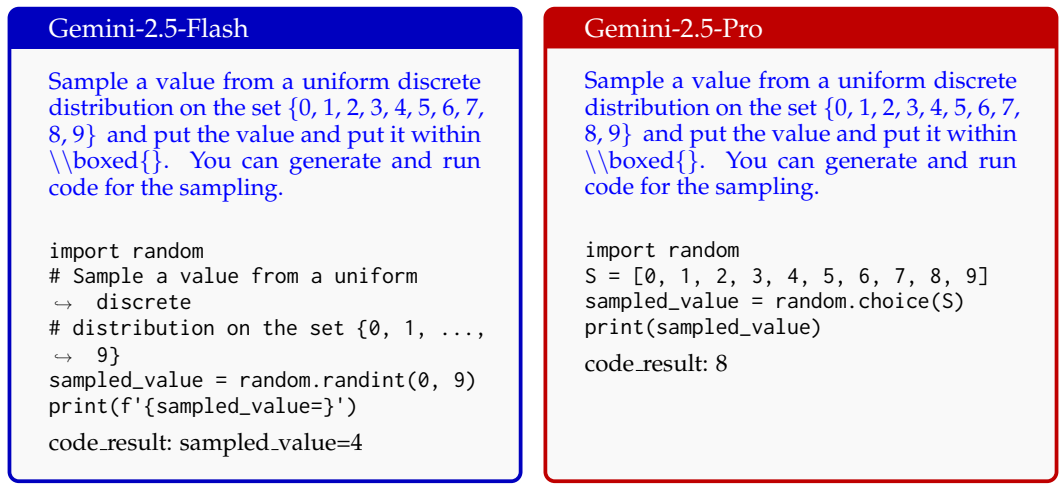

\centering
\begin{minipage}[t]{0.49\textwidth}
\begin{tcolorbox}[
  width=\textwidth,
  colback=gray!5,
  colframe=blue!75!black,
  equal height group=I, 
  title=Gemini-2.5-Flash
]
\small{
\textcolor{blue}{Sample a value from a uniform discrete distribution on the set \{0, 1, 2, 3, 4, 5, 6, 7, 8, 9\}\, and put the value and put it within \textbackslash{}\textbackslash{}boxed\{\}. You can generate and run code for the sampling. } \\
\begin{Verbatim}[breaklines,fontsize=\small]
import random
# Sample a value from a uniform discrete 
# distribution on the set {0, 1, ..., 9}
sampled_value = random.randint(0, 9)
print(f'{sampled_value=}')
\end{Verbatim}
code\_result: sampled\_value=4
}
\end{tcolorbox}
\end{minipage}
\hfill 
\begin{minipage}[t]{0.49\textwidth}
\begin{tcolorbox}[
  width=\textwidth,
  colback=gray!5,
  colframe=red!75!black,
  equal height group=I, 
  title=Gemini-2.5-Pro
]
\small{
\textcolor{blue}{Sample a value from a uniform discrete distribution on the set \{0, 1, 2, 3, 4, 5, 6, 7, 8, 9\}\, and put the value and put it within \textbackslash{}\textbackslash{}boxed\{\}. You can generate and run code for the sampling. } \\
\begin{Verbatim}[breaklines,fontsize=\small]
import random
S = [0, 1, 2, 3, 4, 5, 6, 7, 8, 9]
sampled_value = random.choice(S)
print(sampled_value)
\end{Verbatim}
code\_result: 8
}
\end{tcolorbox}
\end{minipage}
\caption{Example code generation and execution results of (\emph{\textbf{Left}}) Gemini-2.5-Flash and (\emph{\textbf{Right}}) Gemini-2.5-Pro with tool-use.}
\label{tool_use1}
\end{figure*}

We follow the document~\footnote{\url{https://ai.google.dev/gemini-api/docs/code-execution}} to enable Gemini family with a Python code execution tool. Meanwhile, we add extra instruction \texttt{You can generate and run code for the sampling}. In this manner, LLMs can generate a snippet of Python code (using the function ``random.randint()'') and run it to get the results, as shown in Figure~\ref{tool_use1}. However, LLMs still demonstrate strong biases, as illustrated in Figure~\ref{tool-use2} (\emph{\textbf{Left}}). We hypothesize that the execution tool may initialize the random seed on its own. Therefore, we further instruct LLMs to use current time as random seed, as shown in Figure~\ref{tool_use2}. Normally, Gemini-3-Pro follows this instruction well to generate a line of code ``random.seed(time.time())'' while Gemini-2.5-Pro and Gemini-2.5-Flash may not always faithfully use the current time to set the random seed. Consequently, as visualized in (\emph{\textbf{Right}}), Gemini-3.0-Pro is the most reliable model to sample from a distribution with tool-use. Despite the promising results, it can be expected if multiple independent sampling happens concurrently, the biases to certain numbers will emerge.\looseness=-1

\begin{figure*}[htbp]
\centering
\begin{minipage}[t]{0.49\textwidth}
\begin{tcolorbox}[
  width=\textwidth,
  colback=gray!5,
  colframe=blue!75!black,
  equal height group=J, 
  title=Gemini-2.5-Flash
]
\small{
\textcolor{blue}{Sample a value from a uniform discrete distribution on the set \{0, 1, 2, 3, 4, 5, 6, 7, 8, 9\}\, and put the value and put it within \textbackslash{}\textbackslash{}boxed\{\}. You can generate and run code for the sampling. Make sure to seed the random number generator to ensure different results on each run, for example using the current time.} \\
\begin{Verbatim}[breaklines,fontsize=\small]
import random
import time
# Seed the random number generator 
# with the current time
random.seed(time.time())
# Sample a value from a uniform discrete 
# distribution on the set {0, 1, ..., 9}
sampled_value = random.randint(0, 9)
print(f"sampled_value={sampled_value}")
\end{Verbatim}
code\_result: sampled\_value=8
}
\end{tcolorbox}
\end{minipage}
\hfill 
\begin{minipage}[t]{0.49\textwidth}
\begin{tcolorbox}[
  width=\textwidth,
  colback=gray!5,
  colframe=red!75!black,
  equal height group=J, 
  title=Gemini-3.0-Pro
]
\small{
\textcolor{blue}{Sample a value from a uniform discrete distribution on the set \{0, 1, 2, 3, 4, 5, 6, 7, 8, 9\}\, and put the value and put it within \textbackslash{}\textbackslash{}boxed\{\}. You can generate and run code for the sampling. Make sure to seed the random number generator to ensure different results on each run, for example using the current time.} \\
\begin{Verbatim}[breaklines,fontsize=\small]
import random
import time
# Seeding with current time to satisfy 
# the user's request, 
# though standard random init usually 
# does this automatically.
random.seed(time.time())
val = random.randint(0, 9)
print(val)
\end{Verbatim}
code\_result: 6
}
\end{tcolorbox}
\end{minipage}
\caption{Example code generation and execution results of (\emph{\textbf{Left}}) Gemini-2.5-Flash and (\emph{\textbf{Right}}) Gemini-3.0-Pro with tool-use.}
\label{tool_use2}
\end{figure*}

\begin{figure}[t]
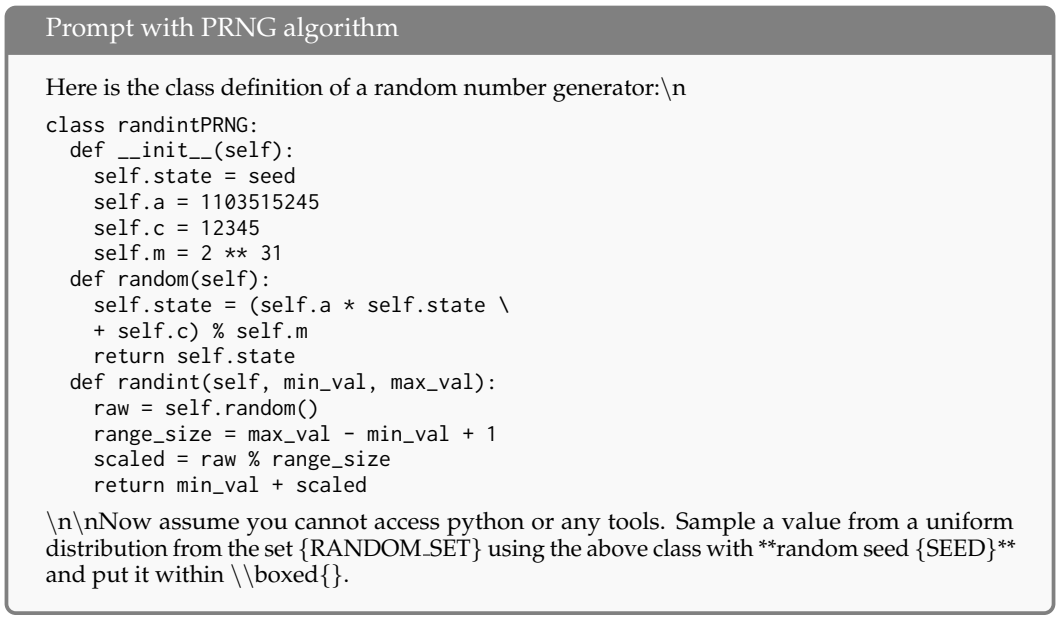

\centering
\begin{tcolorbox}[
  width=\textwidth,
  colback=gray!5,
  colframe=gray,
  title=Prompt with PRNG algorithm,
  fontupper=\small, 
]
Here is the class definition of a random number generator:\textbackslash{}n
\begin{Verbatim}[breaklines,fontsize=\small]
class randintPRNG:
  def __init__(self):
    self.state = seed
    self.a = 1103515245
    self.c = 12345
    self.m = 2 ** 31
  def random(self):
    self.state = (self.a * self.state \
    + self.c) % self.m
    return self.state
  def randint(self, min_val, max_val):
    raw = self.random()
    range_size = max_val - min_val + 1
    scaled = raw % range_size
    return min_val + scaled
\end{Verbatim}
\textbackslash{}n\textbackslash{}nNow assume you cannot access python or any tools. Sample a value from a uniform distribution from the set \{RANDOM\_SET\} using the above class with **random seed \{SEED\}** and put it within \textbackslash{}\textbackslash{}boxed\{\}.
\end{tcolorbox}
\caption{Prompt design for using LLMs to sample from a uniform discrete distribution with a PRNG algorithm and a random seed without tool calling (empty lines and code comments are truncated).\looseness=-1}
\label{prompt_prng}
\end{figure}

\begin{figure*}[t]
\centering
\begin{tcolorbox}[
  width=\textwidth,
  colback=gray!5,
  colframe=gray,
  title=Prompt with PRNG algorithm for uniform continuous distribution,
  fontupper=\small, 
]
Here is the class definition of a random number generator:\textbackslash{}n
\begin{Verbatim}[breaklines,fontsize=\small]
class randPRNG:
    def __init__(self, seed):
        self.state = seed
        self.a = 1103515245
        self.c = 12345
        self.m = 2 ** 31
    def next_raw(self):
        self.state = (self.a * self.state + self.c) % self.m
        return self.state / self.m
    def rand(self, low, high):
        # 1. Get the base random number [0, 1)
        base = self.next_raw()
        # 2. Scale it to the range size (high - low)
        range_size = high - low
        # 3. Shift it to start at 'low'
        return low + (base * range_size)
\end{Verbatim}
\textbackslash{}n\textbackslash{}nNow assume you cannot access python or any tools. Sample a number from a uniform distribution within the continuous range [0, 1] using the above class with **random seed \{SEED\}** and put it within \textbackslash{}\textbackslash{}boxed\{\}.
\end{tcolorbox}
\caption{Prompt design for using LLMs to sample from a uniform continuous distribution with a PRNG algorithm (empty lines and code comments are truncated).}
\label{prompt_prng2}
\end{figure*}

\begin{figure*}[t]
\centering
\begin{tcolorbox}[
  width=\textwidth,
  colback=gray!5,
  colframe=gray,
  title=Prompt with PRNG algorithm for Gaussian distribution,
  fontupper=\small, 
]
Here is the class definition of a random number generator:\textbackslash{}n
\begin{Verbatim}[breaklines,fontsize=\small]
class randnPRNG:
    def __init__(self, seed):
        self.state = seed
        self.a = 1103515245
        self.c = 12345
        self.m = 2 ** 31
    def next_raw(self):
        # Generates a float between 0.0 and 1.0 using LCG.
        self.state = (self.a * self.state + self.c) % self.m
        return self.state / self.m
    def _get_standard_normal(self):
        # Generate two uniform numbers
        u1 = self.next_raw()
        u2 = self.next_raw()
        while u1 == 0:  # Avoid log(0) error
            u1 = self.next_raw()
        # Apply Box-Muller Transform
        R = math.sqrt(-2.0 * math.log(u1))
        theta = 2.0 * math.pi * u2
        z0 = R * math.cos(theta)
        return z0
    def randn(self, mu, sigma):
        # Get a standard normal value Z ~ N(0, 1)
        Z = self._get_standard_normal()
        # Scale and Shift: X = mu + (Z * sigma)
        return mu + (Z * sigma)
\end{Verbatim}
\textbackslash{}n\textbackslash{}nNow assume you cannot access python or any tools. Sample a number from a Gaussian distribution with mean 0 and standard deviation 1 using the above class with **random seed \{SEED\}** and put it within \textbackslash{}\textbackslash{}boxed\{\}.
\end{tcolorbox}
\caption{Prompt design for using LLMs to sample from a Gaussian distribution with a PRNG algorithm (empty lines and code comments are truncated).}
\label{prompt_prng3}
\end{figure*}

\clearpage

\subsection{Simulating PRNG algorithm with LLMs}\label{appendix_simulate}

\begin{figure*}[htbp]
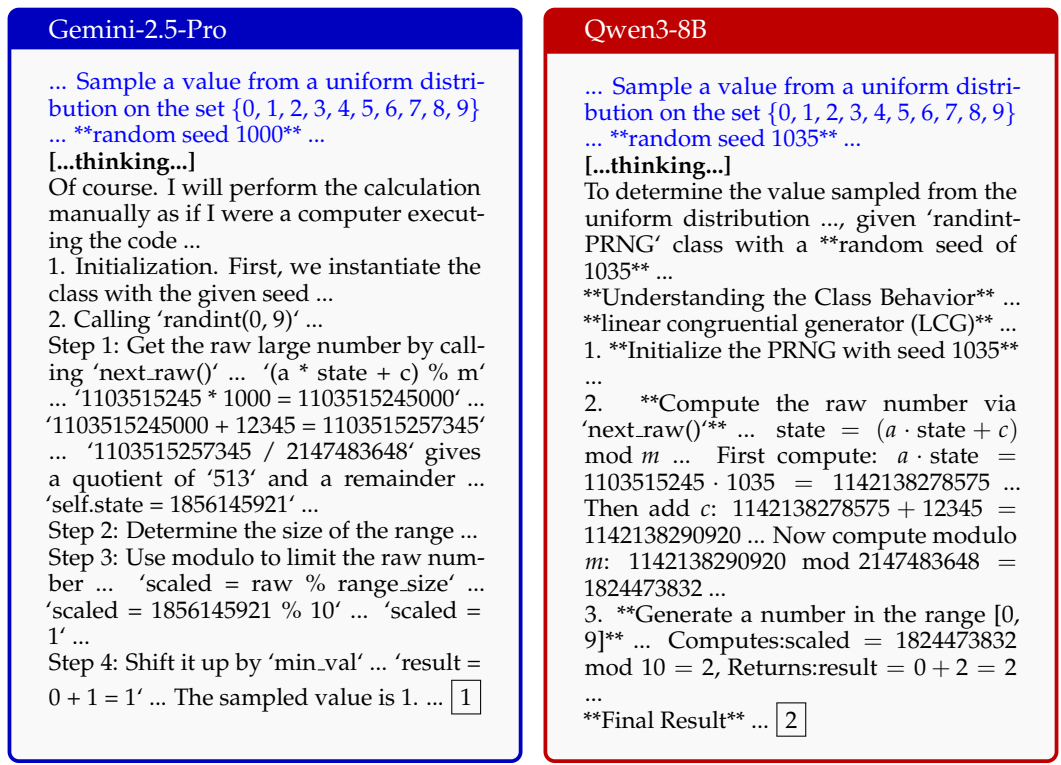

\centering
\begin{minipage}[t]{0.49\textwidth}
\begin{tcolorbox}[
  width=\textwidth,
  colback=gray!5,
  colframe=blue!75!black,
  equal height group=B, 
  title=Gemini-2.5-Pro
]
\small{
\textcolor{blue}{... Sample a value from a uniform distribution on the set \{0, 1, 2, 3, 4, 5, 6, 7, 8, 9\} ... **random seed 1000** ...} \\
\textbf{[...thinking...]}\\
Of course. I will perform the calculation manually as if I were a computer executing the code ...\\ 
1. Initialization. First, we instantiate the class with the given seed ...\\
2. Calling `randint(0, 9)` ... \\
Step 1: Get the raw large number by calling `next\_raw()` ... `(a * state + c) \% m` ... `1103515245 * 1000 = 1103515245000` ... `1103515245000 + 12345 = 1103515257345` ... `1103515257345 / 2147483648` gives a quotient of `513` and a remainder ... `self.state = 1856145921` ... \\
Step 2: Determine the size of the range ... \\
Step 3: Use modulo to limit the raw number ... `scaled = raw \% range\_size` ...   `scaled = 1856145921 \% 10` ... `scaled = 1` ... \\
Step 4: Shift it up by `min\_val` ... `result = 0 + 1 = 1` ...  The sampled value is 1. ... $\boxed{1}$
}
\end{tcolorbox}
\end{minipage}
\hfill 
\begin{minipage}[t]{0.49\textwidth}
\begin{tcolorbox}[
  width=\textwidth,
  colback=gray!5,
  colframe=red!75!black,
  equal height group=B, 
  title=Qwen3-8B
]
\small{
\textcolor{blue}{... Sample a value from a uniform distribution on the set \{0, 1, 2, 3, 4, 5, 6, 7, 8, 9\} ... **random seed 1035** ... } \\
\textbf{[...thinking...]}\\
To determine the value sampled from the uniform distribution ..., given `randintPRNG` class with a **random seed of 1035** ...\\
**Understanding the Class Behavior** ...  **linear congruential generator (LCG)** ...\\
1. **Initialize the PRNG with seed 1035** ... \\
2. **Compute the raw number via `next\_raw()`** ... $\text{state} = (a \cdot \text{state} + c) \mod m$ ... First compute: $a \cdot \text{state} = 1103515245 \cdot 1035 = 1142138278575$ ... Then add $ c$:  $1142138278575 + 12345 = 1142138290920$ ... Now compute modulo $ m $: $1142138290920 \mod 2147483648 = 1824473832$ ... \\
3. **Generate a number in the range [0, 9]** ... Computes:$\text{scaled} = 1824473832 \mod 10 = 2$, Returns:$\text{result} = 0 + 2 = 2$ ... \\ 
**Final Result** ... $\boxed{2}$
}
\end{tcolorbox}
\end{minipage}
\caption{Truncated responses when prompting  (\emph{\textbf{Left}}) Gemini-2.5-Pro and (\emph{\textbf{Right}}) Qwen3-8B to simulate a PRNG algorithm for uniform discrete distribution between 0 and 9. The thinking traces and some special tokens are also truncated as they are too verbose.}
\label{prng_think1}
\end{figure*}

\begin{figure*}[htbp]
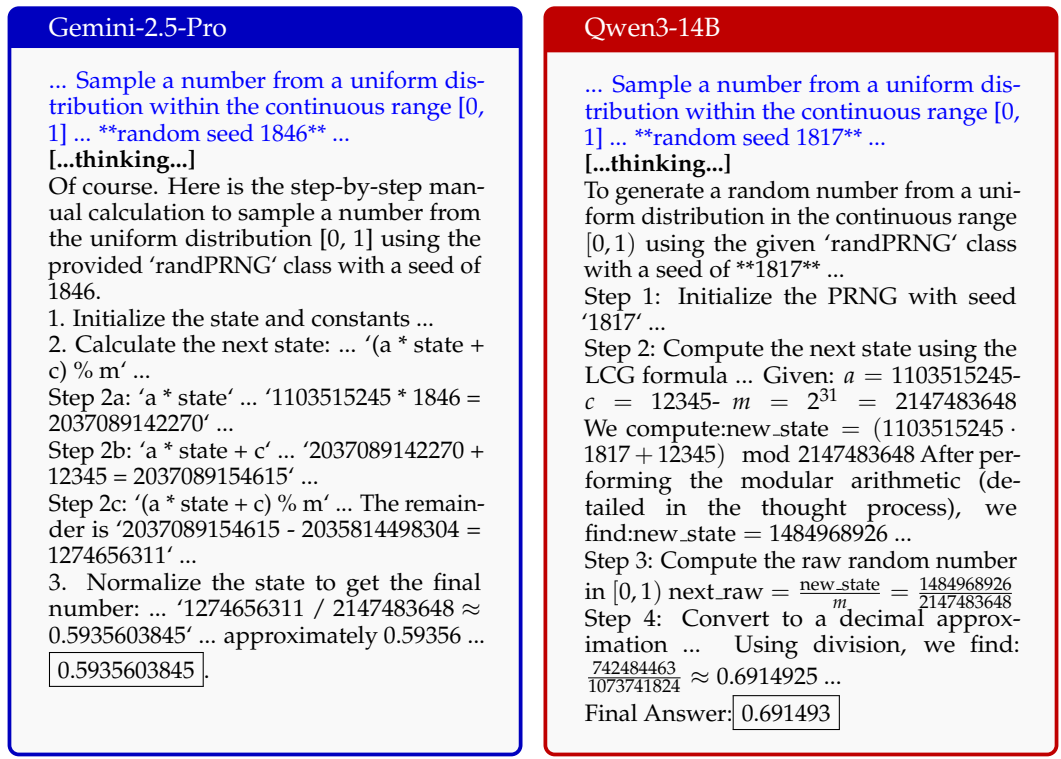

\centering
\begin{minipage}[t]{0.49\textwidth}
\begin{tcolorbox}[
  width=\textwidth,
  colback=gray!5,
  colframe=blue!75!black,
  equal height group=C, 
  title=Gemini-2.5-Pro
]
\small{
\textcolor{blue}{... Sample a number from a uniform distribution within the continuous range [0, 1] ... **random seed 1846** ...} \\
\textbf{[...thinking...]}\\
Of course. Here is the step-by-step manual calculation to sample a number from the uniform distribution [0, 1] using the provided `randPRNG` class with a seed of 1846.\\ 
1. Initialize the state and constants ...\\
2. Calculate the next state: ... `(a * state + c) \% m`  ... \\
Step 2a: `a * state` ...    `1103515245 * 1846 = 2037089142270`  ... \\
Step 2b: `a * state + c`  ... `2037089142270 + 12345 = 2037089154615` ... \\
Step 2c: `(a * state + c) \% m` ... The remainder is `2037089154615 - 2035814498304 = 1274656311` ... \\
3. Normalize the state to get the final number: ... `1274656311 / 2147483648 $\approx$ 0.5935603845` ... approximately 0.59356 ... $\boxed{0.5935603845}$.
}
\end{tcolorbox}
\end{minipage}
\hfill 
\begin{minipage}[t]{0.49\textwidth}
\begin{tcolorbox}[
  width=\textwidth,
  colback=gray!5,
  colframe=red!75!black,
  equal height group=C, 
  title=Qwen3-14B
]
\small{
\textcolor{blue}{... Sample a number from a uniform distribution within the continuous range [0, 1] ... **random seed 1817** ... } \\
\textbf{[...thinking...]}\\
To generate a random number from a uniform distribution in the continuous range $[0, 1)$ using the given `randPRNG` class with a seed of **1817** ...\\
Step 1: Initialize the PRNG with seed `1817` ... \\
Step 2: Compute the next state using the LCG formula ... Given: $ a = 1103515245 $- $ c = 12345 $- $ m = 2^{31} = 2147483648 $ We compute:$\text{new\_state} = (1103515245 \cdot 1817 + 12345) \mod 2147483648$ After performing the modular arithmetic (detailed in the thought process), we find:$\text{new\_state} = 1484968926$ ...\\
Step 3: Compute the raw random number in $[0, 1)$ $\text{next\_raw} = \frac{\text{new\_state}}{m} = \frac{1484968926}{2147483648}$ \\
Step 4: Convert to a decimal approximation ... Using division, we find: $\frac{742484463}{1073741824} \approx 0.6914925$ ... \\
Final Answer:$\boxed{0.691493}$ 
}
\end{tcolorbox}
\end{minipage}
\caption{Truncated responses when prompting  (\emph{\textbf{Left}}) Gemini-2.5-Pro and (\emph{\textbf{Right}}) Qwen3-14B to simulate a PRNG algorithm for uniform continuous distribution [0, 1]. The thinking traces and some special tokens are also truncated as they are too verbose.}
\label{prng_think2}
\end{figure*}

In Figure~\ref{prompt_prng2} and Figure~\ref{prompt_prng3}, we showcase the prompt design for simulating PRNG algorithms for uniform continuous distribution and Gaussian distribution with LLMs. In the main paper, we already demonstrate that LLMs can achieve greater than 90\% accuracy for simulations of sampling from uniform distributions. Here we also conduct goodness-of-fit tests for various scenarios in Table~\ref{prng_gof}. Firstly, the p-values for uniform distributions are significantly larger than 0.05 when LLMs are strong enough. This is reasonable as LLMs have larger than 90\% accuracies when simulating PRNG algorithms. We also visualize several chain-of-thought reasoning processes for uniform distributions in Figure~\ref{prng_think1} and Figure~\ref{prng_think2}, which corroborates the high accuracies.\looseness=-1


\begin{table*}[htbp]
\caption{Goodness-of-fit tests when prompting Qwen3 and Gemini-2.5 families to simulate PRNG algorithms for three representative distributions. We \textbf{bold} the scenarios that achieve $p>0.05$, which shows the tendency of better sampling results w.r.t. model capabilities.\looseness=-1}
\begin{center}
\begin{tabular}{l|lll}
\toprule
Model & Uniform $\{0,1,...,9\}$ & Uniform $[0,1]$ & Gaussian $\mathcal{N}(0,1)$ \\
\midrule
Qwen3-1.7B 
& 5.27e-310 & 2.57e-30 & 1.98e-78 \\

Qwen3-4B 
& 0.02 & \textbf{1.00} & 1.42e-84 \\

Qwen3-8B 
& \textbf{1.00} & \textbf{1.00} & 8.68e-83 \\

Qwen3-14B 
& \textbf{0.98} & \textbf{1.00} & 5.93e-137 \\

Gemini-2.5-Flash 
& \textbf{0.22} & 1.30e-35 & 2.33e-138 \\

Gemini-2.5-Pro 
& \textbf{1.00} & \textbf{1.00} & 2.09e-15 \\
\bottomrule
\end{tabular}
\end{center}
\label{prng_gof}
\end{table*}

\begin{figure}[htbp]
    \centering
    \begin{subfigure}[b]{0.23\textwidth}
        \centering
        \includegraphics[width=\linewidth]{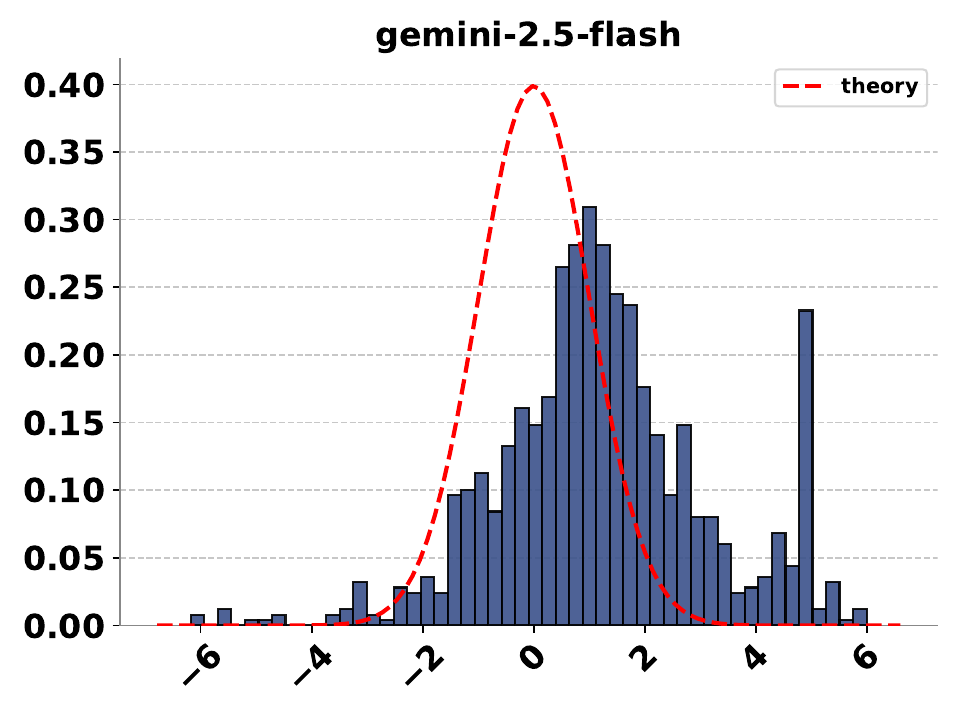}
    \end{subfigure}
    \hfill
    \begin{subfigure}[b]{0.23\textwidth}
        \centering
        \includegraphics[width=\linewidth]{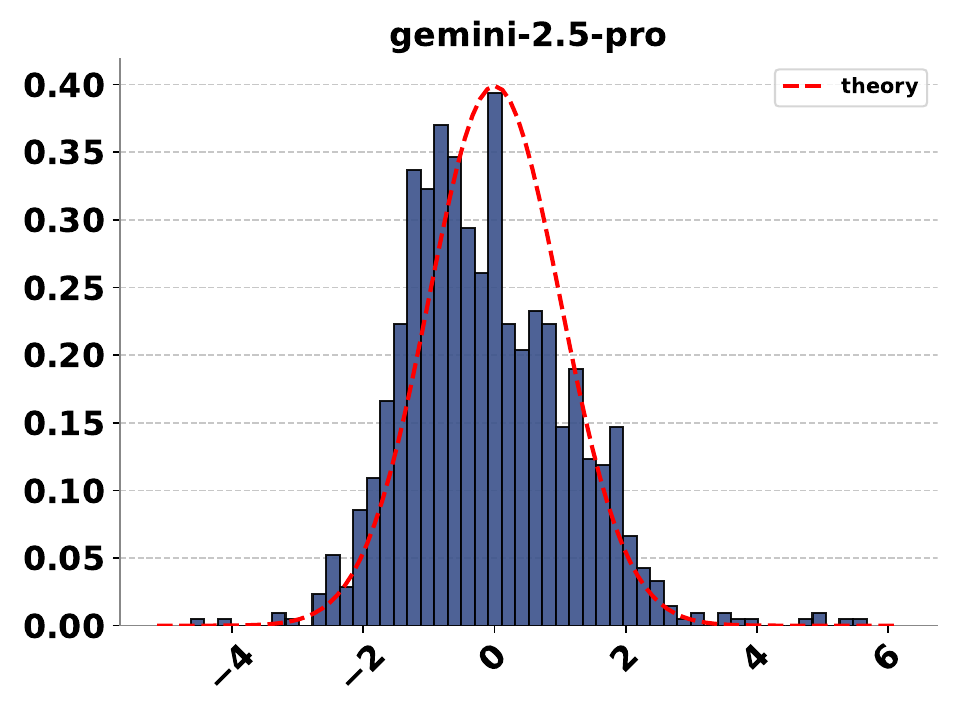}
    \end{subfigure}
    \hfill
    \begin{subfigure}[b]{0.23\textwidth}
        \centering
        \includegraphics[width=\linewidth]{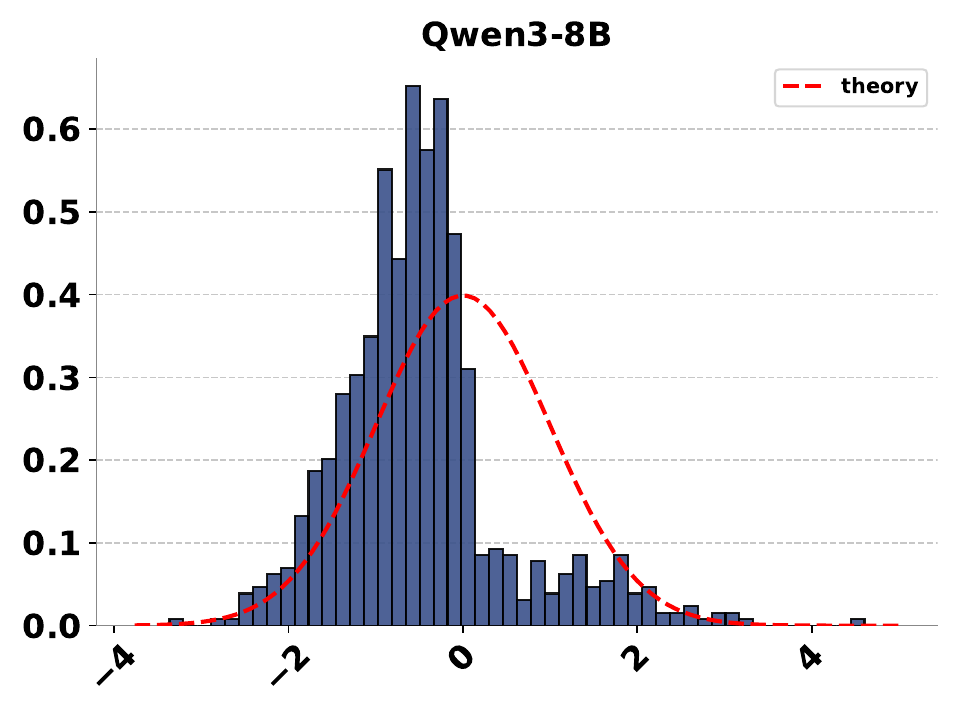}
    \end{subfigure}
    \hfill
    \begin{subfigure}[b]{0.23\textwidth}
        \centering
        \includegraphics[width=\linewidth]{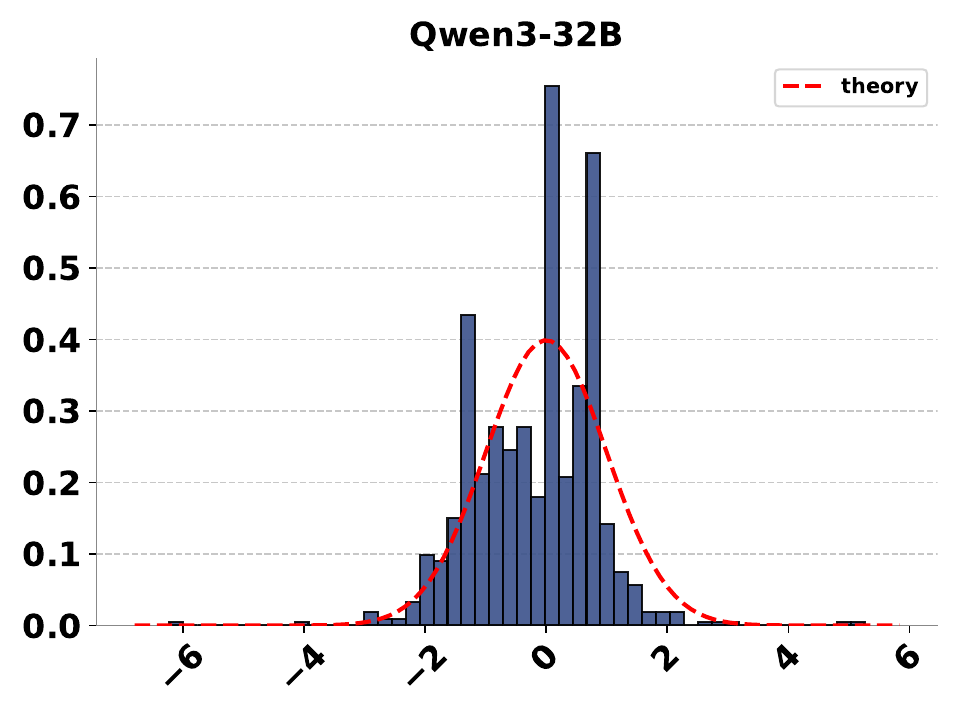}
    \end{subfigure}
    \caption{From \emph{\textbf{left}} to \emph{\textbf{right}}, we visualize the failures of LLMs simulating PRNG algorithms for Gaussian distributions: Gemini-2.5-Flash, Gemini-2.5-Pro, Qwen3-8B, Qwen3-32B.}
    \label{prng_gauss}
\end{figure}

\begin{figure*}[htbp]
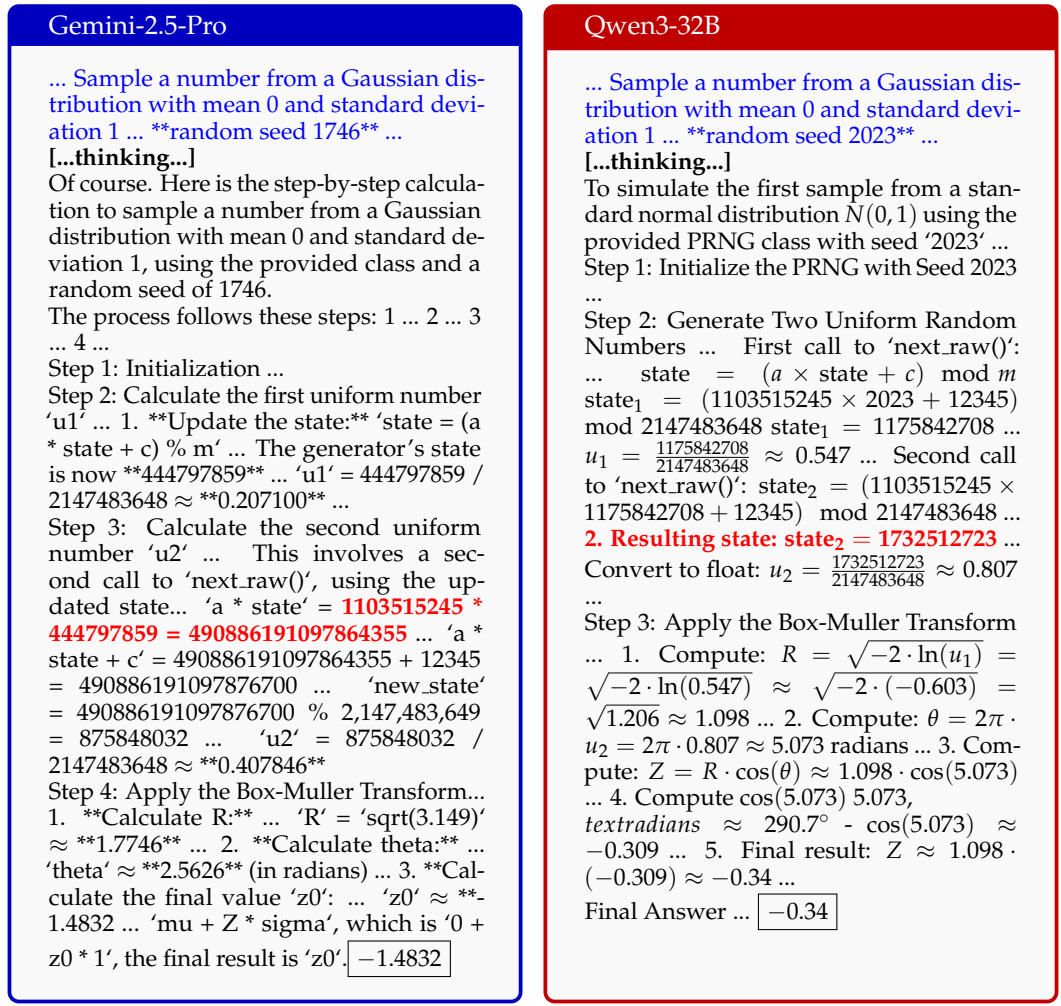

\centering
\begin{minipage}[t]{0.49\textwidth}
\begin{tcolorbox}[
  width=\textwidth,
  colback=gray!5,
  colframe=blue!75!black,
  equal height group=D, 
  title=Gemini-2.5-Pro
]
\small{
\textcolor{blue}{... Sample a number from a Gaussian distribution with mean 0 and standard deviation 1 ... **random seed 1746** ...} \\
\textbf{[...thinking...]}\\
Of course. Here is the step-by-step calculation to sample a number from a Gaussian distribution with mean 0 and standard deviation 1, using the provided class and a random seed of 1746.\\ 
The process follows these steps: 1 ... 2 ... 3 ... 4 ...\\
Step 1: Initialization ... \\
Step 2: Calculate the first uniform number `u1` ... 1.  **Update the state:** `state = (a * state + c) \% m` ... The generator's state is now **444797859** ... `u1` = 444797859 / 2147483648 $\approx$ **0.207100** ... \\
Step 3: Calculate the second uniform number `u2` ... This involves a second call to `next\_raw()`, using the updated state... `a * state` = \textcolor{red}{\textbf{1103515245 * 444797859 = 490886191097864355}} ...   `a * state + c` = 490886191097864355 + 12345 = 490886191097876700 ... `new\_state` = 490886191097876700 \% 2,147,483,649 = 875848032 ... `u2` = 875848032 / 2147483648 $\approx$ **0.407846** \\
Step 4: Apply the Box-Muller Transform... 1. **Calculate R:** ...  `R` = `sqrt(3.149)` $\approx$ **1.7746** ... 2.  **Calculate theta:** ... `theta` $\approx$ **2.5626** (in radians) ... 3.  **Calculate the final value `z0`: ... `z0` $\approx$ **-1.4832 ... `mu + Z * sigma`, which is `0 + z0 * 1`, the final result is `z0`.$\boxed{-1.4832}$
}
\end{tcolorbox}
\end{minipage}
\hfill 
\begin{minipage}[t]{0.49\textwidth}
\begin{tcolorbox}[
  width=\textwidth,
  colback=gray!5,
  colframe=red!75!black,
  equal height group=D, 
  title=Qwen3-32B
]
\small{
\textcolor{blue}{... Sample a number from a Gaussian distribution with mean 0 and standard deviation 1 ... **random seed 2023** ...} \\
\textbf{[...thinking...]}\\
To simulate the first sample from a standard normal distribution $ N(0, 1) $ using the provided PRNG class with seed `2023` ...\\
Step 1: Initialize the PRNG with Seed 2023 ... \\
Step 2: Generate Two Uniform Random Numbers ... First call to `next\_raw()`: ... $   \text{state} = (a \times \text{state} + c) \mod m   $   $   \text{state}_1 = (1103515245 \times 2023 + 12345) \mod 2147483648  $  $ \text{state}_1 = 1175842708   $ ... $   u_1 = \frac{1175842708}{2147483648} \approx 0.547   $ ... Second call to `next\_raw()`: $   \text{state}_2 = (1103515245 \times 1175842708 + 12345) \mod 2147483648  $ ... \textcolor{red}{\textbf{2. Resulting state:   $\textbf{state}_\textbf{2} = \textbf{1732512723}$}} ... Convert to float:   $   u_2 = \frac{1732512723}{2147483648} \approx 0.807   $ ... \\
Step 3: Apply the Box-Muller Transform ... 1. Compute: $   R = \sqrt{-2 \cdot \ln(u_1)} = \sqrt{-2 \cdot \ln(0.547)} \approx \sqrt{-2 \cdot (-0.603)} = \sqrt{1.206} \approx 1.098   $ ... 2. Compute: $   \theta = 2\pi \cdot u_2 = 2\pi \cdot 0.807 \approx 5.073 \text{ radians}   $ ... 3. Compute: $   Z = R \cdot \cos(\theta) \approx 1.098 \cdot \cos(5.073)   $ ... 4. Compute $\cos(5.073) $ $ 5.073, \\text{radians} \approx 290.7^\circ $   - $ \cos(5.073) \approx -0.309 $ ... 5. Final result: $   Z \approx 1.098 \cdot (-0.309) \approx -0.34   $ ...\\
Final Answer ... $\boxed{-0.34}$
}
\end{tcolorbox}
\end{minipage}
\caption{Truncated responses when prompting  (\emph{\textbf{Left}}) Gemini-2.5-Pro and (\emph{\textbf{Right}}) Qwen3-32B to simulate a PRNG algorithm for Gaussian distribution $\mathcal{N}(0, 1)$. The thinking traces and some special tokens are also truncated as they are too verbose. We use \textcolor{red}{\textbf{bold red}} to mark the first mistake LLM made. Typically LLMs struggle with large number multiplications.}
\label{prng_think3}
\end{figure*}

\textbf{Inability of LLMs for Gaussian distributions.} From Figure~\ref{prng_gauss}, with a PRNG algorithm and a random seed as part of prompt, LLMs indeed become more reliable when sampling from a Gaussian distribution than before. However, the low p-values indicate that LLMs begin to struggle with Gaussian distributions, even for Gemini-2.5-Pro. Firstly, we find that the accuracy of LLMs simulating PRNG algorithm for Gaussian distributions drops significantly, e.g., smaller than 5\% for Gemini-2.5-Pro. After further exploring the chain-of-thought reasoning process, we find the pattern of how LLMs fail. Specifically, different from uniform distributions, Gaussian distributions typically require generating two random numbers. We visualize the chain-of-thought reasoning processes in Figure~\ref{prng_think3}. The first generation uses the random seed provided in the prompt (normally small) as the initial state, which can be precisely computed by LLMs. While for the second generation, the random seed already becomes very large. Therefore, LLMs fail to precisely compute the product of  two very large numbers, corroborating the inability of LLMs to execute PRNG algorithms shown in \citet{markeeva2024clrs}.\looseness=-1

%% file: appendix/appendix_d.tex
\clearpage

\section{More Results on Distribution Conversion with LLMs}\label{appendix_convert}

\begin{figure*}[htbp]
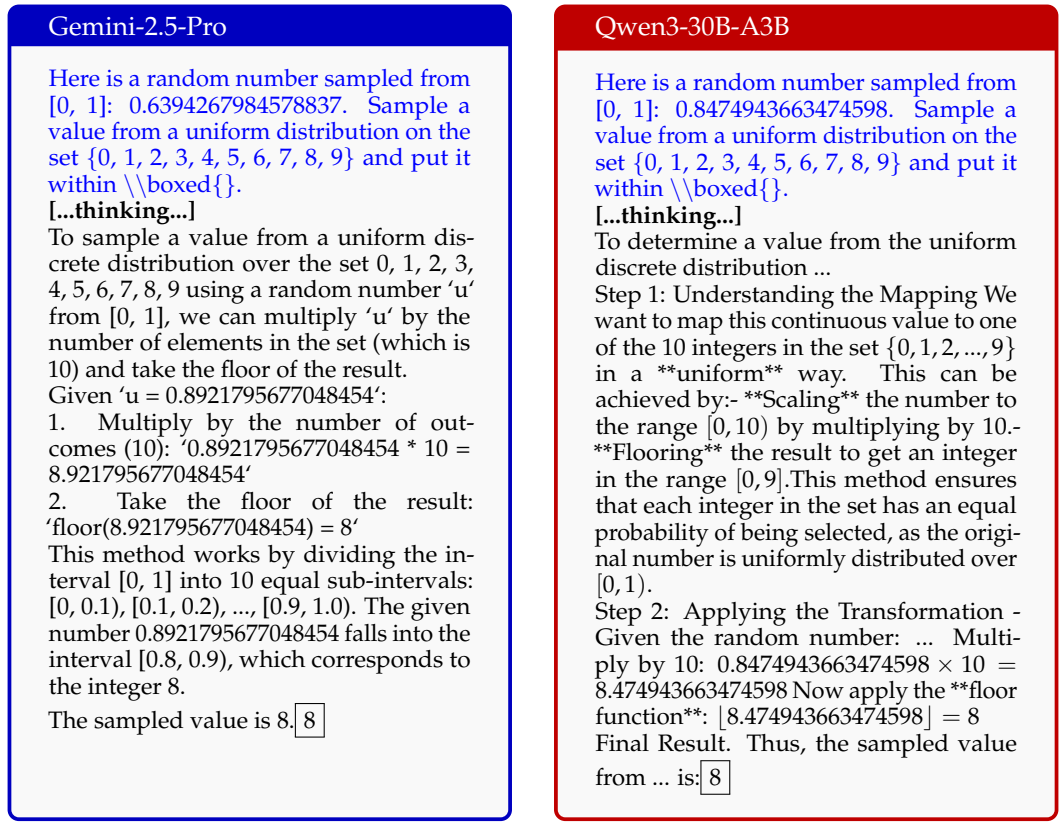

\centering
\begin{minipage}[t]{0.48\textwidth}
\begin{tcolorbox}[
  width=\textwidth,
  colback=gray!5,
  colframe=blue!75!black,
  equal height group=K, 
  title=Gemini-2.5-Pro
]
\small{
\textcolor{blue}{Here is a random number sampled from [0, 1]: 0.6394267984578837. Sample a value from a uniform distribution on the set \{0, 1, 2, 3, 4, 5, 6, 7, 8, 9\} and put it within \textbackslash{}\textbackslash{}boxed\{\}.} \\
\textbf{[...thinking...]}\\
To sample a value from a uniform discrete distribution over the set {0, 1, 2, 3, 4, 5, 6, 7, 8, 9} using a random number `u` from [0, 1], we can multiply `u` by the number of elements in the set (which is 10) and take the floor of the result.\\
Given `u = 0.8921795677048454`:\\
1.  Multiply by the number of outcomes (10):    `0.8921795677048454 * 10 = 8.921795677048454`\\
2.  Take the floor of the result:    `floor(8.921795677048454) = 8`\\
This method works by dividing the interval [0, 1] into 10 equal sub-intervals: [0, 0.1), [0.1, 0.2), ..., [0.9, 1.0). The given number 0.8921795677048454 falls into the interval [0.8, 0.9), which corresponds to the integer 8.\\
The sampled value is 8.$\boxed{8}$
}
\end{tcolorbox}
\end{minipage}
\hfill 
\begin{minipage}[t]{0.48\textwidth}
\begin{tcolorbox}[
  width=\textwidth,
  colback=gray!5,
  colframe=red!75!black,
  equal height group=K, 
  title=Qwen3-30B-A3B
]
\small{
\textcolor{blue}{Here is a random number sampled from [0, 1]: 0.8474943663474598. Sample a value from a uniform distribution on the set \{0, 1, 2, 3, 4, 5, 6, 7, 8, 9\} and put it within \textbackslash{}\textbackslash{}boxed\{\}.} \\
\textbf{[...thinking...]}\\
To determine a value from the uniform discrete distribution ... \\
Step 1: Understanding the Mapping We want to map this continuous value to one of the 10 integers in the set $\{0, 1, 2, ..., 9\}$ in a **uniform** way. This can be achieved by:- **Scaling** the number to the range $[0, 10)$ by multiplying by 10.- **Flooring** the result to get an integer in the range $[0, 9]$.This method ensures that each integer in the set has an equal probability of being selected, as the original number is uniformly distributed over $[0, 1)$. \\
Step 2: Applying the Transformation - Given the random number: ... Multiply by 10: $0.8474943663474598 \times 10 = 8.474943663474598$ Now apply the **floor function**: $\lfloor 8.474943663474598 \rfloor = 8$ \\
Final Result. Thus, the sampled value from ... is:$\boxed{8}$
}
\end{tcolorbox}
\end{minipage}
\caption{Responses when prompting (\emph{\textbf{Left}}) Gemini-2.5-Pro (\emph{\textbf{Right}}) Qwen3-30B-A3B to use a random number in [0, 1] to sample from a uniform discrete distribution between 0 and 9.\looseness=-1}
\label{convert1}
\end{figure*}

\begin{figure*}[htbp]
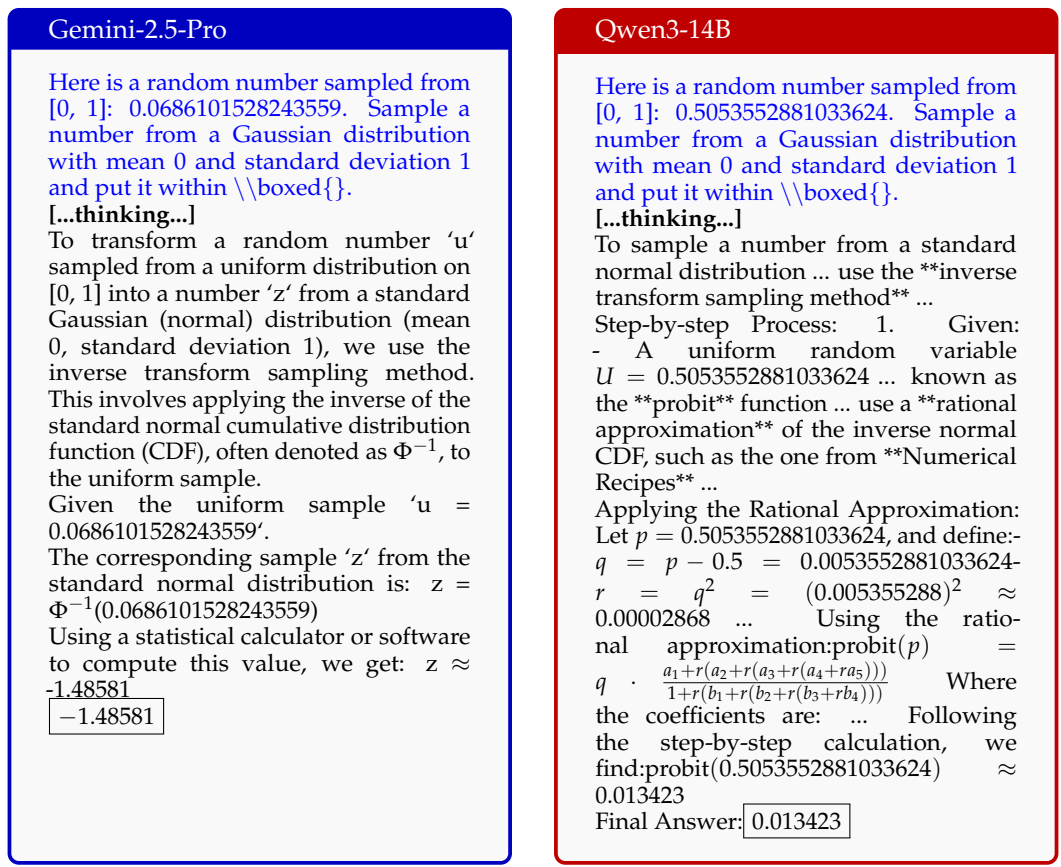

\centering
\begin{minipage}[t]{0.48\textwidth}
\begin{tcolorbox}[
  width=\textwidth,
  colback=gray!5,
  colframe=blue!75!black,
  equal height group=L, 
  title=Gemini-2.5-Pro
]
\small{
\textcolor{blue}{Here is a random number sampled from [0, 1]: 0.0686101528243559. Sample a number from a Gaussian distribution with mean 0 and standard deviation 1 and put it within \textbackslash{}\textbackslash{}boxed\{\}.} \\
\textbf{[...thinking...]}\\
To transform a random number `u` sampled from a uniform distribution on [0, 1] into a number `z` from a standard Gaussian (normal) distribution (mean 0, standard deviation 1), we use the inverse transform sampling method. This involves applying the inverse of the standard normal cumulative distribution function (CDF), often denoted as $\Phi^{-1}$, to the uniform sample.\\
Given the uniform sample `u = 0.0686101528243559`.\\
The corresponding sample `z` from the standard normal distribution is: z = $\Phi^{-1}$(0.0686101528243559)\\
Using a statistical calculator or software to compute this value, we get: z $\approx$ -1.48581\\
$\boxed{-1.48581}$
}
\end{tcolorbox}
\end{minipage}
\hfill 
\begin{minipage}[t]{0.48\textwidth}
\begin{tcolorbox}[
  width=\textwidth,
  colback=gray!5,
  colframe=red!75!black,
  equal height group=L, 
  title=Qwen3-14B
]
\small{
\textcolor{blue}{Here is a random number sampled from [0, 1]: 0.5053552881033624. Sample a number from a Gaussian distribution with mean 0 and standard deviation 1 and put it within \textbackslash{}\textbackslash{}boxed\{\}.} \\
\textbf{[...thinking...]}\\
To sample a number from a standard normal distribution ... use the **inverse transform sampling method** ...\\
Step-by-step Process:  1. Given: - A uniform random variable $ U = 0.5053552881033624 $ ... known as the **probit** function ...  use a **rational approximation** of the inverse normal CDF, such as the one from **Numerical Recipes** ... \\
Applying the Rational Approximation: Let $ p = 0.5053552881033624 $, and define:- $ q = p - 0.5 = 0.0053552881033624 $- $ r = q^2 = (0.005355288)^2 \approx 0.00002868 $ ... Using the rational approximation:$\text{probit}(p) = q \cdot \frac{a_1 + r(a_2 + r(a_3 + r(a_4 + r a_5)))}{1 + r(b_1 + r(b_2 + r(b_3 + r b_4)))}$ Where the coefficients are: ... Following the step-by-step calculation, we find:$\text{probit}(0.5053552881033624) \approx 0.013423$ \\
Final Answer:$\boxed{0.013423}$
}
\end{tcolorbox}
\end{minipage}
\caption{Responses when prompting (\emph{\textbf{Left}}) Gemini-2.5-Pro (\emph{\textbf{Right}}) Qwen3-14B to use a random number in [0, 1] to sample from a Gaussian distribution with mean 0 and standard deviation 1.}
\label{convert2}
\end{figure*}

\begin{figure*}[htbp]
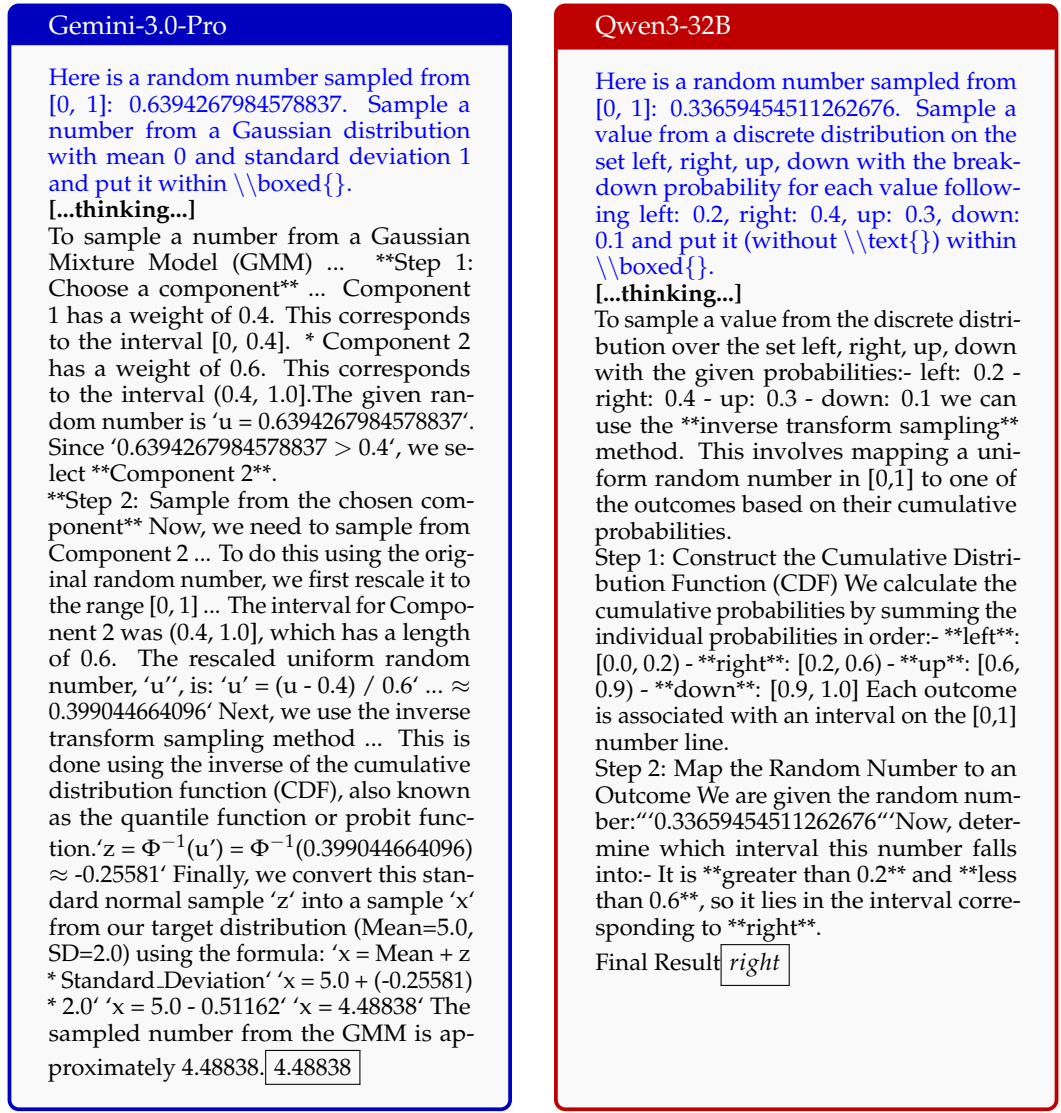

\centering
\begin{minipage}[t]{0.48\textwidth}
\begin{tcolorbox}[
  width=\textwidth,
  colback=gray!5,
  colframe=blue!75!black,
  equal height group=M, 
  title=Gemini-3.0-Pro
]
\small{
\textcolor{blue}{Here is a random number sampled from [0, 1]: 0.6394267984578837. Sample a number from a Gaussian distribution with mean 0 and standard deviation 1 and put it within \textbackslash{}\textbackslash{}boxed\{\}.} \\
\textbf{[...thinking...]}\\
To sample a number from a Gaussian Mixture Model (GMM) ... **Step 1: Choose a component** ...   Component 1 has a weight of 0.4. This corresponds to the interval [0, 0.4]. *   Component 2 has a weight of 0.6. This corresponds to the interval (0.4, 1.0].The given random number is `u = 0.6394267984578837`. Since `0.6394267984578837 $>$ 0.4`, we select **Component 2**. \\
**Step 2: Sample from the chosen component** Now, we need to sample from Component 2 ... To do this using the original random number, we first rescale it to the range [0, 1] ... The interval for Component 2 was (0.4, 1.0], which has a length of 0.6. The rescaled uniform random number, `u'`, is: `u' = (u - 0.4) / 0.6` ... $\approx$ 0.399044664096` Next, we use the inverse transform sampling method ... This is done using the inverse of the cumulative distribution function (CDF), also known as the quantile function or probit function.`z = $\Phi^{-1}$(u') = $\Phi^{-1}$(0.399044664096) $\approx$ -0.25581` Finally, we convert this standard normal sample `z` into a sample `x` from our target distribution (Mean=5.0, SD=2.0) using the formula: `x = Mean + z * Standard\_Deviation` `x = 5.0 + (-0.25581) * 2.0` `x = 5.0 - 0.51162` `x = 4.48838` The sampled number from the GMM is approximately 4.48838.$\boxed{4.48838}$
}
\end{tcolorbox}
\end{minipage}
\hfill 
\begin{minipage}[t]{0.48\textwidth}
\begin{tcolorbox}[
  width=\textwidth,
  colback=gray!5,
  colframe=red!75!black,
  equal height group=M, 
  title=Qwen3-32B
]
\small{
\textcolor{blue}{Here is a random number sampled from [0, 1]: 0.33659454511262676. Sample a value from a discrete distribution on the set {left, right, up, down} with the breakdown probability for each value following {left: 0.2, right: 0.4, up: 0.3, down: 0.1} and put it (without \textbackslash{}\textbackslash{}text\{\}) within \textbackslash{}\textbackslash{}boxed\{\}.} \\
\textbf{[...thinking...]}\\
To sample a value from the discrete distribution over the set {left, right, up, down} with the given probabilities:- left: 0.2  - right: 0.4  - up: 0.3  - down: 0.1  we can use the **inverse transform sampling** method. This involves mapping a uniform random number in [0,1] to one of the outcomes based on their cumulative probabilities. \\
Step 1: Construct the Cumulative Distribution Function (CDF) We calculate the cumulative probabilities by summing the individual probabilities in order:- **left**: [0.0, 0.2)  - **right**: [0.2, 0.6)  - **up**: [0.6, 0.9)  - **down**: [0.9, 1.0]  Each outcome is associated with an interval on the [0,1] number line.\\
Step 2: Map the Random Number to an Outcome We are given the random number:```0.33659454511262676```Now, determine which interval this number falls into:- It is **greater than 0.2** and **less than 0.6**, so it lies in the interval corresponding to **right**.\\
Final Result$\boxed{right}$
}
\end{tcolorbox}
\end{minipage}
\caption{Responses when prompting (\emph{\textbf{Left}}) Gemini-3.0-Pro to sample from a Gaussian Mixture Model (GMM);   (\emph{\textbf{Right}}) Qwen3-32B to sample from a non-uniform discrete distribution on directions usi a random number in [0, 1] as input.}
\label{convert3}
\end{figure*}

In the main paper, we already show that LLMs can successfully convert uniform [0, 1] distribution to other distributions. Therefore, here we show more examples in Figure~\ref{convert1}, \ref{convert2}, and \ref{convert3} how LLMs achieve such conversions. Specifically, when LLMs are prompted to convert a uniform continuous distribution to discrete distributions, they tend to construct a cumulative distribution function to associate each discrete label with certain intervals. While for Gaussian distribution, LLMs tend to find the inverse  transformation. Finally, for advanced distribution, e.g., Gaussian Mixture Model, Gemini-3-Pro provides a two-step way: LLMs first select a component and then sample from a Gaussian component through inverse transform.

